\documentclass[acmtog,nonacm]{acmart}
\acmSubmissionID{242}
\usepackage{multirow}
\usepackage{booktabs} % For formal tables
\usepackage{natbib}
\usepackage{graphicx}
\usepackage{float}
\usepackage{enumitem}
\usepackage{bbding}
\usepackage{soul}
\usepackage{bm}
\usepackage{xspace}
\usepackage{array}
\usepackage{multirow}
\usepackage{overpic}

\usepackage{ulem}

\usepackage[ruled]{algorithm2e} % For algorithms

\SetAlFnt{\small}
\SetAlCapFnt{\small}
\SetAlCapNameFnt{\small}
\SetAlCapHSkip{0pt}

\newcommand{\tb}[1]{{\textbf{#1}}}

\newcommand{\cell}[1]{{\begin{tabular}{@{}c@{}} #1 \end{tabular}}}

\begin{document}
\title{CLAY: A Controllable Large-scale Generative Model for Creating High-quality 3D Assets}

\author{Longwen Zhang}\authornote{Equal contributions.}
\orcid{0000-0001-8508-3359}
\affiliation{%
 \institution{ShanghaiTech University and Deemos Technology Co., Ltd.}
 \city{Shanghai}
 \country{China}}
\email{zhanglw2@shanghaitech.edu.cn}

\author{Ziyu Wang}\authornotemark[1]
\orcid{0000-0002-4697-5183}
\affiliation{%
 \institution{ShanghaiTech University and Deemos Technology Co., Ltd.}
 \city{Shanghai}
 \country{China}}
\email{wangzy6@shanghaitech.edu.cn}

\author{Qixuan Zhang}\authornote{Project leader.}
\orcid{0000-0002-4837-7152}
\affiliation{%
 \institution{ShanghaiTech University and Deemos Technology Co., Ltd.}
 \city{Shanghai}
 \country{China}}
\email{zhangqx1@shanghaitech.edu.cn}

\author{Qiwei Qiu}
\orcid{0009-0005-0213-4744}
\affiliation{%
 \institution{ShanghaiTech University and Deemos Technology Co., Ltd.}
 \city{Shanghai}
 \country{China}}
\email{qiuqw@shanghaitech.edu.cn}

\author{Anqi Pang}
\orcid{0000-0003-2746-6946}
\affiliation{%
 \institution{ShanghaiTech University}
 \city{Shanghai}
 \country{China}}
\email{pangaq@shanghaitech.edu.cn}

\author{Haoran Jiang}
\orcid{0009-0006-9673-8545}
\affiliation{%
 % \institution{ShanghaiTech University}
 \institution{ShanghaiTech University and Deemos Technology Co., Ltd.}
 \city{Shanghai}
 \country{China}}
\email{jianghr1@shanghaitech.edu.cn}

\author{Wei Yang}
\orcid{0000-0002-1189-1254}
\affiliation{%
 \institution{Huazhong University of Science and Technology}
 \city{Wuhan}
 \country{China}}
\email{weiyangcs@hust.edu.cn}

\author{Lan Xu}\authornote{Corresponding author.}
\orcid{0000-0002-8807-7787}
\affiliation{%
 \institution{ShanghaiTech University}
 \city{Shanghai}
 \country{China}}
\email{xulan1@shanghaitech.edu.cn}

\author{Jingyi Yu}\authornotemark[3]
\orcid{0000-0001-9198-6853}
\affiliation{%
 \institution{ShanghaiTech University}
 \city{Shanghai}
 \country{China}}
\email{yujingyi@shanghaitech.edu.cn}

\renewcommand\shortauthors{Zhang and Wang, et al}

\newcommand{\eqtr}{\mathrel{\raisebox{-0.1ex}{%
\scalebox{0.8}[0.6]{$\vartriangle$}}}}

\begin{abstract}

In the realm of digital creativity, our potential to craft intricate 3D worlds from imagination is often hampered by the limitations of existing digital tools, which demand extensive expertise and efforts. To narrow this disparity, we introduce CLAY, a 3D geometry and material generator designed to effortlessly transform human imagination into intricate 3D digital structures. CLAY supports classic text or image inputs as well as 3D-aware controls from diverse primitives (multi-view images, voxels, bounding boxes, point clouds, implicit representations, etc). At its core is a large-scale generative model composed of a multi-resolution Variational Autoencoder (VAE) and a minimalistic latent Diffusion Transformer (DiT), to extract rich 3D priors directly from a diverse range of 3D geometries. Specifically, it adopts neural fields to represent continuous and complete surfaces and uses a geometry generative module with pure transformer blocks in latent space. We present a progressive training scheme to train CLAY on an ultra large 3D model dataset obtained through a carefully designed processing pipeline, resulting in a 3D native geometry generator with 1.5 billion parameters. For appearance generation, CLAY sets out to produce physically-based rendering (PBR) textures by employing a multi-view material diffusion model that can generate 2K resolution textures with diffuse, roughness, and metallic modalities. We demonstrate using CLAY for a range of controllable 3D asset creations, from sketchy conceptual designs to production ready assets with intricate details. Even first time users can easily use CLAY to bring their vivid 3D imaginations to life, unleashing unlimited creativity.

\end{abstract}




\begin{teaserfigure}
    \setlength{\abovecaptionskip}{3pt}
    \centering
    \includegraphics[width=1\textwidth,trim=0 0 0 0,clip]{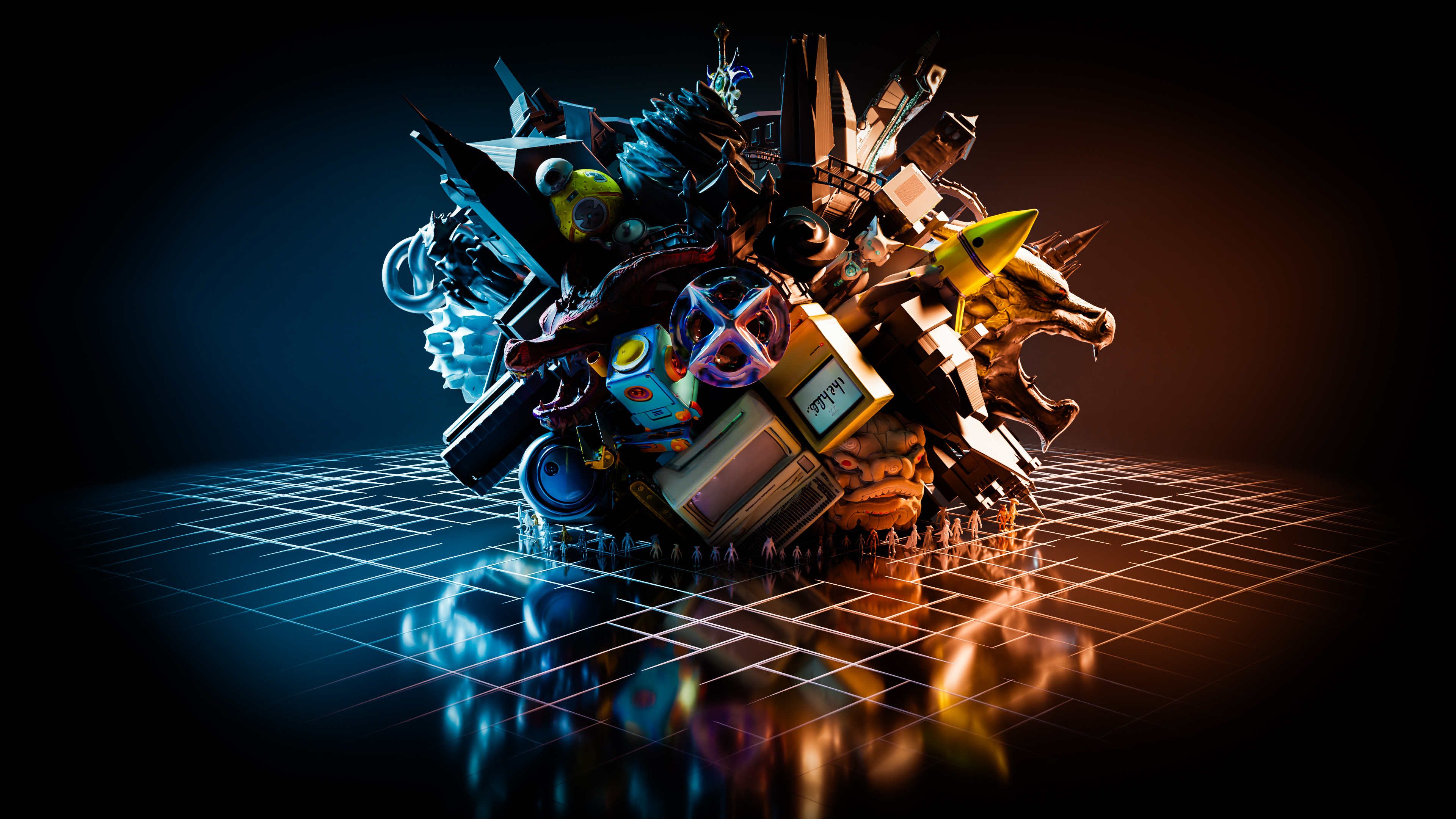}
    \caption{
{
\textit{Against the backdrop of the great digital expanse, CLAY orchestrates a vibrant explosion of 3D creativity, unleashing unlimited imagination.}
}
    }
    \label{fig:teaser}
\end{teaserfigure}

\maketitle

\section{Introduction}

Three-dimensional (3D) imagination allows us humans to visualize and design structures, spaces, and systems before they are physically constructed. When we were kids, we learned to build objects using this imagination, with as simple as clay, stones, or wood sticks, and for the lucky few, LEGO blocks. To us then, a building formed by a few simple blocks can imaginatively transform to a magnificent castle and a wood stick attached to a stone into a LightSaber, Jedi's or Sith's. In fact, with a diverse range of pieces in different shapes, sizes, and colors in hand, we once imagined having virtually unlimited capabilities for creating objects. This boundless imagination has fundamentally transformed the entertainment industry, from feature films to computer games, and has led to significant advances in the field of computer graphics, from modeling to rendering. In contrast, the capabilities of producing creative content by far fall far behind our imagination. For example, the current 3D creation workflow still requires immense artistic expertise and tedious manual labor. An ideal 3D creation tool should conveniently convert our kid-like vibrant imagination into digital reality - it should effortlessly craft geometry and textures and support diverse controllable strategies for creation, translating abstract concepts into tangible, digital forms.

Latest progresses on AI Generated Content (AIGC)~\cite{po2023state} reignite the hope and enthusiasm to bridge imagination and creation, epitomized by the text-based 2D image generation that benefits from the consolidation of large image datasets, effective neural network architectures (e.g., Transformer~\cite{AttentionAllYouNeed}, Diffusion Model~\cite{ddpm}), adaptation schemes (e.g., LoRA~\cite{hu2022lora}, ControlNet~\cite{zhang2023controlnet}), etc. It is not an exaggeration that the 2D creation workflow has largely been revolutionized, perhaps symbolized by the controversial triumph of Midjourney's AI-generated ``Théâtre D’opéra Spatial'' at a digital arts competition. In a similar vein, we have also witnessed rapid progress in 3D asset generation. Yet compared with 2D generation, 3D generation has not yet reached the same level of progress that can fundamentally reshape the 3D creation pipeline. Its model scalability and adaptation capabilities fall far behind mature 2D techniques. The challenges are multi-fold, stemming from the limited scale of quality 3D datasets as well as the inherent entanglement of geometry and appearance of 3D assets.

State-of-the-art 3D asset generation techniques largely build on two distinct strategies: either lifting 2D generation into 3D or embracing 3D native strategies. In a nutshell, the former line of work leverages 2D generative models~\cite{sd, imagen} via intricate optimization techniques such as score distillations~\cite{poole2022dreamfusion,wang2023prolificdreamer}, or further refines 2D models for multi-view generation~\cite{liu2023zero,shi2024mvdream}. They address the diverse appearance generation problem by employing pretrained 2D generative models. As 2D priors do not easily translate to coherent 3D ones, methods based on 2D generation generally lack concise 3D controls (preserving lines, angles, planes, etc) that one would expect in a foundational model and they consequently fail to maintain high geometric fidelity. In comparison, 3D native approaches attempt to train generative models directly from 3D datasets~\cite{chang2015shapenet,deitke2023objaverse} where 3D shapes can be represented in explicit forms such as point clouds~\cite{nichol2022point}, meshes~\cite{nash2020polygen,siddiqui2023meshgpt} or implicit forms such as neural fields~\cite{chen2019learning, zhang20233dshape2vecset}. They can better ``understand'' and hence preserve geometric features, but have limited generation ability unless they employ much larger models, as shown in concurrent works~\cite{yariv2023mosaic,ren2023xcube}. Yet larger models subsequently require training on larger datasets, which are expensive to obtain, the problem that 3D generation aims to address in the first place.

In this paper, we aim to bring together the best of 2D-based and 3D-based generations by following the ``pretrain-then-adaptation'' paradigm adopted in text/image generation, effectively mitigating 3D data scarcity issue. We present \textit{CLAY}, a novel \textit{C}ontrollable and \textit{L}arge-scale generative scheme to create 3D \textit{A}ssets with high-qualit\textit{Y} geometry and appearance. CLAY manages to scale up the foundation model for 3D native geometry generation at an unprecedented quality and variety, and at the same time it can generate appearance with rich multi-view physically-based textures. The 3D assets generated by CLAY contain not only geometric meshes but also material properties (diffuse, roughness, metallic, etc.), directly deployable to existing 3D asset production pipelines. As a versatile foundation model, CLAY also supports a rich class of controllable adaptations and creations (from text prompts to 2D images, and to diverse 3D primitives), to help conveniently convert a user's imagination to creation.

The core of CLAY is a large-scale generative model that extracts rich 3D priors directly from a diverse range of 3D geometries. Specifically, we adopt the neural field design from 3DShape2VecSet~\cite{zhang20233dshape2vecset} to depict continuous and complete surfaces along with a tailored multi-resolution geometry Variational Autoencoder (VAE). We customize the geometry generative module in latent space with an adaptive latent size. To conveniently scale up the model, we adopt a minimalistic latent diffusion transformer (DiT) with pure transformer blocks to accommodate the adaptive latent size. We further propose a progressive training scheme to carefully increase both the latent size and model parameters, resulting in a 3D native geometry generator with 1.5 billion parameters. The quality of training samples is crucial for fine-grained geometry generation, especially considering the limited size of available 3D datasets. We hence present a new data processing pipeline to standardize the diverse 3D data and enhance the data quality. Specifically, it includes a remeshing process that converts various 3D surfaces into occupancy fields, preserving essential geometric features such as sharp edges and flat surfaces. At the same time, we harness the capabilities of GPT-4V~\cite{openai2023gpt4v} to produce robust annotations that accentuate these geometric characteristics.

The combination of new architecture, training scheme, and training data in CLAY leads to a novel 3D native generative model that can create high-quality geometry, serving as the foundation to downstream model adaptations. 
For appearance generation, the scarcity of abundant data poses a significant challenge for synthesizing material texture maps. To tackle this issue, CLAY sets out to generate multi-view physically-based rendering (PBR) textures, and subsequently project them onto geometry.
We construct a multi-view material diffusion model analogous to 2D diffusion model~\cite{sd} but trained on high-quality PBR textures from Objaverse~\cite{deitke2023objaverse}, to efficiently generate diffuse, roughness, and metallic modalities while avoiding tedious distillation. We further extend the diffusion model to support super-resolution as well as to accurately map the multi-view textures onto the generated geometry. The modified model allows for much faster high-quality textures generation than traditional optimization methods, producing 2K resolution in the UV space for realistic rendering.

We further explore various adaptation schemes including LoRA-like fine-tuning and cross-attention-based conditioning, to support classic text or image-based creations as well as 3D-aware controls from diverse primitives (multi-view images, voxels, bounding boxes, point clouds, implicit representations, etc). These extensive adaptation capabilities of CLAY hence enable controllable 3D asset creation ranging from sketchy conceptual designs to more sophisticated ones with intricate details. Even first time users can use CLAY to bring their vivid 3D imaginations to life with our tailored interactive controls: a bustling village can be generated from scattered bounding boxes across a barren landscape, a spacecraft with futuristic wings and propulsion system from craft blocks with textual descriptions, and ultimately creations from imaginations.

 %   \setlength\itemsep{0em} 

  %  \item We explore the pretraining-adaptation paradigm for 3D generation and introduce a large 3D native model (up to 1.5B parameters) to generate high-quality geometry, equipped with a concise scalable training scheme.

   % \item Building upon the generated geometry, we introduce a novel asset enhancement to efficiently generate physically-based textures, enhancing the realism of our 3D creations.

    %\item We showcase diverse adaptation techniques in our 3D generative model, enabling various interactive and controllable applications from text, image, or even 3D modalities.

\begin{figure*}
    \centering
    \includegraphics[width=\linewidth]{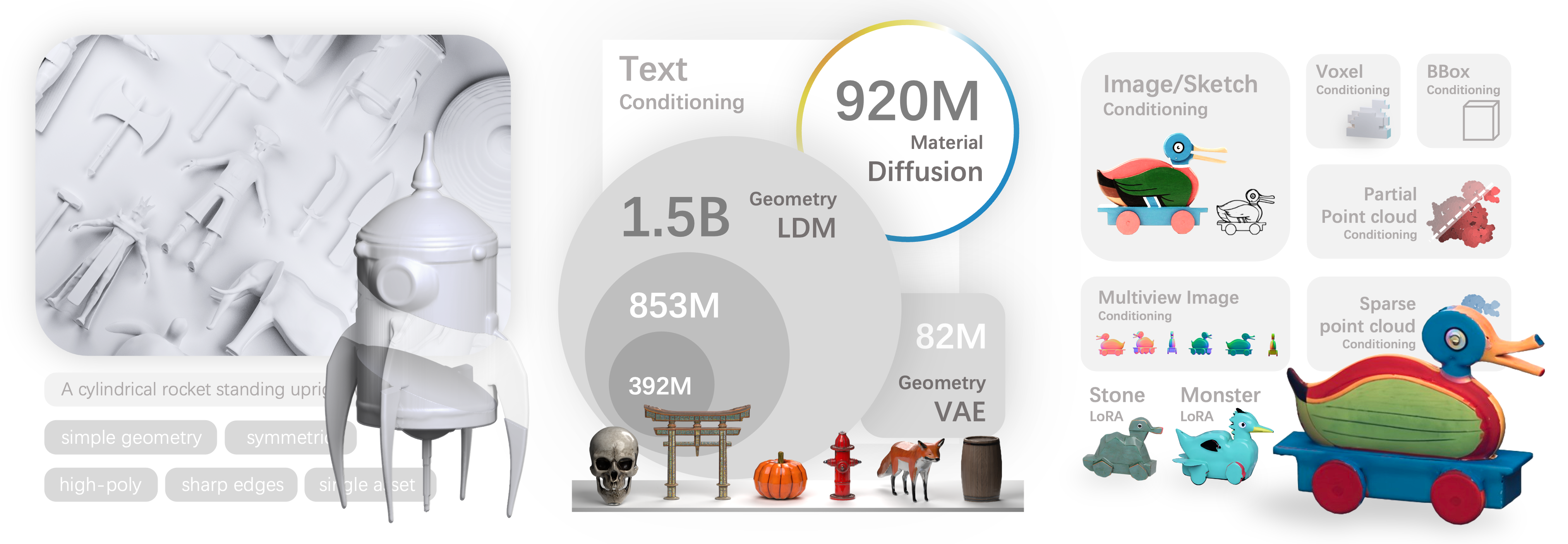}
    \vspace{-0.9cm}
    \caption{An overview of our CLAY framework for 3D generation. Central to the framework is a large generative model trained on extensive 3D data, capable of transforming textual descriptions into detailed 3D geometries. The model is further enhanced by  physically-based material generation and versatile modal adaptation, to enable the creation of 3D assets from diverse concepts and ensure their realistic rendering in digital environments. }
    \label{fig:overview}
    % \vspace{-0.3cm}
\end{figure*}

\section{Related Work}
3D generation is undoubtedly the fastest-growing research arena in AIGC. Efficient and high quality 3D asset creation via generation benefits entertainment and gaming industry as well as film and animation productions. Previous practices have explored different routes, ranging from directly training on 3D datasets, to imposing generated 2D images as priors, and to imposing 3D priors on top of 2D generation.

\paragraph{Imposing 2D Images as Prior}

3D generation methods in this category attempt to exploit significant strides made in 2D image generation, exemplified by latest advances such as DALL·E~\cite{Dall-e}, Imagen~\cite{imagen} and Stable Diffusion~\cite{sd}. Extending this prowess to 3D generation, many approaches have adopted image-based techniques, focusing on transforming 2D images into 3D structures or imposing 2D images as priors. DreamFusion~\cite{poole2022dreamfusion} pioneered this practice by introducing Score Distillation Sampling (SDS) and employed 2D image generation with viewpoint prompts to produce 3D shapes via NeRF~\cite{mildenhall2021nerf} optimization. Although the idea is intriguing, earlier attempts struggled to consistently produce high-quality and diverse results. Often, generating satisfactory results requires repeated adjustments to parameters and long waits of optimizations.
Subsequent enhancements in SDS have explored the possibility of extending the idea to various neural fields
~\cite{
lin2023magic3d, yu2023points,zhu2023hifa,huang2023dreamtime,wu2024hd,GSGEN}, ranging from DMTet~\cite{shen2021deep} to the most recent 3D Gaussian splatting~\cite{kerbl3Dgaussians}, 
Various modifications managed to elevate the performance~\cite{seo2024let, wang2023prolificdreamer,li2023sweetdreamer,chen2023fantasia3d, dreamface, latentnerf}. 
Yet a critical challenge remains: 2D image diffusion models utilized in SDS still lack an explicit understanding of neither geometry nor viewpoint. The lack of perspective information and explicit 3D supervision can lead to the 
multi-head Janus problem, where realistic 3D renderings do not translate to view consistency and every rendered view can be deemed as the front view.

To mitigate the problem, Zero-1-to-3~\cite{liu2023zero} proposes to integrate view information into the image generation process. This can be achieved by training an additional mapping from the transformation matrix to the pretrained Stable Diffusion model, enabling the network to obtain some prior knowledge on view position and distribution. 
Alternative solutions attempt to employ SDS to optimize a coherent neural field~\cite{qian2023magic123, zhang2023optimized, tang2023dreamgaussian, sun2023dreamcraft3d}, but they generally require long optimization time. 
Latest developments~\cite{liu2024syncdreamer,shi2024mvdream,shi2023zero123++,li2023instant3d,long2023wonder3d,qiu2023richdreamer,blattmann2023stable} have focused on directly generating multi-view images with view consistency, by employing enhanced attention mechanisms. These approaches have significantly improved multi-view image generation, achieving a higher level of consistency.

The downside there is the need to fine-tune Stable Diffusion using additional images either by conducting multi-view rendering ~\cite{deitke2023objaverse} or using auxiliary multi-view  datasets~\cite{reizenstein2021common,wu2023omniobject3d, Yu_2023_CVPR}. Since the multi-view results can already be used to extract 3D shapes (e.g., via multi-view stereo or neural methods), techniques such as SyncDreamer~\cite{liu2024syncdreamer} and Wonder3D~\cite{long2023wonder3d} employed NeuS~\cite{wang2021neus} to accelerate generation.
One-2-3-45~\cite{liu2023one2345} has gone one step further to train generalizable NeuS~\cite{long2022sparseneus} on 3D datasets, to tackle sparse view inputs.
Since the starting point of all these approaches are 2D images, they unanimously focus on the quality of generated images without attempting to preserve geometric fidelity.
As a result, the generated geometry often suffers from incompleteness and lacks details.

\paragraph{Imposing 3D Geometry as Priors}

To address challenges in 2D-based techniques, an emerging class of solutions attempt to impose 3D shapes as priors. 
Even though One-2-3-45~\cite{liu2023one2345} is viewed as using 2D image priors, the clever use NeuS as geometry proxy reveals the possibility of imposing 3D shape priors. For example, Instant3D~\cite{li2023instant3d}, LRM~\cite{hong2023lrm, wang2023pf}, 
DMV3D~\cite{xu2023dmv3d} and TGS~\cite{zou2023triplane}
further utilized sparse-view or single-view reconstructors that leverage a Vision Transformer (ViT) as the vision backbone, coupled with a deep transformer architecture to directly reconstruct NeRF with both color and density attributes. They are hence commonly referred to Large Reconstruction Models (LRMs).
Yet these techniques still focus on minimizing the volume rendering loss rather than explicitly generating surfaces, resulting in coarse or noisy geometry.  

Apparently, the most straightforward practice to generate 3D would be to train on 3D datasets, rather than 2D images or image-induced 3D shapes. 
Early approaches~\cite{choy20163d,fan2017point,mescheder2019occupancy,groueix2018papier,tang2019skeleton,tang2021skeletonnet} primarily utilized 3D convolutional networks to understand the 3D grid structure.
Point-E~\cite{nichol2022point} 
took a pioneering step by leveraging a pure transformer-based diffusion model for denoising directly on the point clouds. 
This method is notable for its simplicity and efficiency, yet it faces great difficulties in transforming the generated point clouds into precise, common mesh surfaces. 
Polygen~\cite{nash2020polygen} and MeshGPT~\cite{siddiqui2023meshgpt} take a different approach by natively representing meshes through points and surface sequences. These models are capable of producing extremely high-quality meshes, but their dependence on small, high-quality datasets restricts their broader applicability.
XCube~\cite{ren2023xcube} introduces a strategy that simplifies geometry into multi-resolution voxels before diffusion. It streamlines the process but faces challenges in managing complex prompts and supporting a broad range of downstream tasks, limiting its overall flexibility. It is worth mentioning that different 3D generation techniques have relied on different datasets. This is not surprising as they are based on different geometric representation but problematic as it is essential to have a unified dataset that includes all available shapes.

One such attempt is to represent geometry uniformly in terms of Signed Distance Field (SDF)~\cite{park2019deepsdf,yariv2023mosaic}, occupancy fields~\cite{peng2020convolutional,tang2021sa}, or both~\cite{zheng2023locally, liu2023one}, and train directly on 3D datasets. Such approaches provide a more explicit mechanism than NeRF for learning and extracting surfaces but require the latent encoding of watertight meshes for generation. 
Models such as DeepSDF~\cite{park2019deepsdf} and Mosaic-SDF~\cite{yariv2023mosaic} utilize optimization techniques to create unique representations for each geometry in the training dataset, which is not efficient during training as they do not benefit from autoencoders.
Other models such as SDFusion~\cite{cheng2023sdfusion} and ShapeGPT~\cite{yin2023shapegpt} adopt an intuitive 3D VAE (Variational Autoencoder) for encoding geometries and reconstructing SDF fields. These methods, primarily trained or tested on the ShapeNet~\cite{chang2015shapenet} dataset, are limited in the diversity and variety of shapes they can generate. 
3DGen~\cite{gupta20233dgen} employs a triplane VAE for both encoding and decoding SDF fields whereas Shap-E~\cite{jun2023shap}, 3DShape2VecSet~\cite{zhang20233dshape2vecset}, and Michelangelo~\cite{zhao2023michelangelo} adopt a different trajectory by utilizing transformers to encode the input point clouds into parameters for the decoding networks, signifying a shift towards more sophisticated neural network architectures in 3D generative models.

By far methods that aim to direct learning from 3D datasets, while capable of producing better geometries than 2D-based generation, still cannot match the hand-crafted ones by artists, in either detail or complexity. We observe, through the development of CLAY, this is mainly because they have not sufficiently explored rich geometric features embedded in the datasets. In addition, their small model size limits the capability of generalization and diversification. In CLAY, we resort to tailored geometry processing to mine a variety groups of 3D datasets as well as discuss effective techniques to scale up the generation model.

\section{Large-scale 3D Generative Model}

An effective 3D generative model should be able to generate 3D contents from different conditional inputs such as text, images, point clouds, and voxels. As aforementioned, the task is challenging in how to define a 3D model: should 3D asset be viewed in terms of geometry with per-vertex color or geometry with a texture map? should the 3D geometry be inferred from the generated appearance data or be directly generated? In CLAY, we adopt a minimalist approach, i.e., we separate the geometry and texture generation processes. This indicates that we choose not to use 2D generation techniques which potentially help 3D geometry generation (e.g., through reconstruction). In our experiment, we find that once we manage to scale up the 3D generation model and train it with sufficiently large amount of high quality data, the directly generated 3D geometry by CLAY exceeds previous 2D generation based/assisted techniques by a large margin, in both diversity and quality (e.g., geometric details).

In a nutshell, CLAY is a large 3D generative model with 1.5 billion parameters, pretrained on high-quality 3D data. The significant upscaling from prior art is key to improving its capabilities in generation diversity and quality. %
Architecture-wise, CLAY extends the generative model in 3DShape2VecSet~\cite{zhang20233dshape2vecset} with a new multi-resolution Variational Autoencoder (VAE). This extension enables more efficient geometric data encoding and decoding. In addition, we complement CLAY with an advanced latent Diffusion Transformer (DiT) for probabilistic geometry generation.
Dataset-wise, we have developed a remeshing pipeline, along with annotation schemes powered by GPT-4V~\cite{openai2023gpt4v}, to standardize and unify existing 3D datasets. These datasets historically have not been used together for training a 3D generation model as they are in different formats and lack consistencies. Our combined dataset after processing maintains a consistent representation and coherent annotations. We show that putting the model architecture and training dataset together greatly improves 3D generation.

\begin{figure}
    \centering
    \includegraphics[width=1\linewidth]{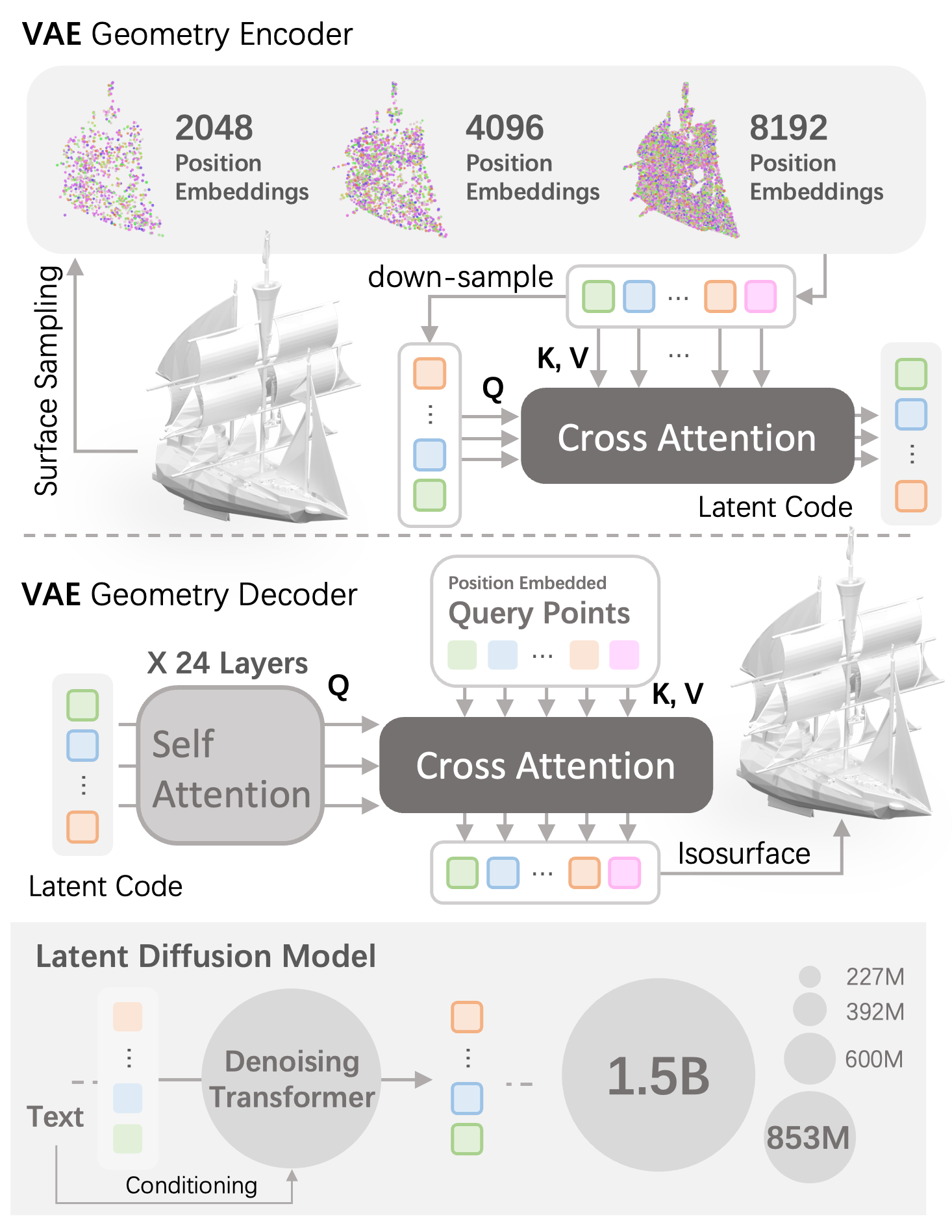}
    \vspace{-0.8cm}
    \caption{Network design of our VAE and DiT. With a minimalist design, our DiT supports scalable training and VAE operates effectively across various geometric resolutions.  }
    \label{fig:network}
    \vspace{-0.2cm}
\end{figure}

\begin{table*}[t]
\centering
\caption{DiT specifications and training hyper parameters.}
    \vspace{-0.3cm}
\begin{tabular}{lcccccccc}\hline\hline
Model size & $n_\text{params}$ & $n_\text{layers}$ & $d_\text{model}$ & $n_\text{heads}$ & $d_\text{head}$ & Latent length & Batch size & Learning rate \\\hline\hline
Tiny   & 227M & 24 & 768   & 12 & 64   & 512                       & 1024                      & 1e-4                      \\\hline
Small   & 392M & 24 & 1024  & 16 & 64   & \cell{512\\1024}          & \cell{16384\\8192}        & \cell{1e-5\\5e-6}       \\\hline
Medium  & 600M & 24 & 1280  & 16 & 80   & \cell{512\\1024}          & \cell{16384\\8192}        & \cell{1e-4\\5e-5}         \\\hline
Large   & 853M & 24 & 1536  & 16 & 96   & \cell{512\\1024\\2048}    & \cell{8192\\4096\\2048}   & \cell{1e-4\\1e-5\\5e-6} \\\hline
XL      & 1.5B & 24 & 2048  & 16 & 128  & \cell{512\\1024\\2048}    & \cell{4096\\2048\\1024}   & \cell{1e-4\\1e-5\\5e-6}   \\\hline

\end{tabular}
\label{tab:model_specs}
\vspace{-0.4cm}
\end{table*}

\subsection{Representation and Model Architecture}

Our approach for a 3D generative model emphasizes on learning to denoise 3D data in a compressed latent space, analogous to the foundation 2D generative models. This strategy significantly reduces the complexity and is computationally much more efficient than directly working in 3D space. We adopt the representation and architecture from 3DShape2VecSet but augment it with new scaling-up strategies. 
Specificially, we encode a 3D geometry into latent space by sampling a point cloud $\mathbf{X}$ from a 3D mesh surface $\mathbf{M}$. 
This point cloud is encoded into a latent code with dynamic shape $\mathbf{Z} = \mathbb{R}^{L\times 64}$ with a length $L$ and channel size 64 using the encoder $\mathcal{E}$ of a transformer-based VAE, expressed as $\mathbf{Z} = \mathcal{E}(\mathbf{X})$. We then learn a DiT to denoise the latent code  $\mathbf{Z}_t$ with noise at step $t$. Finally, the VAE decoder $\mathcal{D}$ decodes the generated latent codes from DiT into a neural field, as $\mathcal{D}(\mathbf{Z}_0, \mathbf{p}) \rightarrow [0, 1]$, where $\mathbf{p}$ is a testing coordinate in space, and $\mathcal{D}$ determines if $\mathbf{p}$ is inside or outside the 3D shape.
Recall our objective is to achieve substantial scaling-up of this architectural model. To maintain robust scale-up while facilitating effective training, we develop a new scheme based on multi-resolution encoding. Such an extension not only enhances the model's capacity to manage large-scale data but also ensures refined training outcomes, underpinning the model's performance, scalability, and adaptability.

\paragraph{Multi-resolution VAE} In the design of our VAE module, we follow the structure outlined in 3DShape2VecSet. This involves embedding the input point cloud $\mathbf{X}\in \mathbb{R}^{N\times 3}$ sampled from a mesh $\mathbf{M}$ into a latent code using a learnable embedding function and a cross-attention encoding module:
\begin{equation}
     \mathbf{Z} =\mathcal{E}(\mathbf{X})=\text{CrossAttn}(\text{PosEmb}(\mathbf{\tilde X}),\text{PosEmb}(\mathbf{X})),
\end{equation}
where $\mathbf{\tilde X}$ denotes a down-sampled version of $\mathbf{X}$ at 1/4 scale, effectively reducing the latent code's length $L$ to a quarter of the input point cloud size $N$.
The VAE's decoder, consisting of 24 self-attention layers and a cross-attention layer, processes these latent codes and a list of query points $\mathbf{p}$, outputting occupancy logits:
\begin{equation}
\mathcal{D}(\mathbf{Z},\mathbf{p})=\text{CrossAttn}(\text{PosEmb}(\mathbf{p}),\text{SelfAttn}^{24}(\mathbf{Z})).
\end{equation}
Our VAE is dimensioned at 512 with 8 attention heads, culminating in a total of 82 million parameters. The latent code size is configured as $L\times 64$, with $L$ varying based on the input point cloud size.

In 3DShape2VecSet, the point clouds are generally of small sizes and therefore are insufficient to capture fine geometric details. We adopt a multi-resolution approach. 
At each iteration, we first randomly choose a sampling size $N$ from 2048, 4096, or 8192, to ensure variability. Next, we sample the corresponding number of surface points from the input mesh $\mathbf{M}$.

\paragraph{Coarse-to-fine DiT}
Our DiT employs a minimalistic yet effective structure, consisting of a 24-layer pure transformer, with added cross-attention mechanisms for accommodating text prompt conditions. The encoding process involves sampling $N=4L$ surface points from a 3D mesh, which are subsequently encoded into a latent code $\mathbf{Z}\in \mathbb{R}^{L\times 64}$ using $\mathcal{E}(\cdot)$. In parallel, a pretrained language model, specifically CLIP-ViT-L/14~\cite{radford2021clip}, processes the text prompt into textual features $\mathbf{c}$. The DiT's role, defined as $\epsilon(\cdot)$, is to predict the noise in $\mathbf{Z}_t$ at timestep $t$:
\begin{equation}
\epsilon(\mathbf{Z}_t,t,\mathbf{c})=\{\text{CrossAttn}(\text{SelfAttn}(\mathbf{Z}_t\#\#\mathbf{t}),\mathbf{c})\}^{24},
\end{equation}
where the symbol $\#\#$ signifies concatenation, and for clarity, certain elements like projection and feed-forward layers are omitted from this description.
To efficiently capture fine geometric details, we optimize the DiT on high-dimensional latent sets. Specifically, we employ a progressive training scheme, varying the latent code length for quicker convergence and time efficiency. Starting with a length of latent code $L=512$ at a higher learning rate, we gradually increase to 1024, then to 2048, each time reducing the learning rate based on empirical observations. This progressive scaling method ensures robust and efficient training of our DiT.

\paragraph{Scaling-up Scheme}
Scaling-up CLAY requires enhancing both the VAE and DiT architectures with pre-normalization and GeLU activation, to facilitate faster computation of attention mechanism.  The feed-forward dimension is four times of the model dimension.
For noise scheduling, a discrete scheduler with 1000 timesteps is employed, and a cosine beta schedule is utilized during training. Following the latest practice on diffusion training~\cite{lin2024common}, we implement zero terminal SNR by rescaling betas and opt for ``v-prediction'' as our training objective, a strategy that promotes stable inference.
To evaluate the impact of model size on performance, we train five DiTs with sizes varying from 227 million to 1.5 billion parameters, as outlined in Table.~\ref{tab:model_specs}. Our smallest model, designed for verification, can be trained on a single node with 8 NVidia A800 GPUs due to its smaller batch size, to support preliminary experiments. For larger models, we employed  larger batch sizes, resulting in improved training stability and faster convergence rates. Our largest model, the XL, was trained on a cluster of 256 NVidia A800 GPUs, for approximately 15 days, with progressive training.

Following the insights in \citet{gesmundo2023composable} of \textit{Head addition}, \textit{Heads expansion} and \textit{Hidden dimension expansion}, we progressively scale up the DiT during training. This approach offers benefits such as enhanced time efficiency, improved knowledge retention, and a reduced risk of the model trapped in the local optima. This scaling-up process in DiT training, leveraging the suggested training techniques, is designed to optimize the model's learning trajectory and overall performance.

Our model, once trained on our expanded dataset (Sec.~\ref{sec:data}), demonstrates strong capabilities to generate 3D objects from text prompts at a high quality and accuracy. 
During inference, we utilize a 100-timestep denoising process with linear-space timestep spacing for efficient 3D geometry generation. 
The model then engages in dense sampling at a $512^3$ grid resolution with our VAE's geometry decoder, precisely determining occupancy values for detailed geometry capture, which are then converted to mesh using Marching Cubes.

\subsection{Data Standardization for Pretraining}
\label{sec:data}

The effectiveness and robustness of large-scale 3D generative models rely on the quality and the scale of 3D datasets. Unlike text and 2D images which are abundant and hence can support Stable Diffusion, 3D datasets such as ShapeNet~\cite{chang2015shapenet} and Objaverse~\cite{deitke2023objaverse} are limited in size or quality. To obtain large-scale high quality 3D data, it is essential to overcome challenges such as non-watertight meshes, inconsistent orientations and inaccurate annotation. Our solution is to apply a remeshing method for geometry unification and GPT-4V~\cite{openai2023gpt4v} for precise automatic annotation. 
Our standardization starts with filtering out unsuitable data, such as complex scenes and fragmented scans, resulting in a refined collection of 527K objects from ShapeNet and Objaverse, laying a robust groundwork for enhanced model performance through tailored unification and annotation techniques.

\paragraph{Geometry Unification}

\begin{figure}[t]
    \centering

\setlength{\tabcolsep}{-1pt}
\newcommand{\w}{2.2cm}
\newcommand{\lef}{5}
\newcommand{\dow}{-1}
\footnotesize
\begin{tabular}{cccc}

    % \multirow{2}{*}{{\includegraphics[width=1.8cm,trim=150 0 150 0,clip]{fig/remesh1/table2.png}}\hspace{2pt}} 
    % \raisebox{0.5cm}{\multirow{2}{*}{{\includegraphics[width=1.8cm,trim=170 0 170 0,clip]{fig/remesh1/table2.png}}\hspace{2pt}}}

    % &

    %  \raisebox{-2cm}{
    % \begin{overpic}[width=\w,trim=150 0 150 0,clip]{fig/remesh1/04379243_719c8fe47dc6d3d9b6b5a7b8c31557c_paint.png}
    % \put(\lef,\dow){\color{white} Input}
    % \end{overpic}
    % }
    %   &
    %  \raisebox{-2cm}{
    % \begin{overpic}[width=\w,trim=150 0 150 0,clip]{fig/remesh1/manifold_paint.png}
    % \put(\lef,\dow){\color{white} Manifold}
    % \end{overpic}
    % }
    %   &
    %  \raisebox{-2cm}{
    % \begin{overpic}[width=\w,trim=150 0 150 0,clip]{fig/remesh1/manifoldplus_paint.png}
    % \put(\lef,\dow){\color{white} ManifoldPlus}
    % \end{overpic}
    % }
    % \\
    % &
    %  \raisebox{-2cm}{
    % \begin{overpic}[width=\w,trim=150 0 150 0,clip]{fig/remesh1/mesh_to_sdf_paint.png}
    % \put(\lef,\dow){\color{white} mesh-to-sdf}
    % \end{overpic}
    % }
    %   &
    %  \raisebox{-2cm}{
    % \begin{overpic}[width=\w,trim=150 0 150 0,clip]{fig/remesh1/mesh2sdf_paint.png}
    % \put(\lef,\dow){\color{white} mesh2sdf}
    % \end{overpic}
    % }
    %   &
    %  \raisebox{-2cm}{
    % \begin{overpic}[width=\w,trim=150 0 150 0,clip]{fig/remesh1/zlw_paint.png}
    % \put(\lef,\dow){\color{white} Ours}
    % \end{overpic}
    % }
    
    %  \\\\
 
    % \multirow{2}{*}{{\includegraphics[width=1.8cm,trim=150 0 150 0,clip]{fig/remesh2/chair2.png}}\hspace{2pt}} 
    \raisebox{-1cm}{\multirow{2}{*}{{\includegraphics[width=1.8cm,trim=60 0 40 0,clip]{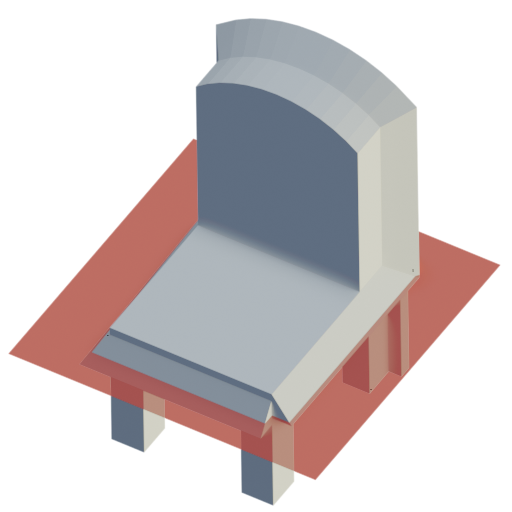}}\hspace{2pt}}}

    &

     \raisebox{-2cm}{
    \begin{overpic}[width=\w,trim=150 0 150 0,clip]{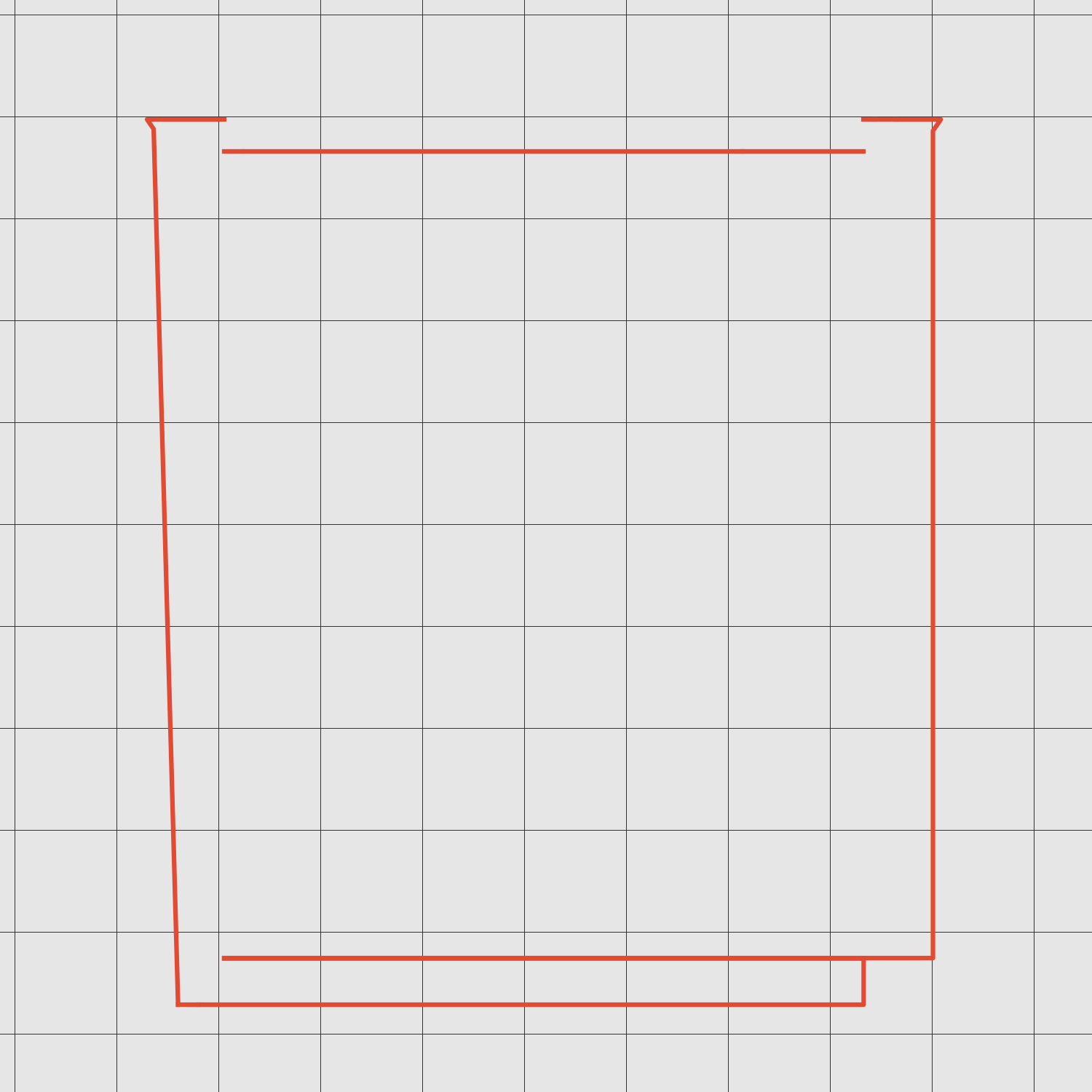}
    \put(\lef,\dow){\color{black} Input}
    \end{overpic}
    }
      &
     \raisebox{-2cm}{
    \begin{overpic}[width=\w,trim=150 0 150 0,clip]{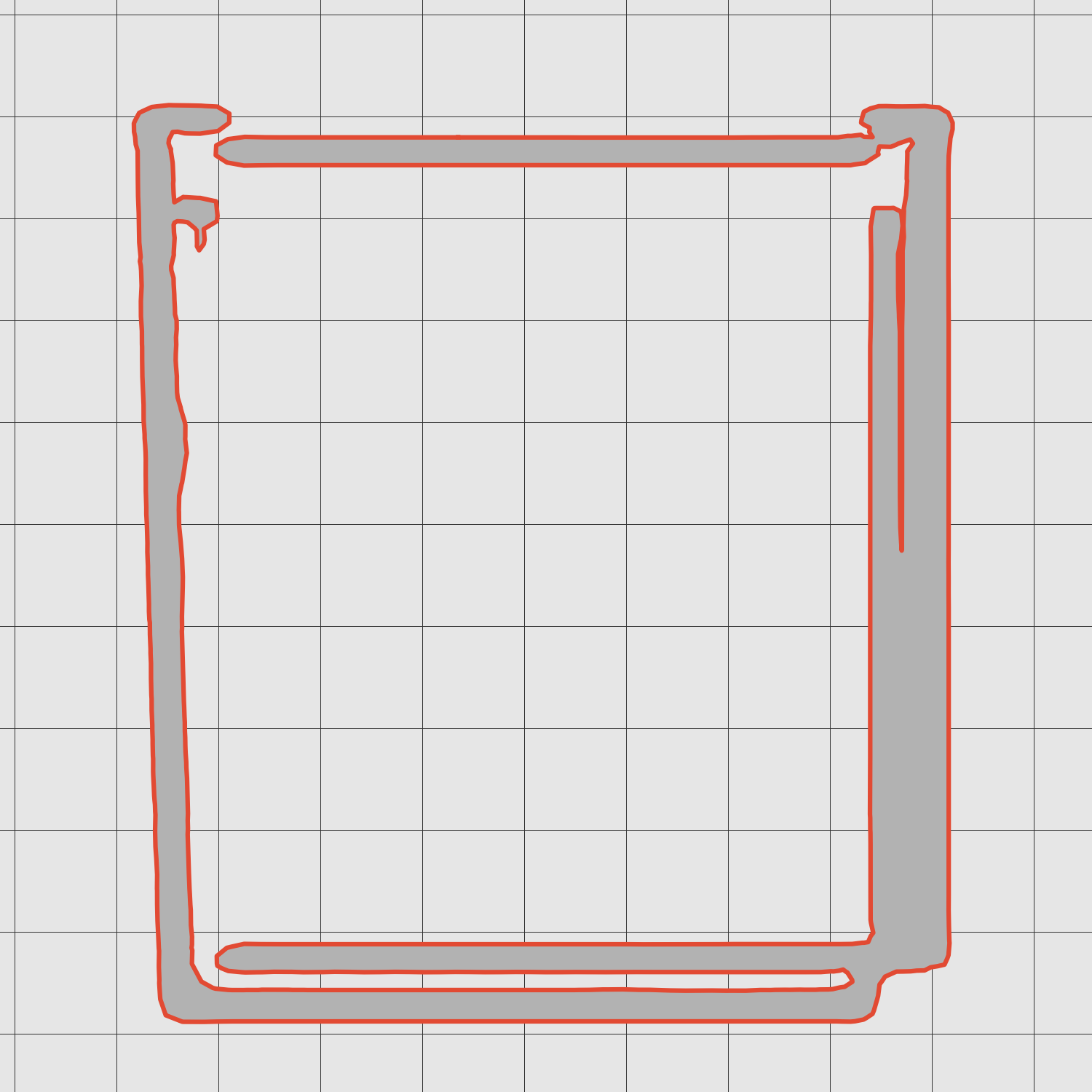}
    \put(\lef,\dow){\color{black} Manifold [\citeyear{huang2018robust}]}
    \end{overpic}
    }
      &
     \raisebox{-2cm}{
    \begin{overpic}[width=\w,trim=150 0 150 0,clip]{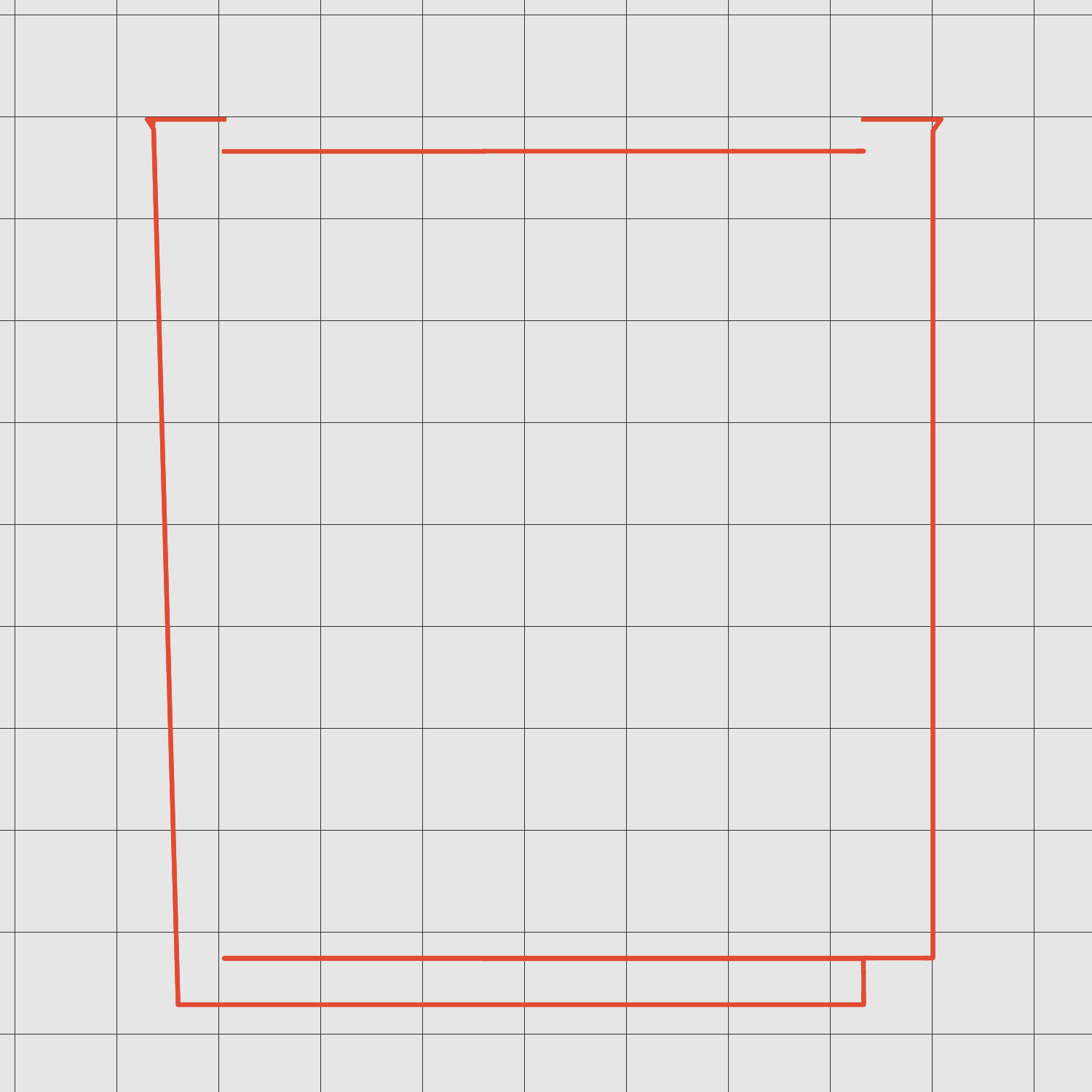}
    \put(\lef,\dow){\color{black} ManifoldPlus [\citeyear{huang2020manifoldplus}]}
    \end{overpic}
    }\vspace{0.1cm}
    \\
    &
     \raisebox{-2cm}{
    \begin{overpic}[width=\w,trim=150 0 150 0,clip]{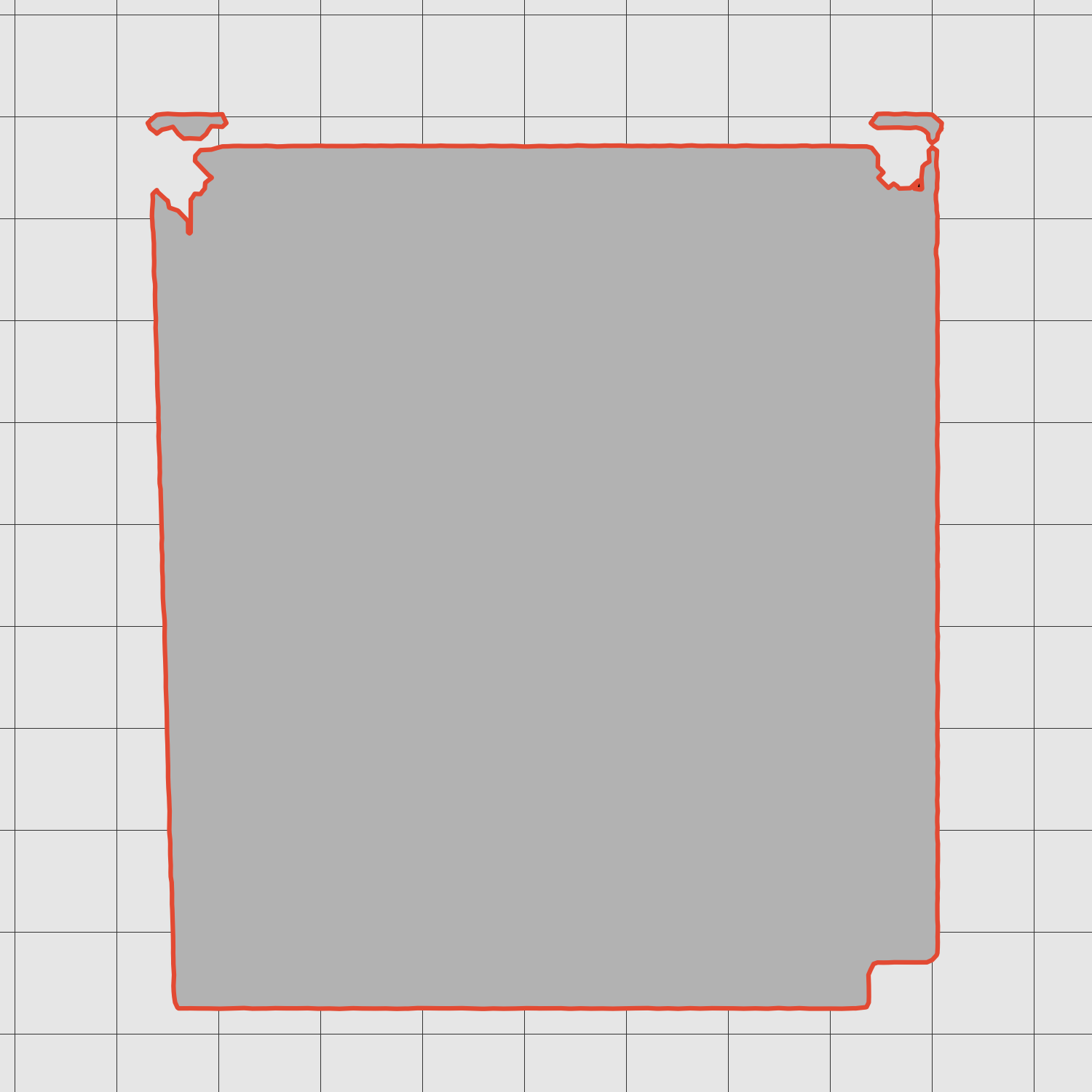}
    \put(\lef,\dow){\color{black} mesh-to-sdf [\citeyear{mesh_to_sdf}]}
    \end{overpic}
    }
      &
     \raisebox{-2cm}{
    \begin{overpic}[width=\w,trim=150 0 150 0,clip]{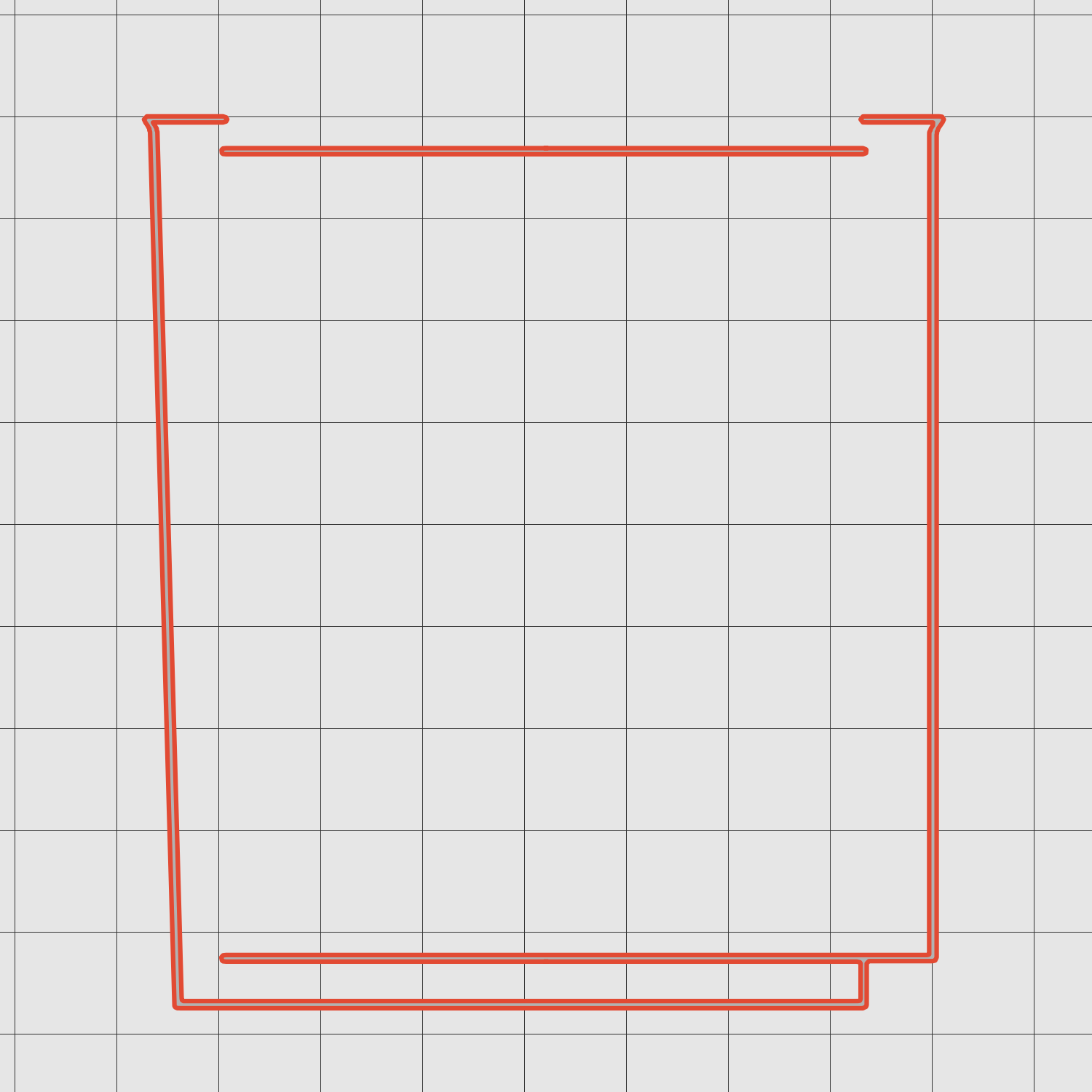}
    \put(\lef,\dow){\color{black} DOGN [\citeyear{Wang-2022-dualocnn}]}
    \end{overpic}
    }
      &
     \raisebox{-2cm}{
    \begin{overpic}[width=\w,trim=150 0 150 0,clip]{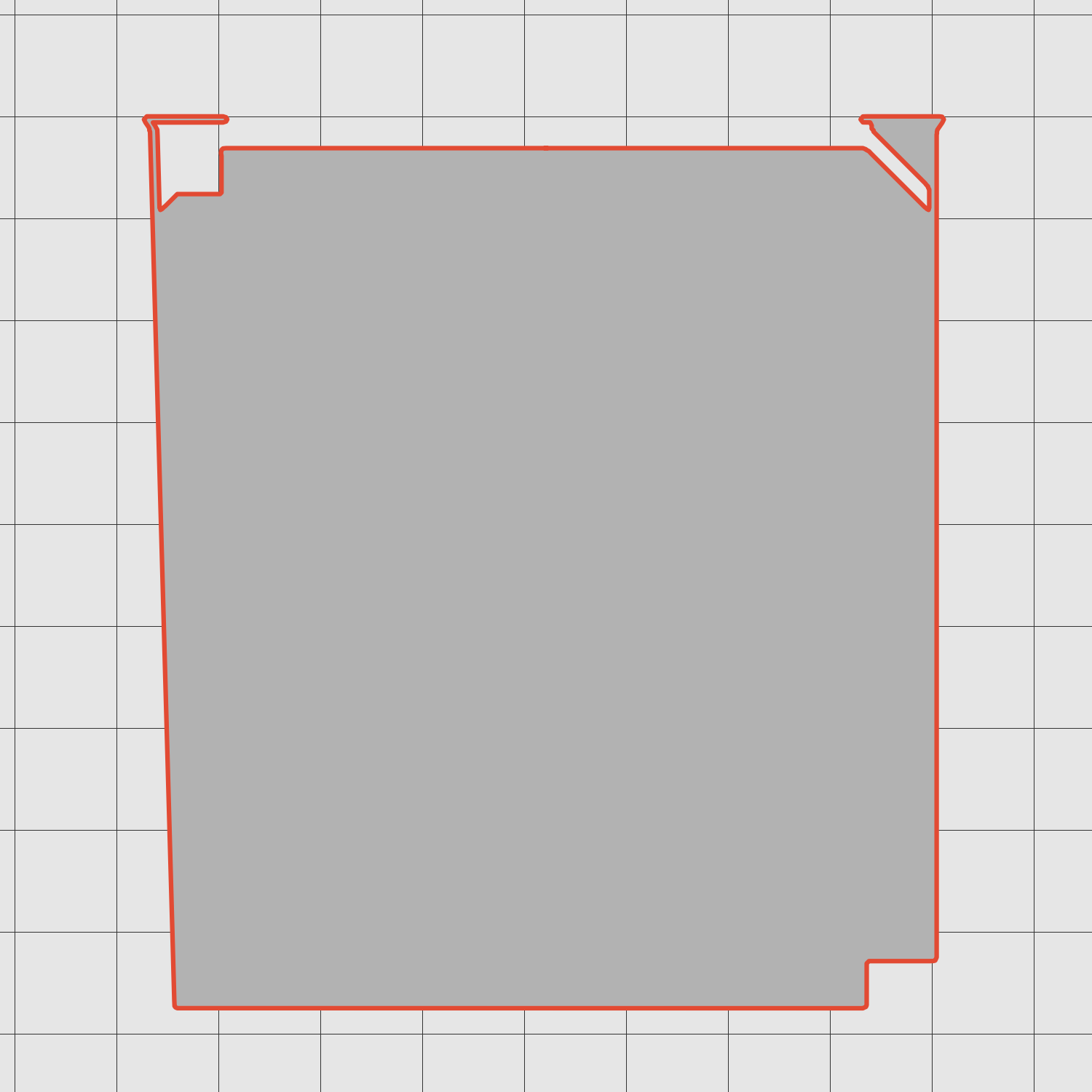}
    \put(\lef,\dow){\color{black} Ours}
    \end{overpic}
    }
\end{tabular}

    \vspace{-0.2cm}
    
    \caption{Comparison against existing mesh preprocessing methods using cross-sectional analysis.
    % The input includes a table with the tabletop represented by a single face, and a non-watertight chair with its surface not closed. 
    The input is a non-watertight chair with its surface not closed. 
    Red lines correspond to the faces of meshes, light gray indicates ``outside'' and dark gray indicates ``inside''. Our method maximizes positive volume while faithfully preserving geometric features. This robustness extends to non-watertight input meshes, ensuring consistent and reliable results.}
    \label{fig:remesh}
    \vspace{-0.4cm}
\end{figure}

To address the challenge of predicting a 3D shape's occupancy field in the presence of non-watertight meshes after data filtration, we propose a standardized geometry remeshing protocol to ensure watertightness while avoiding discarding useful data in the training set.
Popular remeshing tools such as Manifold~\cite{huang2018robust}, while efficient, tend to smooth edges and corners, with its updated version, ManifoldPlus~\cite{huang2020manifoldplus}, showing improved but inconsistent results. Alternatives such as ``mesh-to-sdf''~\cite{mesh_to_sdf} and Dual Octree Graph Networks (DOGN)~\cite{wang2022mesh2sdf, Wang-2022-dualocnn} set out to compute Signed and Unsigned Distance Fields but they are computationally costly.
As depicted in Fig.~\ref{fig:remesh}, the quality of training data for advanced 3D models is affected by these remeshing techniques, underscoring the need for a strategy that balances precision and efficiency. Specific criteria for effective remeshing include: (1) Geometric Preservation - maintaining essential geometric features with minimal alteration; (2) Volume Conservation - ensuring the integrity of all structural elements; and (3) Adaptability to Non-Watertight Meshes - proficiently managing non-watertight models to preserve volumetric accuracy essential for model training.

Inspired by DOGN~\cite{wang2022mesh2sdf, Wang-2022-dualocnn}, we adopt the Unsigned Distance Field (UDF) representation because of its seamless conversion capabilities between mesh formats and correction of inconsistencies in vertex and face density.
In addition, the traditional Marching Cubes algorithm for isosurface extraction can produce a mere thin shell in scenarios involving mesh holes. To address this, we employ a grid-based visibility computation before isosurface extraction. Specifically, we label a grid point as ``inside'' when completely obscured from all angles, maximizing volume for stable VAE training.

\begin{figure*}
    \centering
    \includegraphics[width=1\linewidth]{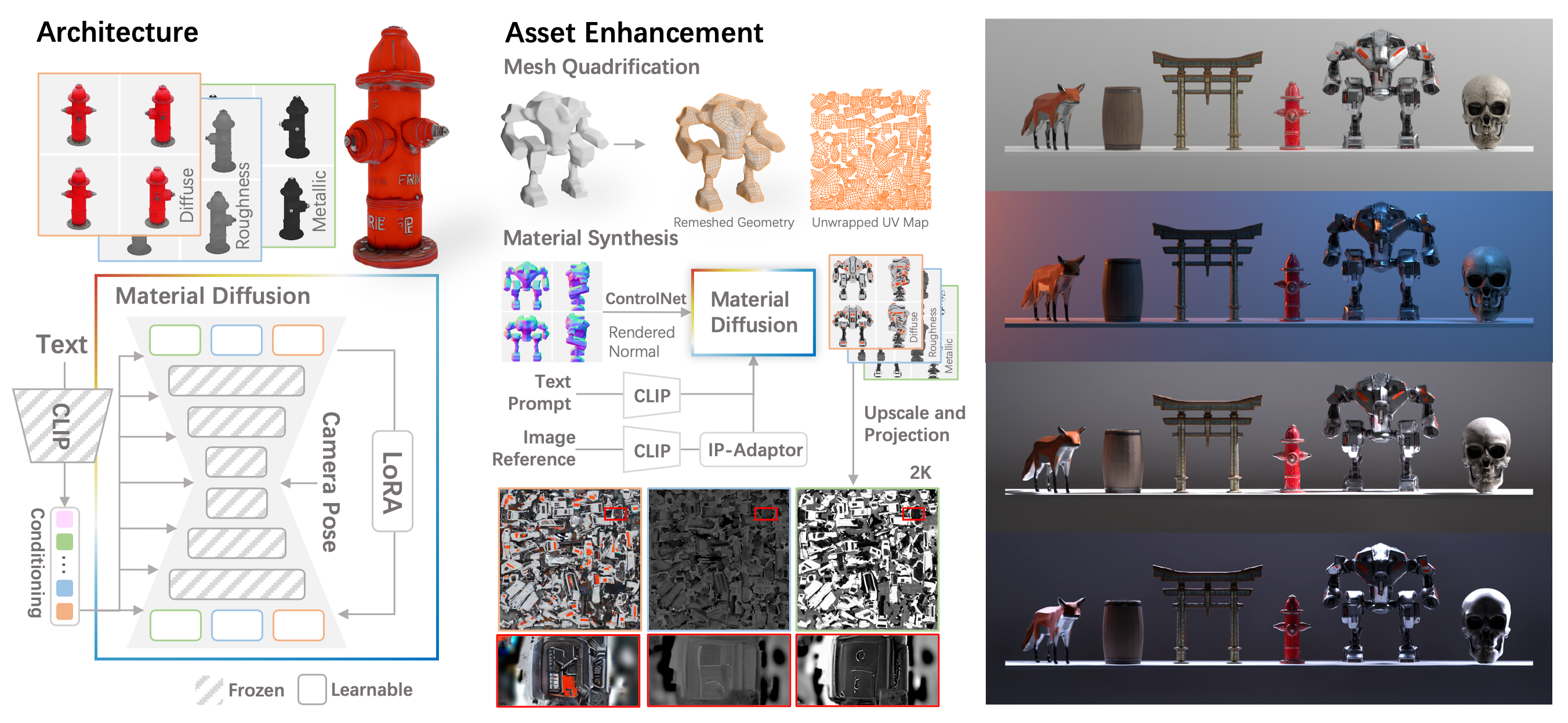}
    \vspace{-0.8cm}
    \caption{Our Material Diffusion architecture and Asset Enhancement pipeline. Our Material Diffusion network, derived from existing diffusion models, facilitates efficient fine-tuning. Following mesh quadrification and atlasing, it generates textures through a multi-view approach and subsequently back-projecte them onto UV maps. The resultant materials, closely aligned with geometries and user inputs (text/image), faithfully respond to diverse lighting conditions, culminating in realistic renderings.}
    \label{fig:enhancement}
    % \vspace{-0.2cm}
\end{figure*}

\paragraph{Geometry Annotation}

The impact of text prompts on 2D image generation by models such as Stable Diffusion~\cite{sd} and SDXL~\cite{podell2023sdxl} reveals the importance of precise prompts in any successful 3D generative model. Previous studies have demonstrated how ``magic prompts'' guide specific content and style. Recognizing this, we emphasize accurate textual prompts in our 3D model to capture geometric and stylistic details of objects. We have developed unique prompt tags and utilized GPT-4V~\cite{openai2023gpt4v} for producing detailed annotation, enhancing the model's capability to interpret and generate complex 3D geometries with nuanced details and diverse styles.

\section{Asset Enhancement}

To make the generated digital assets directly usable in existing CG pipelines, we further adopt a two-stage scheme: post-generation geometry optimization and material synthesis. Geometry optimization ensures structural integrity and compatibility as well as refines the model's form aesthetically and functionally. Material synthesis is crucial for adding lifelike qualities through realistic textures and materials. Together, these steps transform coarse meshes into more engaging assets in digital environments.

\paragraph{Mesh Quadrification and Atlasing}

In CLAY, the initial geometric meshes via the Marching Cubes algorithm typically consist of millions of uneven triangles. While suitable for early stages, such structure poses challenges in editing and application, notably when exported to mesh editing tools or game engines. In addition, it would require complicated automatic UV unwrapping — a crucial step in texture mapping and material synthesis.
To overcome these challenges, we transform these triangle-faced meshes into quad-faced ones using off-the-shelf tools~\cite{quadriflow,Blender}, preserving key geometric features such as sharp edges and flat surfaces. This quadrification process is highly crucial for yielding high-quality final meshes, facilitating the effective conversion from coarse 3D models to the refined assets.

\paragraph{Material Synthesis}

In addition to geometry generation, it is equally important to produce high quality textures in 3D generation. The physically-based rendering (PBR) materials, typically consisting of diffuse, metallic, and roughness textures, are essential for conveying convincing visual experiences in digital environments. Existing methods in PBR texture generation by far have focused on creating a very small subset of these materials. In addition, these approaches lack supervision on specific material attributes, limiting the rendering quality. For example, RichDreamer~\cite{qiu2023richdreamer} generates diffuse maps without roughness and metallic predictions. Fantasia3D~\cite{chen2023fantasia3d} and UniDream~\cite{liu2023unidream} can produce roughness and metallic attributes but do not consider richer attributes. Therefore they cannot generate richer material types.

We aim to synthesize a wide range of PBR materials including diffuse, roughness, and metallic textures. From Objaverse~\cite{deitke2023objaverse}, we carefully choose over 40,000 objects, each characterized by high-quality PBR materials. Utilizing this dataset, we developed a multi-view Material Diffusion to synthesize textures with a significantly speed-up over existing methods, which are then accurately mapped onto the geometries' UV space in a way similar to TEXTure~\cite{TEXTure}.

We modify MVDream~\cite{shi2024mvdream}, originally designed for image space generation, to suit the need for generation from texture attributes with additional channels and modalities. Inspired by HyperHuman~\cite{liu2023hyperhuman}, we integrate three branches into its UNet's outer most convolutional layers, each with skip connections, allowing concurrent denoising across various texture modalities and ensuring view consistency. Similar to MVDream, our training process includes selecting orthogonal-view rendered texture images for each 3D object in training data, and applying both full-parameter for add-on layers and LoRA-based fine-tuning for inside layers, focusing on generating high-quality, view-consistent PBR materials.
Following the same training regimen, our model capably synthesizes texture images from four camera viewpoints, aligned precisely with the input geometry. This is achieved by applying the pretrained ControlNet~\cite{zhang2023controlnet}, with each target view's rendered normal map as inputs. Such an approach not only ensures geometric accuracy but also allows for image-based input customization via IPAdapter~\cite{ye2023ip-adapter}.
To further enhance texture detail, we employ a targeted inpainting approach as introduced in Text2Tex~\cite{chen2023text2tex},
and integrate advanced super-resolution techniques Real-ESRGAN~\cite{wang2021realesrgan} and MultiDiffusion~\cite{bar2023multidiffusion}, achieving 2K texture resolution sufficient for most realistic rendering tasks. 
Our Material Diffusion scheme enables the creation of high-quality textures, resulting in production quality rendering. Our generation results are of a much higher quality and visual pleasantness than previous 3D generation schemes
enhancing engagement and realism of the generated 3D assets.

\begin{figure}
    \centering
    \includegraphics[width=\linewidth]{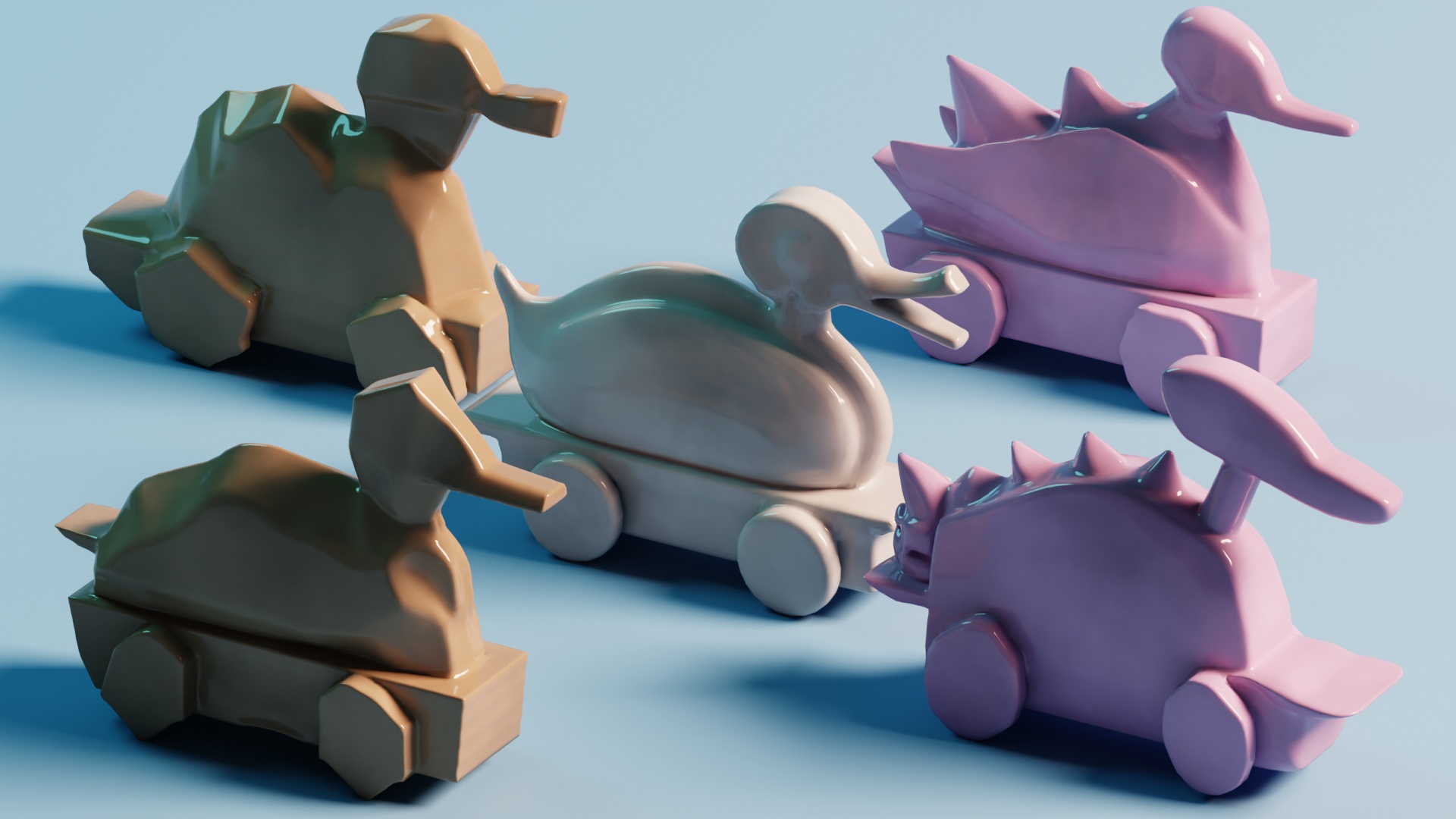}
    \caption{Generation after LoRA fine-tuning on different specific datasets including the rock dataset and the pocket monster dataset. 
    After generating a LEGO duck (center), which was one of the first toys designed by LEGO founder \textit{Ole Kirk Kristiansen}, CLAY can further generate variants in stone styles (left) and pocket monster styles (right).}
    \label{fig:lora}
    \vspace{-0.4cm}
\end{figure}

\begin{figure*}
    \centering
    \includegraphics[width=\linewidth]{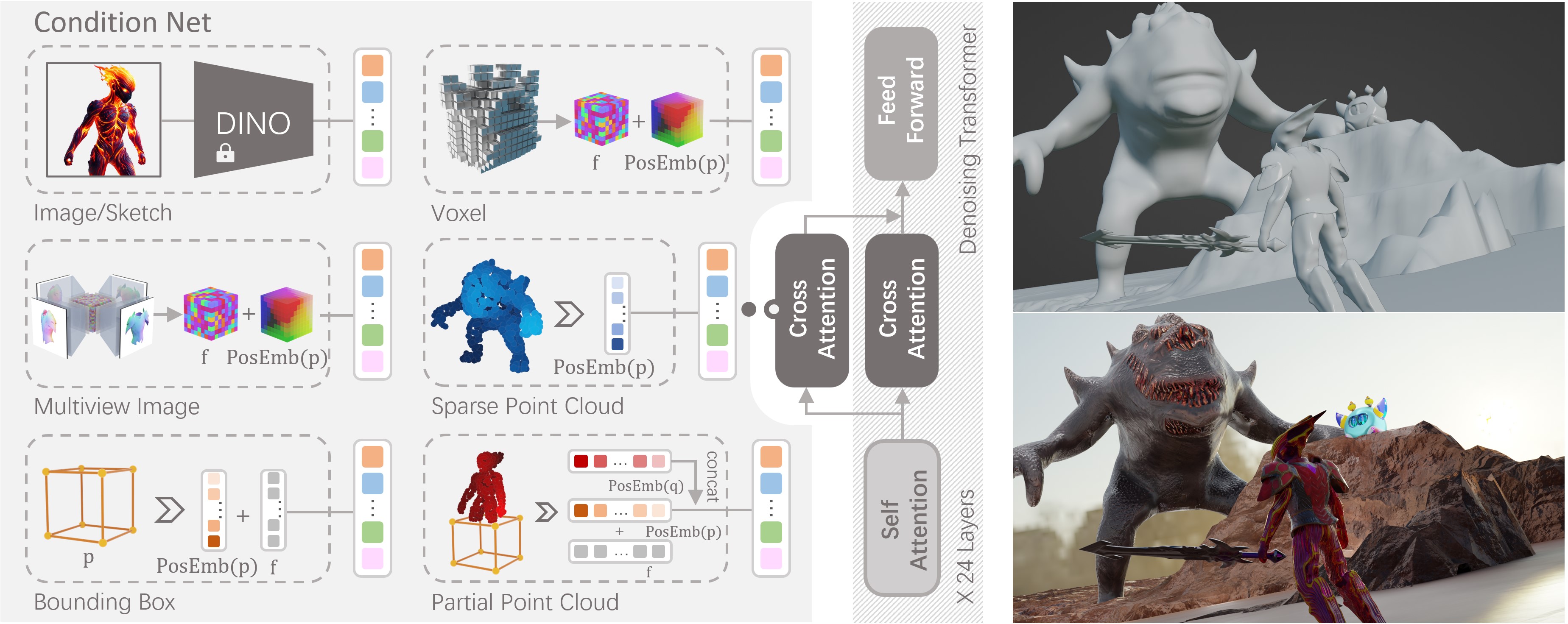}
        \vspace{-0.7cm}
    \caption{Illustration of our network's conditioning design across various modalities. When used together, they support the creation of cinematic scenes with lifelike renderings. }
    \label{fig:condition}
\end{figure*}

\section{Model Adaptation}

CLAY, when pretrained, also serves as a versatile foundation model. For example, CLAY directly supports Low-Rank Adaptation (LoRA) on the attention layers of our DiT. This allows for efficient fine-tuning, enabling the generation of 3D content targeted to specific styles, as illustrated in Fig.~\ref{fig:lora}.
Further, the minimalistic architecture enables us to efficiently support various conditional modalities to support conditioned generation. We implement several exemplary conditions that can be easily provided by a user, including text, which is natively supported, as well as image/sketch, voxel, multi-view images, point cloud, bounding box, and partial point cloud with an extension box. These conditions, which can be applied individually or in combination, enable the model to either faithfully generate content based on a single condition or create 3D content with styles and user controls blended from multiple conditions, offering a wide range of creative possibilities.

\subsection{Conditioning Scheme} 

Building upon our existing text prompt conditioning, we extend the model to incorporate additional conditions in parallel. 
Our use of pre-normalization~\cite{xiong2020layer} converts the attention results into residuals, enabling the addition of extra conditions as parallel residuals alongside the text condition, which can be expressed as:
\begin{equation}
\mathbf{Z} \gets \mathbf{Z}+\text{CrossAttn}(\mathbf{Z},\mathbf{c})+\sum_{i=1}^n\alpha_i\text{CrossAttn}_i(\mathbf{Z},\mathbf{c}_i),
\label{eqn:condition}
\end{equation}
where  $\text{CrossAttn}$ denotes the original text conditioning, $\text{CrossAttn}_i$ denotes the $i$-th additional trainable module and $\mathbf{c}_i$ is the $i$-th condition. The inclusion of scalar $\alpha_i$ in this residual framework allows for direct manipulation of the influence exerted by each additional condition.

While this conditioning scheme is general, obtaining the embedded condition $\mathbf{c}_i$ requires careful calibration. For image/sketch conditions, we utilize the pretrained DINOv2~\cite{oquab2023dinov2} model to extract features as conditions and directly integrate using the cross-attention in the above equation. However, for spatially related modalities such as voxel, multi-view images, point cloud, bounding box, and partial point cloud with an extension box, directly applying cross-attention on features do not guarantee to preserve spatial information pertaining to those conditions. To maintain spatial integrity, we have devised a specific learning strategy.

\paragraph{Spatial Control}

Our 3D geometry generative model incorporates conditions in 3D modalities, a unique feature absent in previous approaches. This allows for spatial controls similar to those in 2D diffusion models. However, different from 3D UNet structures with convolutional backbones that naturally maintain spatial resolution, our approach uses a VAE that dynamically generates latent codes interwoven with spatial coordinates, imposing a new set of challenges for achieving precise spatial controls.

To address the integration of 3D conditions, we set out to learn additional positional embeddings for spatial features. This allows our attention layer to differentiate point coordinates from their features effectively. We start by associating the feature embedding $\mathbf{f}\in \mathbb{R}^{M\times C}$, learned during fine-tuning or extracted from a backbone network, with sparse 3D points $\mathbf{p}\in \mathbb{R}^{M\times 3}$ sampled based on the type of condition being used, where $M$ and $C$ are the length and channels of specific conditioning embedding. The exact sampling strategy is tailored to each condition type and will be detailed subsequently. 
We then apply cross attention more specifically as:
\begin{equation}
\text{CrossAttn}_i(\mathbf{Z},\mathbf{f}+\text{PosEmb}(\mathbf{p})),
\end{equation}
where $\text{PosEmb}(\cdot)$ is the learnable positional embedding. This method allows for the effective integration of various 3D modalities into our model.

\begin{table}
    \vspace{-0.2cm}
    \centering
    \caption{Conditioning module specifications.} 
    \vspace{-0.4cm}
    \begin{tabular}{lcccc}\hline\hline
    Conditioning  & $n_\text{params}$ & $M$ & $C$  & Backbone \\\hline\hline
    Image/Sketch     & 352M & 257 & 1536 & DINOv2-Giant\\
    Voxel     & 260M & 8$^3$ & 512 & /\\
    Multi-view images  & 358M  &  8$^3$   & 768 &  DINOv2-Small\\
    Point cloud & 252M & 512   & 512 & /\\
    Bounding box  &  252M & 8  & 512 & /\\
    Partial point cloud & 252M & 2048+8 & 512 & /\\
    \hline
    \end{tabular}
    \label{tab:conditionnet}
    % \vspace{-0.3cm}
    \vspace{-0.4cm}
\end{table}

\subsection{Implementation}
We discuss how to implement a variety of conditions for controlled 3D content generation. Each condition involves independently training an additional $\text{CrossAttn}_i(\cdot)$ while keeping other parameters fixed.
Fig.~\ref{fig:condition} and Table.~\ref{tab:conditionnet} showcase the specifications and hyperparameters of training for each condition. The base model and training data is described in Sec.~\ref{sec:result}.

\paragraph{Images and Sketches}
For image and sketch conditions, we use the pretrained Vision Transformer (ViT) DINOv2 to extract both patch and global features.
These features are integrated into CLAY via cross-attention, as indicated in Eqn.~\ref{eqn:condition}. This module is trained using rendered RGB images and corresponding sketches from our dataset, ensuring alignment between the generated 3D models and the visual characteristics of the conditioning images or sketches.

\paragraph{Voxel}
Voxels represent spatial cubes and provide an intuitive medium for 3D construction. To integrate voxel-based guidance, we initially construct a $16^3$ voxel grid for each 3D object in our dataset, marking each cell as occupied or vacant. These voxel grids are down-sampled to a $8^3$ feature volume using 3D convolution. The volume features $\mathbf{f}\in \mathbb{R}^{8^3\times C}$, added with positional embeddings of volume centers $\text{PosEmb}(\mathbf{p})$, are then flattened and integrated into the DiT through cross-attention. After training, CLAY can generate 3D geometries that correspond to user-defined voxel structures, effectively translating abstract voxel designs into intricate 3D forms.

\paragraph{Bounding Boxes}
Bounding boxes provide a straightforward way for users to control the aspect ratio and position of 3D objects, essential in interactive generation applications. 
The bounding box features $\mathbf{f}\in\mathbb{R}^{8\times C}$, added with positional embeddings $\text{PosEmb}(\mathbf{p})$, are learned during condition fine-tuning, enabling precise spatial control.

\paragraph{Sparse Point Cloud}
Point clouds offer an easily accessible abstraction for 3D shapes. CLAY can use sparse point clouds as conditions, to generate variants from input meshes or points. For this, we set feature embeddings $\mathbf{f}=0$, which indicates no feature embedding, and sample 512 points as $\mathbf{p}$ and learn the corresponding positional embedding $\text{PosEmb}(\mathbf{p})$. This allows CLAY to generate detailed 3D geometries based on sparse surface point clouds while maintaining the overall shape and appearance.

\paragraph{Multi-view Images}

CLAY also supports multi-view images or multi-view normal maps as conditions, offering spatial control through projected views of 3D geometries. As a demonstration, we use DINOv2 to extract features from various views' images generated by the Wonder3D. These features are back-projected into a 3D volume similar to previous method~\cite{liu2024syncdreamer}, then down-sampled and flattened for integration into the DiT using cross-attention, a similar procedure to the voxel condition.

\paragraph{Partial Point Cloud with Extension Box}
This condition specifically aims to address the point cloud completion task, where a certain bounding box indicates the generation region of missing parts. We merge the input point cloud with the corner points of an extension box, applying a similar approach for learning bounding box conditioning and sparse point cloud conditioning by concatenating these two set of features. This integration is instrumental in the effective reconstruction of incomplete geometries, precisely within the specified extension areas.

\begin{figure*}
    \centering
    \includegraphics[width=\linewidth]{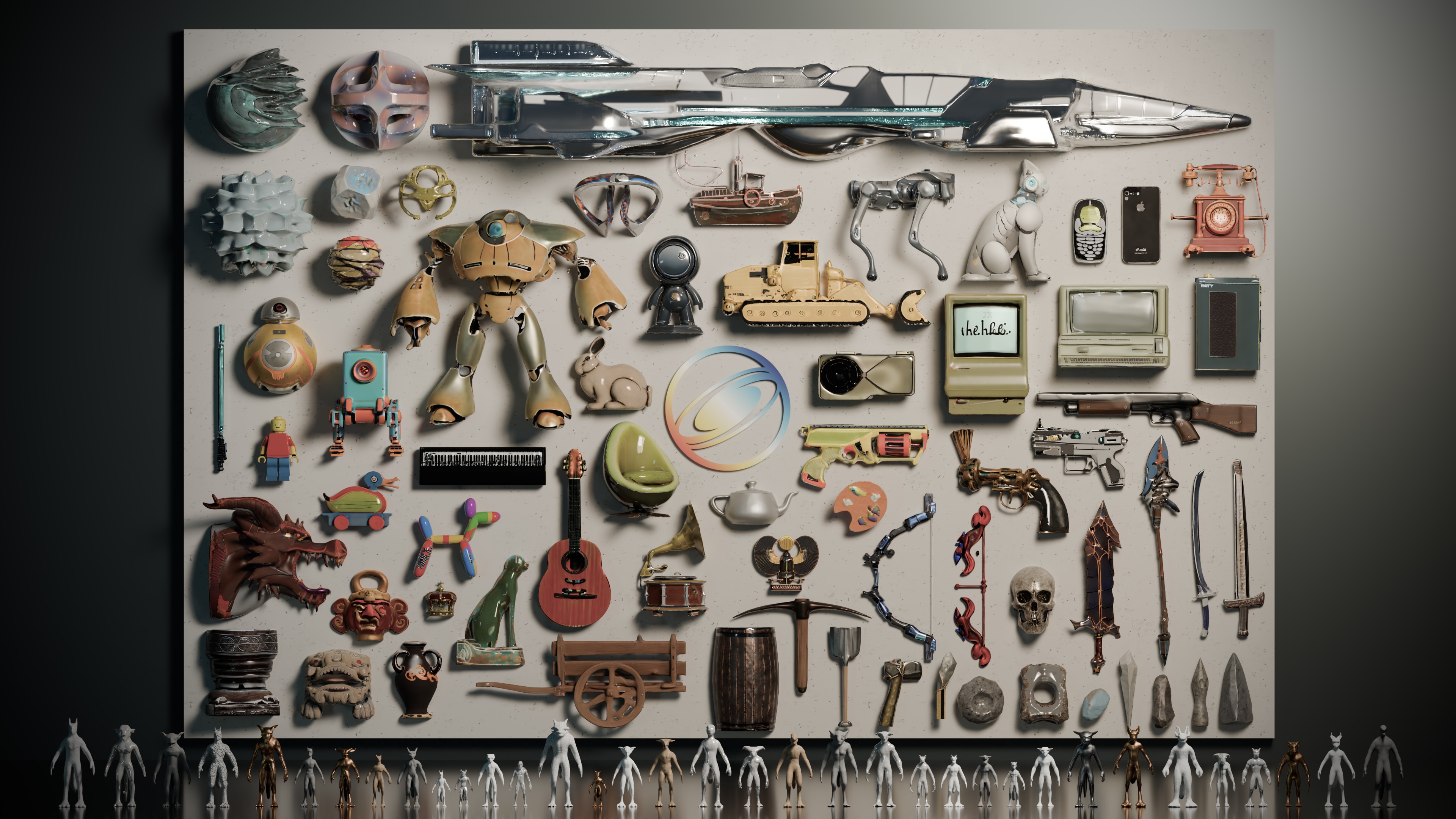}
    \vspace{-0.7cm}
    \caption{Evolution of human innovation, from primitive tools and cultural artifacts to modern electronics and futuristic imaginings, generated by CLAY.}
    \label{fig:gallery}
    \vspace{-0.1cm}
\end{figure*}

\begin{figure*}
    \centering
    \includegraphics[width=1\linewidth]{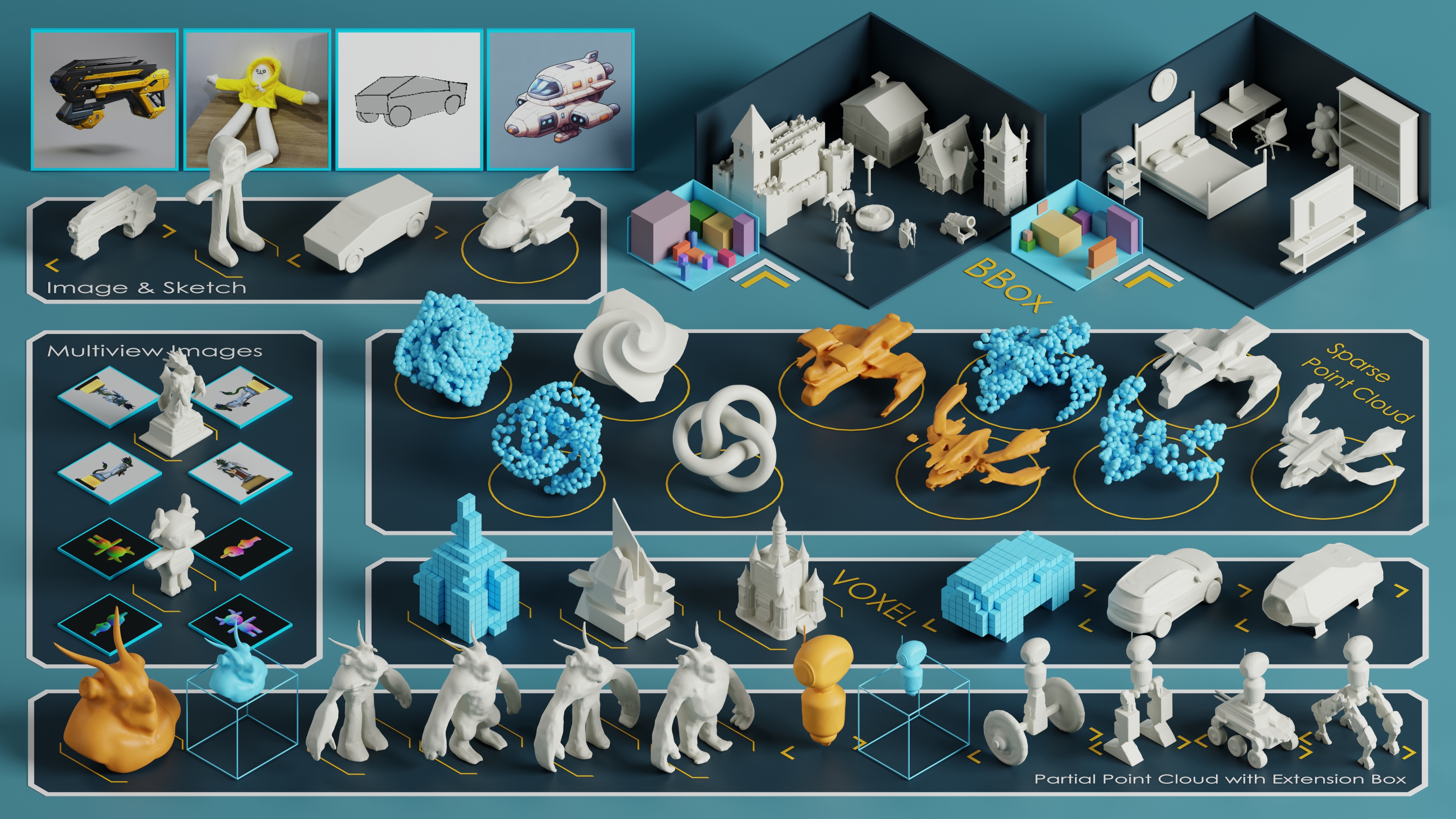}
    \vspace{-0.7cm}
    \caption{
    Sample creations using CLAY, with conditions marked in sky blue and input geometries for respective conditioning (if available) in sandy brown.
    % gallery for condition
    % sky blue, sandy brown
    }
    \label{fig:gallery2}
\end{figure*}

\begin{table*}
    \centering
    \caption{Quantitative evaluation of Text-to-3D for models of different sizes.}
    \vspace{-0.3cm}
    % \small
    \begin{tabular}{lccccccc}\hline\hline
    Model name & Latent length &render-FID$\downarrow$ &render-KID$(\times10^3)$$\downarrow$   & P-FID$\downarrow$ & P-KID$(\times10^3)$$\downarrow$ & CLIP(I-T)$\uparrow $  & ULIP-T$\uparrow$ \\\hline\hline
    Tiny-base   &  1024  & 12.2241    & 3.4861                      & 2.3905 & 4.1187                    & 0.2242        & 0.1321     \\
    Small-base  &  1024  & 11.2982    & 4.2074                      & 1.9332 & 4.1386                    & 0.2319        & 0.1509      \\
    Medium-base &  1024  & 13.0596    & 5.4561                      & 1.4714 & 2.7708                    & 0.2311        & 0.1511 \\
    Large-base  &  1024  & 6.5732     & 2.3617                      & 0.8650 & 1.6377                    & 0.2358        & 0.1559  \\
    XL-base     &  1024  & 5.2961     & 1.8640                      & 0.7825 & 1.3805                    & 0.2366        & 0.1554  \\ \hline
    Large-P     &  1024  & 5.7080     & 1.9997                      & 0.7148 & 1.2202                    & 0.2360        & 0.1565  \\
    XL-P        &  1024  & \tb{4.0196}& \tb{1.2773}                 & 0.6360 & 1.0761                    & 0.2371        & 0.1564 \\  \hline
    Large-P-HD  &  2048  & 5.5634     & 1.8234                      & 0.6394 & 0.9170                    & \tb{0.2374}   & \tb{0.1578} \\
    XL-P-HD     &  2048  & 4.4779     & 1.4486                      & \tb{0.5072} & \tb{0.5180}          & 0.2372        & 0.1569     \\  \hline
    \end{tabular}
    \label{table:eval_text2shape}
\end{table*}

\begin{table*}
    \centering
    \caption{Quantitative evaluation of Multi-modal-to-3D for different conditions and their combinations.}
    \vspace{-0.3cm}
    \begin{tabular}{lcccccccc}\hline\hline
        Condition  &CD$(\times10^3)$$\downarrow$ &EMD$(\times10^2)$$\downarrow$ &Voxel-IoU$\uparrow$ & F-Score$\uparrow$ &P-FID$\downarrow$ & P-KID$(\times10^3)$$\downarrow$  & 
        ULIP-T$\uparrow$ & ULIP-I$\uparrow$ \\\hline\hline
        Image       &  12.4092      & 17.6155      & 0.4513        & 0.4070      & 0.9946        & 1.9889          & 0.1329        & 0.2066\\
        MVN         &  0.9924       & 5.7283       & 0.7697        & 0.8218      & 0.3038        & 0.2420          & 0.1393        & 0.2220\\
        Voxel  &  \tb{0.5676}  & 8.4254       & 0.6273        & 0.6049      & 2.6963        & 5.0008          & 0.1186        & 0.1837\\
        Image-Bbox  &  5.4733       & 14.0811      & 0.5122        & 0.4909      & 1.5884        & 3.2994          & 0.1275        & 0.2028\\
        Image-Voxel &  0.7491       & 8.1174       & 0.6514        & 0.6541      & 2.4866        & 6.8767          & 0.1262        & 0.2017\\
        Text-Image  &  7.7198       & 14.5489      & 0.4980        & 0.4609      & 0.7996        & 1.4489          & 0.1407        & 0.2122\\
        Text-MVN     &  0.7301       & \tb{5.4034}  & \tb{0.7842}   & \tb{0.8358} & \tb{0.2184}   & \tb{0.1233}     & \tb{0.1424}   & \tb{0.2240}\\
        Text-Bbox   &  5.6421       & 14.6170      & 0.4921        & 0.4659      & 2.0074        & 4.0355          & 0.1417        & 0.1838\\
        Text-Voxel  &  0.6090       & 7.4981       & 0.6737        & 0.6689      & 1.0427        & 1.0903          & 0.1397        & 0.2036\\  \hline
    \end{tabular}
    \label{table:eval_condition}
    % \vspace{-0.2cm}
\end{table*}

\section{Results}
\label{sec:result}
We have trained five base models of different model sizes using our full training data with length of latent code $L=1024$, ranging from Tiny-base to XL-base.
Based on Large-base and XL-base, we have trained Large-P and XL-P on a high-quality subset of our training data including 300K objects, using length of latent code $L=1024$.
Based on Large-P and XL-P, we have further trained using the same subset data but with a longer length of latent code $L=2048$.
For adaptations including LoRA fine-tuning and conditioning, we have trained these modules based on XL-P using the same high-quality subset data, with each module independently trained for 8 hours.

Next, we demonstrate the generation results with various conditioning using the XL-P model of CLAY. 
Fig.~\ref{fig:gallery} illustrates a sample collection of 3D models generated by CLAY, demonstrating its versatility in producing a wide range of objects with intricate details and textures. 
From ancient tools to futuristic spacecraft, the collection traces through a fascinating human history of imagination, celebrating the fusion of art, tech, and human ingenuity as well as embracing our rich cultural heritage. The array also includes technologically advanced vehicles, cultural artifacts, everyday items, and imaginative elements, all of which highlight the model's capacity for high-fidelity and varied 3D creations suitable for applications in gaming, film, and virtual simulations. 

Fig.~\ref{fig:gallery2} showcases CLAY's conditioning capabilities across different modalities. With image conditioning, CLAY generates geometric entities that faithfully resemble the input images, be it real-world photos, AI-generated concepts, or hand-drawn sketches. CLAY also allows for the creation of entire towns or bedrooms from scattered bounding boxes. Using multi-view images, it reliably reconstructs 3D geometries from multiple perspectives and normal maps. CLAY further manages to generate from sparse point cloud, indicating it can also serve as an effective surface reconstruction tool, analogous but outperforming GCNO~\cite{Rui2023GCNO} from as few as 512 points in the ``knot'' case. Additionally, CLAY can be used to further improve 3D geometries generated by existing techniques while maintaining sharp edges and flat surfaces largely missing in prior art. Diversity wise, CLAY excels in generating rich varieties in shapes from the same voxel input, transforming the same coarse shape into anything from a futuristic monument to a Medieval Castle, from an SUV to a space shuttle, resembling our unlimited imagination. Finally, CLAY can be used to complete missing parts from partially available geometry and therefore serves as both a geometry completion tool and an editing tool. For example, it allows us to alter a monster's body or turn a companion robot into a battle-ready counterpart, a Star Wars fantasy for many.

\begin{figure}
    \centering
    \newcommand{\ct}{37}
    \begin{overpic}[width=\linewidth]{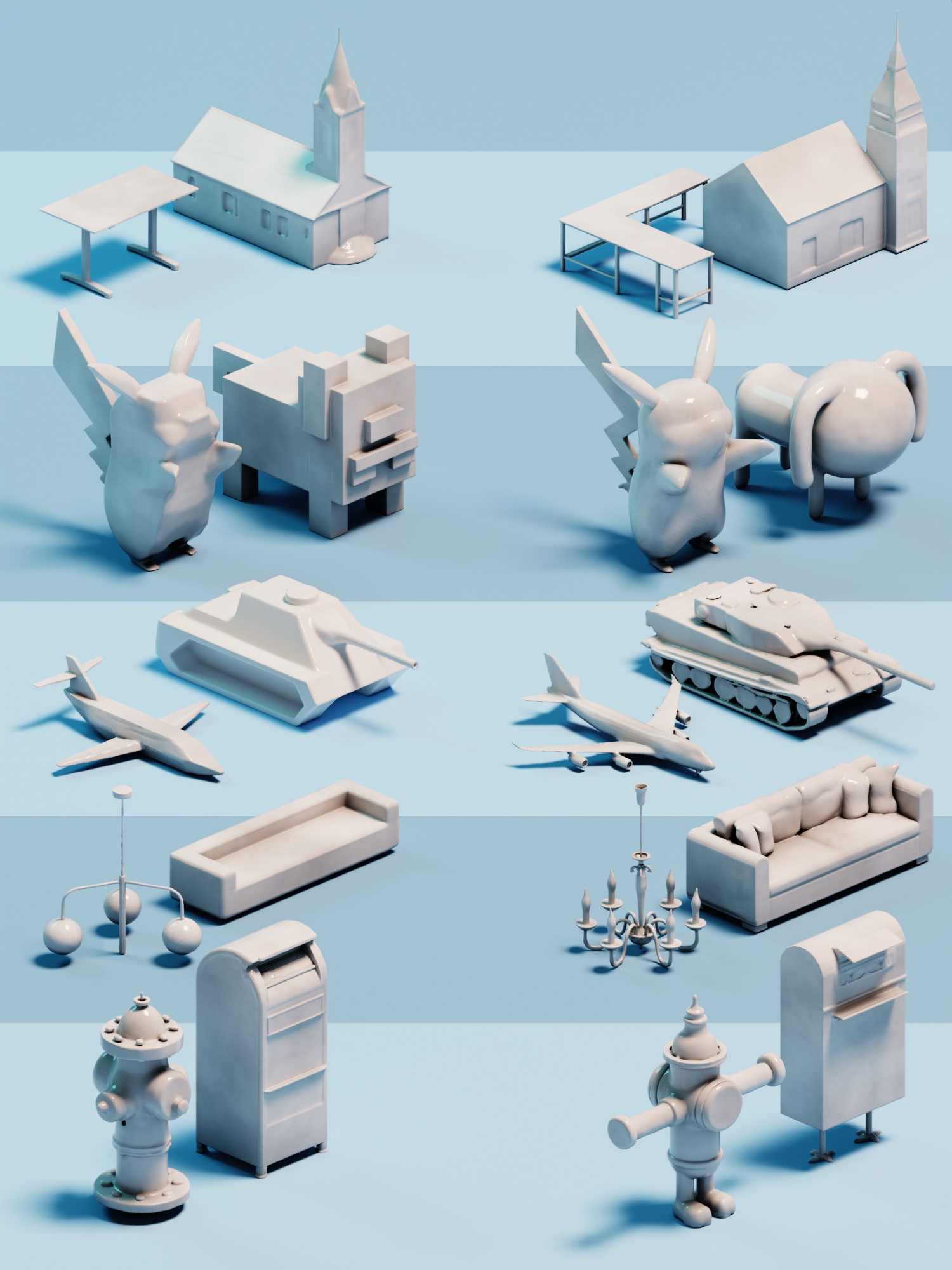}
    \put(\ct,78){\makebox(0,0){\small $\leftarrow$ symmetric}}
    \put(\ct,75){\makebox(0,0){\small asymmetric $\rightarrow$ }}
    
    \put(\ct,58){\makebox(0,0){\small $\leftarrow$ sharp}}
    \put(\ct,55){\makebox(0,0){\small smooth $\rightarrow$ }}

    \put(\ct,41){\makebox(0,0){\small $\leftarrow$ low-poly}}
    \put(\ct,38){\makebox(0,0){\small high-poly $\rightarrow$ }}

    \put(\ct,25){\makebox(0,0){\small $\leftarrow$ simple}}
    \put(\ct,22){\makebox(0,0){\small complex $\rightarrow$ }}

    \put(\ct,6){\makebox(0,0){\small $\leftarrow$ original}}
    \put(\ct,3){\makebox(0,0){\small character $\rightarrow$ }}

    \end{overpic}
    \caption{Evaluation of the CLAY's ability to alter generated content by incorporating different geometric feature tags in the prompt. We showcase precise controls over the geometry style, in the extreme case transforming a fireplug into a T-pose character.}
    \label{fig:prompt}

    \vspace{-0.2cm}
\end{figure}

\subsection{Evaluations}
We have conducted comprehensive evaluations on CLAY, focusing on various aspects including model sizes, conditioning types, prompt engineering, multi-view conditioning, and geometry diversity.

\begin{figure}
    \centering
    \begin{overpic}[width=\linewidth]{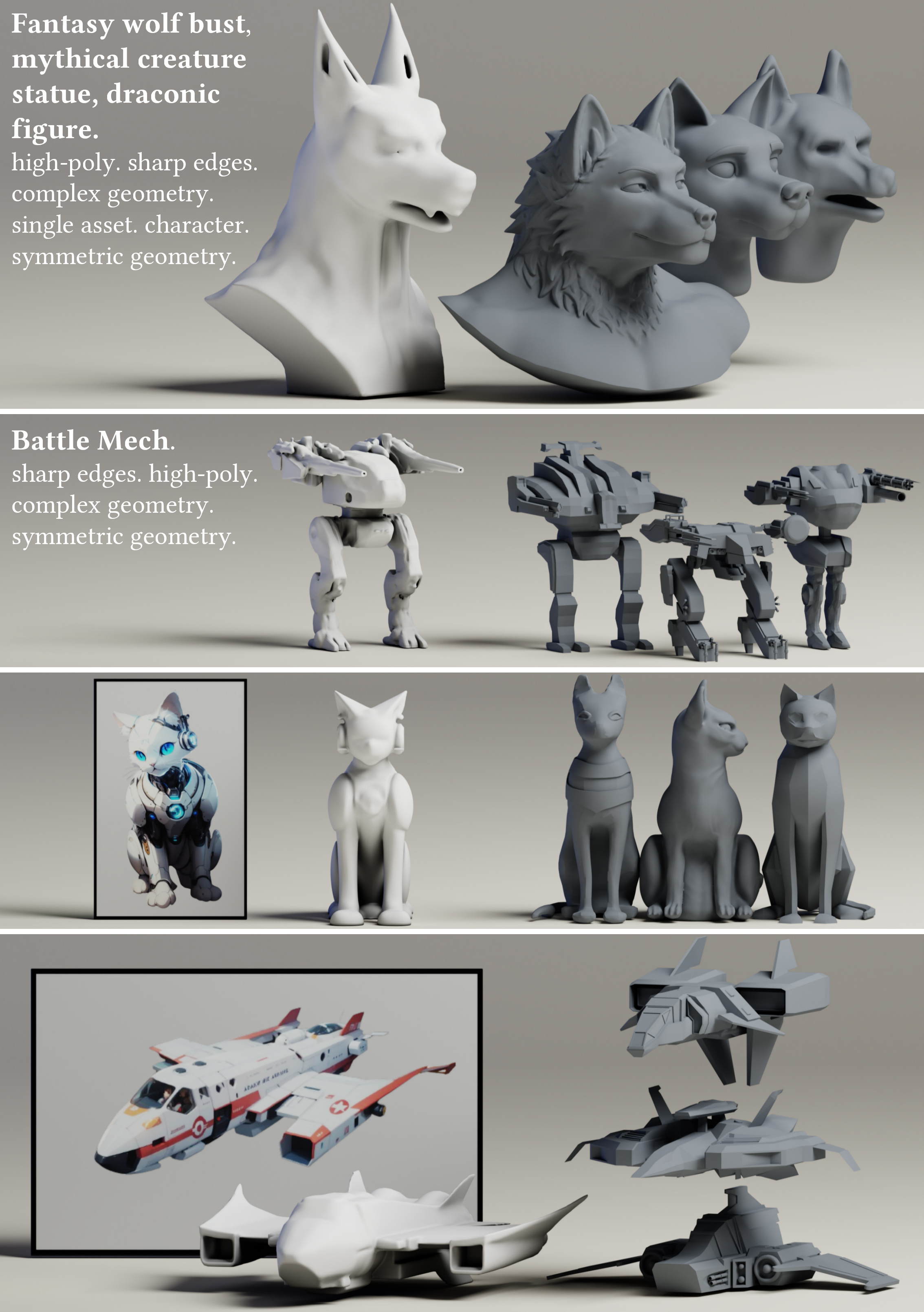}
        \put(7,-3){Input}
        \put(25,-3){CLAY}
        \put(42,-3){Nearest dataset samples}
    \end{overpic}
    \vspace{0.1cm}
    \caption{Evaluation of the geometry diversity. We present top-3 nearest samples retrieved from the dataset.
CLAY generates high-quality geometries that match the description but are distinct from the ones in the dataset.}
    \label{fig:diversity}
    
    \vspace{-0.4cm}
\end{figure}

\paragraph{Quantitative Evaluations}
Here we evaluate nine versions of CLAY as illustrated in Table.~\ref{table:eval_text2shape}.
The text-to-shape evaluation employs metrics including render-FID, render-KID, P-FID, P-KID, CLIP, and ULIP-T, using a 16K text-shape pair validation set. We apply FID and KID to both 2D (image rendering) and 3D (point cloud) feature spaces. For render-FID and render-KID, images are rendered from eight views, and PointNet++~\cite{qi2017pointnet2} is used to extract 3D features for P-FID and P-KID assessments. Additionally, we utilize CLIP-ViT-L/14~\cite{radford2021clip} for evaluating text-rendering similarity and ULIP-2~\cite{xue2023ulip} for text-shape alignment. Specifically, ULIP-T is defined as $\mathrm{\text{ULIP-T}}(T, S) = \langle \mathbf{E}_T, \mathbf{E}_S \rangle$, corresponding to the inner product of normalized ULIP features of caption $T$ and generated geometry $S$.
Table.~\ref{table:eval_text2shape} reveals the apparent trend that larger models excel over the smaller ones in text-to-shape generation tasks, demonstrated by higher scores and more accurate text-shape alignment.

We have also evaluated various conditioning modules, including image, multi-View normal, bounding box, and voxel, using XL-P as the base model. Additional metrics such as Chamfer Distance (CD), Earth Mover's Distance (EMD), Voxel-IoU, and F-score are employed to assess conditioned shape generation accuracy. We further introduce ULIP-I to evaluate alignment between the condition image and the generated shapes. Both ULIP-T and ULIP-I are assessed across all conditions, except a few, such as voxel, that do not utilize text or image inputs.
Table.~\ref{table:eval_condition} shows that with as few as a single condition, CLAY already manages to generate geometry of very high fidelity. Applying additional conditions  further improves geometric details while maintaining high alignment with the ground truth text or image at the feature level.
It is worth mentioning that among all settings, our multi-view normal (MVN) conditioning model exhibits one of the most outstanding performances. Therefore, CLAY can be also deemed as a reliable reconstruction back-end for other multi-view generation models~\cite{shi2024mvdream, long2023wonder3d}.

\begin{figure}
    \centering
    \begin{overpic}[width=\linewidth]{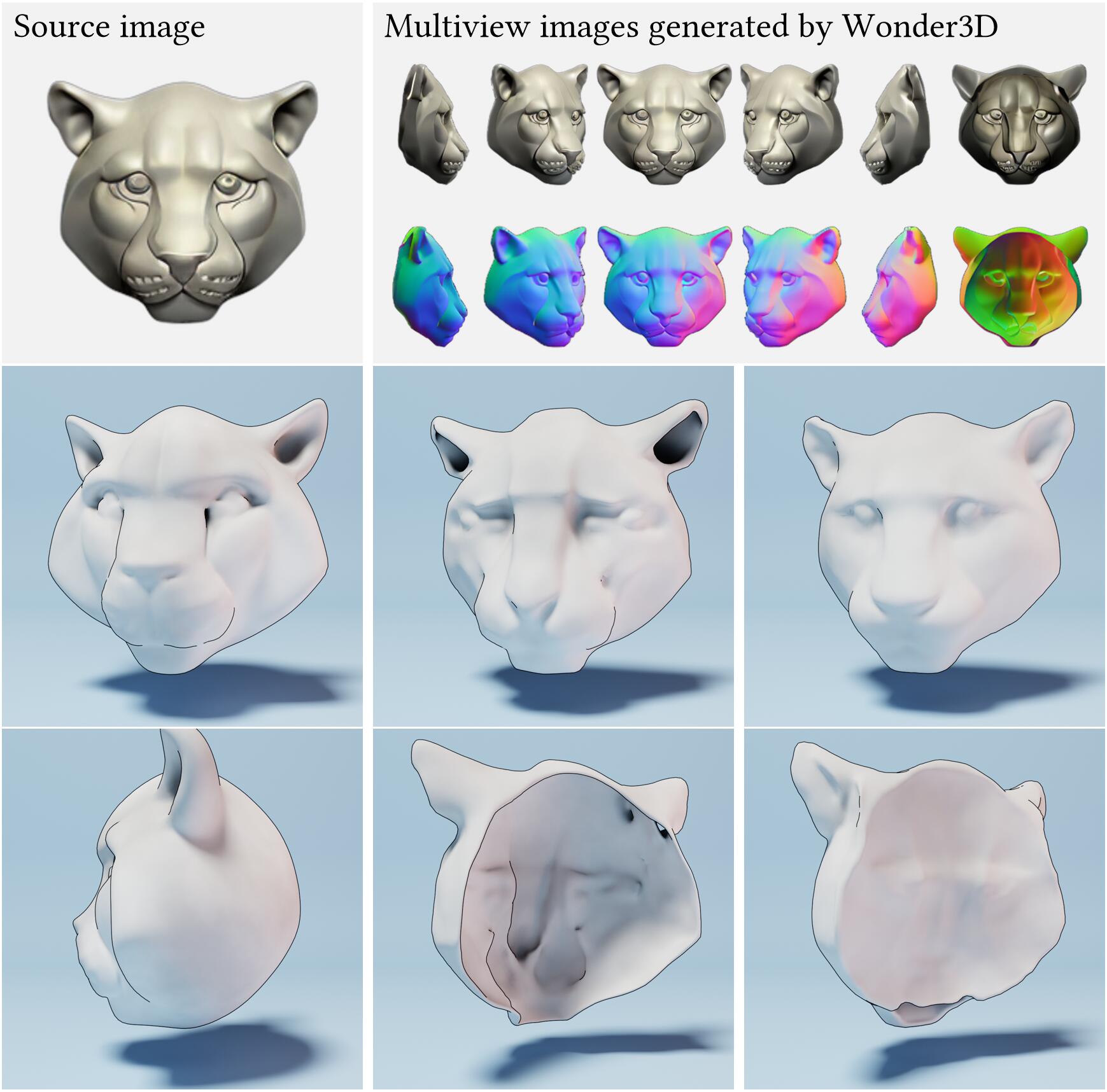}
    \put(3,-4){CLAY-single image}
    \put(41,-4){CLAY-MVN}
    \put(39.5,-8){(from 4 views)}
    \put(71,-4){Wonder3D-NeuS}
    \put(73,-8){(from 6 views)}
    \end{overpic}
    \vspace{0.1cm}
    \caption{
    Geometry generation via single image and multi-view image conditioning with multi-view RGB and normal images generated by Wonder3D.
    % By utilizing our multi-view conditioning, we generate a panther face mask that faithfully corresponds to the multi-view images produced by Wonder3D.
    % The source images are from Wonder3D.
    }
    \label{fig:mv}
    % \vspace{-0.4cm}
\end{figure}

\begin{figure}[t]
    \centering
    \small
    \begin{overpic}[width=\linewidth]{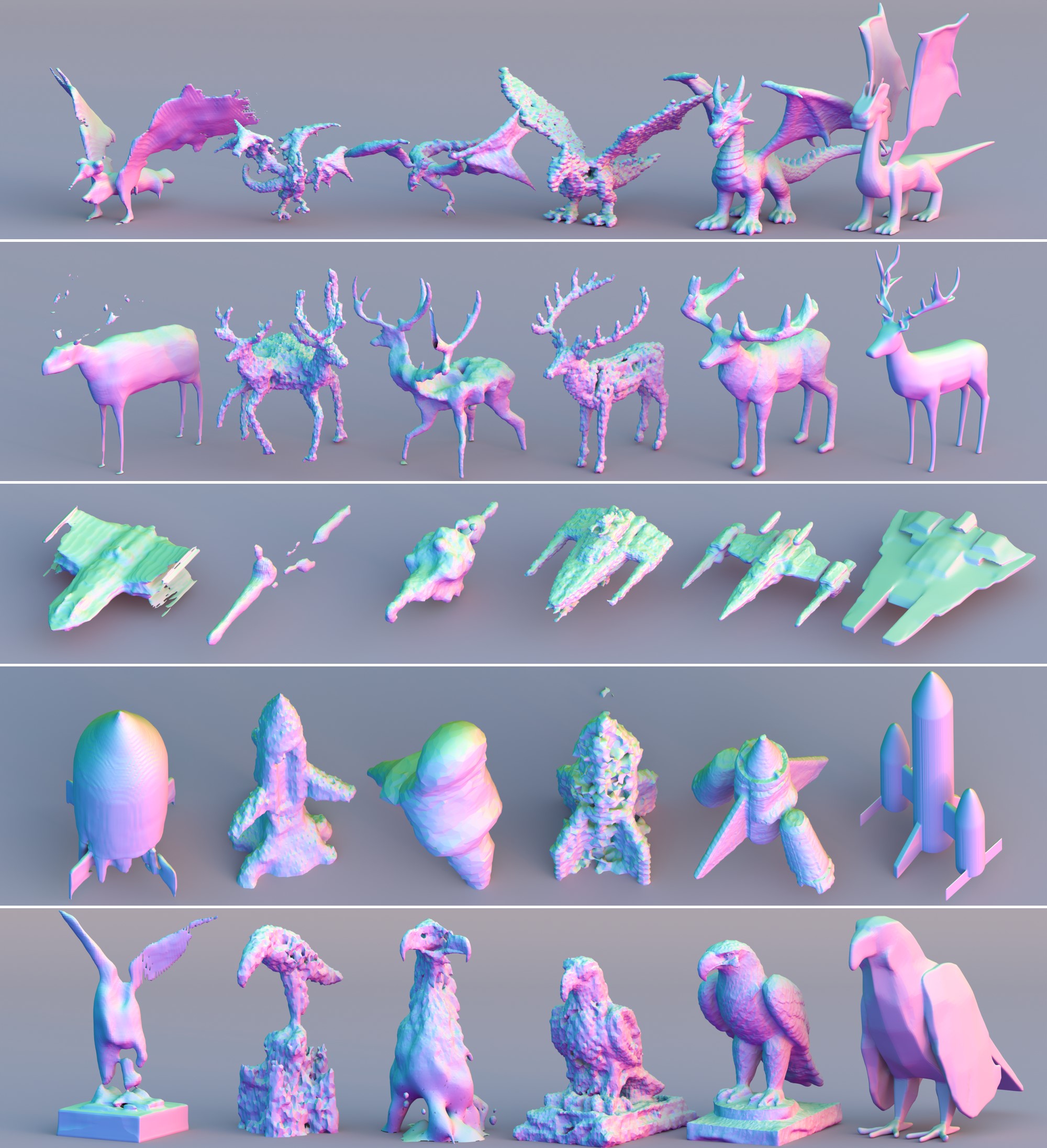}
        \put(6.5,-2.5){Shap-E}
        \put(7,-5.5){($\sim$10s)}
        \put(20.5,-2.5){Dream-}
        \put(21,-5.5){Fusion}
        \put(20.5,-8.5){($\sim$1.5h)}
        \put(33,-2.5){Magic3D}
        \put(34,-5.5){($\sim$1.5h)}
        \put(46,-2.5){MVDream}
        \put(48,-5){($\sim$1.5h)}
        \put(63.5,-2.5){Rich-}
        \put(61,-5.5){Dreamer}
        \put(63.5,-8.5){($\sim$2h)}
        \put(77.5,-2.5){CLAY}
        \put(77.5,-5.5){($\sim$45s)}
    \end{overpic}
    \vspace{0.2cm}
    \caption{Comparisons of CLAY vs. state-of-the-art methods on text-conditioned generation. From top to bottom: ``Mythical creature dragon'', ``Stag deer'', ``Interstellar warship'', ``Space rocket'', and ``Eagle, wooden statue''.  
    % Our model excels in translating input text into detailed geometries, seamlessly preserving essential geometric features such as flat surfaces and structural integrity, all achieved without the extensive time investment required for SDS optimization.
    }
    \label{fig:comp_text}
    \vspace{-0.2cm}
\end{figure}

\begin{figure*}
    \centering
    \small
    \begin{overpic}[width=\linewidth]{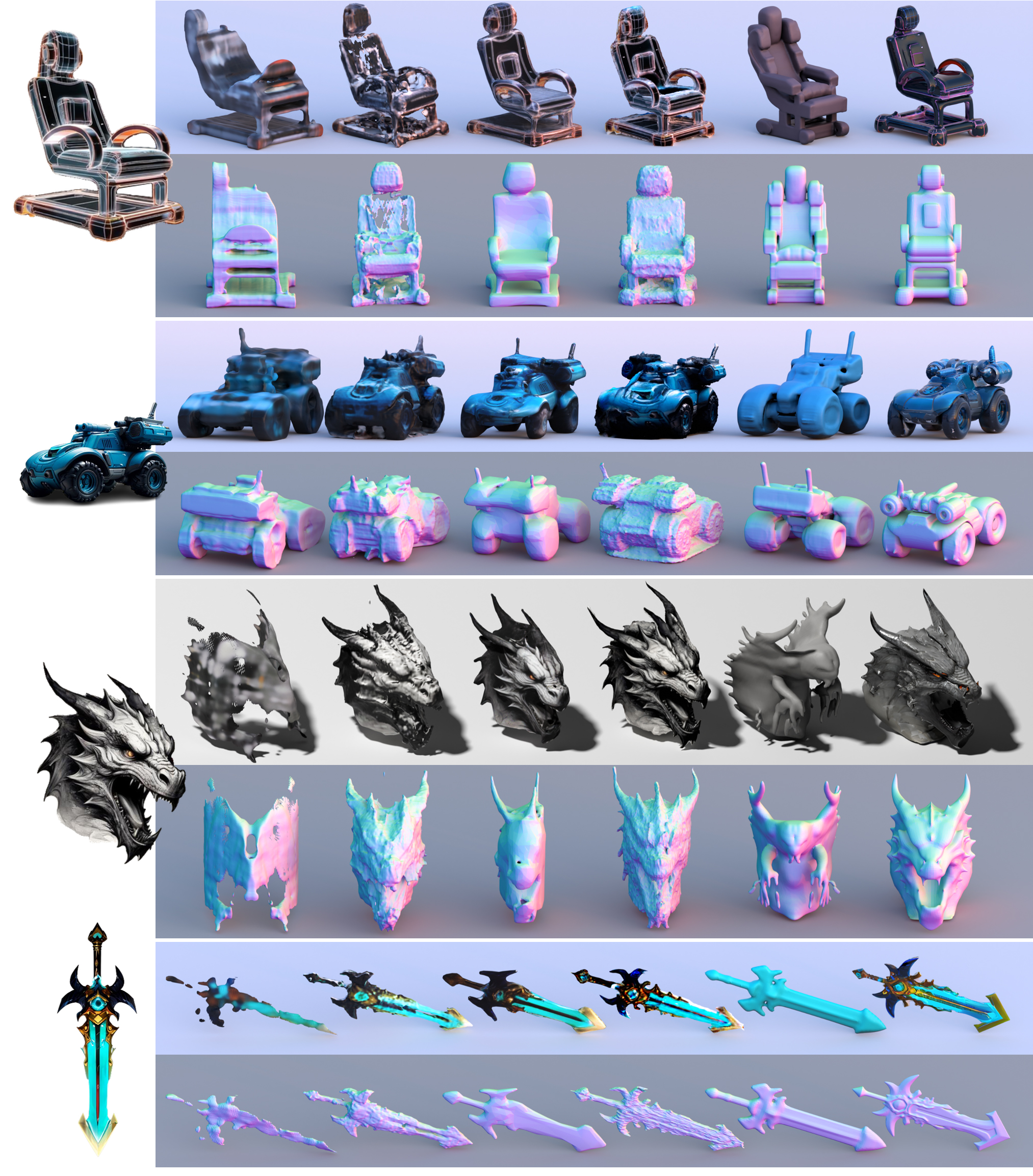}
        \put(6.75,-0.3){Input}
        \put(19,-0.3){Shap-E}
        \put(19.25,-1.8){($\sim$10s)}
        \put(30,-0.3){Wonder3D}
        \put(30.75,-1.8){($\sim$4min)}
        \put(41,-0.3){One-2-3-45++}
        \put(43,-1.8){($\sim$90s)}
        \put(52.5,-0.3){DreamCraft3D}
        \put(55,-1.8){($\sim$4h)}
        \put(64.5,-0.3){Michelangelo}
        \put(66.5,-1.8){($\sim$10s)}
        \put(79,-0.3){CLAY}
        \put(78.75,-1.8){($\sim$45s)}
    \end{overpic}
    \vspace{0.1cm}
    \caption{Comparison with state-of-the-art methods on image-conditioned generation. Even without performing optimization using the target view, CLAY still generates high-quality and detailed geometries that faithfully resemble the input image, preserving essential geometric features, including straight lines and matching surface curvatures. Note that all input images are generated by Stable Diffusion. %and SDXL. 
    Colors of Michelangelo are manually set.}
    \label{fig:comp_image}
\end{figure*}

\paragraph{Prompt engineering}
We further explore the effects of varied prompt tags on geometry generation, as illustrated in Fig.~\ref{fig:prompt}.
For example, by incorporating ``asymmetric geometry'' into our prompts, CLAY successfully generates asymmetric table and church. Similarly, the transition from ``sharp edges'' to ``smooth edges'' prompts manages to modify Pikachu and a dog into more rounded shapes. Interestingly, typical 3D models composed of high-polygon meshes such as aircrafts and tanks can be transformed into low-polygon variants using CLAY. In contrast, the ``complex geometry'' tag prompts the generation of intricate details in a chandelier and a sofa. Adding ``character'' will transform inanimate objects such as a fireplug and a mailbox into anthropomorphic figures, reminiscent to magics taught at the Hogwarts.
This experiment further indicates that specific annotated tags applied during training can effectively steer the model to produce geometries with desired complexities and styles, enhancing the quality and specificity of the generated shapes.

\paragraph{Geometry Diversity}
CLAY also excels at generating high-quality geometries with rich diversity.
In Fig.~\ref{fig:diversity}, we showcase the results generated by CLAY conditioned on either text or image inputs, alongside the most relevant samples retrieved from the dataset. To perform geometry retrieval, we utilize cosine similarity to compare the normalized ULIP feature of the generated geometry with that of geometries in the dataset.
With text inputs, CLAY manages to generate novel shapes that differ from any existing ones in the dataset. When presented with image inputs, CLAY faithfully reconstructs the content of the image while introducing novel structural combinations that are absent from the dataset. 
For instance, the airplane depicted at the bottom of Fig.~\ref{fig:diversity} represents a novel concept art piece generated by AI. It features the fuselage of a passenger airplane, uniquely merged with square air intakes and the tail fins reminiscent of a fighter jet — a design composite that is never seen in the training data. Nevertheless, CLAY accurately generates its 3D geometry, capturing a high degree of resemblance to the provided image.

\paragraph{Effectiveness of MVN Conditioning}
While single image conditioning tends to allow for more liberty in creation, multi-view conditioning harnesses multiple perspectives to deliver more detailed and precise control over the targeted generation, akin to a pixel-align sparse-view reconstruction approach.
Fig.~\ref{fig:mv} shows an example where we use an initial image of a panther's head (top left) as a starting point. This image, when processed through our single image conditioning, yields a solid 3D geometry (left column). 
In contrast, when the concept is further solidified using Wonder3D to generate multi-view images and corresponding normal maps, it results in a panther face mask with a notably thin surface (top right).
Based on these multi-view images, our multi-view images conditioning using normal maps successfully harnesses these multiple views, leading to a faithful yet efficient synthesis of the thin surface (center column), distinct from the traditional NeuS method applied to Wonder3D’s outputs (right column). 
This comparison underscores the precision and efficiency of our multi-view image conditioning in guiding the generation of detailed 3D geometries.

\begin{table}
    \vspace{0.4cm}
    \centering
    \setlength{\tabcolsep}{1.5mm}{}
    \caption{Quantitative comparison with state-of-the-art methods.}
    \vspace{-0.3cm}
    \begin{tabular}{lccccc}\hline\hline
    Method & CLIP & CLIP & ULIP-T & ULIP-I & Time \\\hline\hline
    \tb{Text-to-3D} &(N-T)& (I-T)&& \\\hline
    Shap-E      & 0.1761        & 0.2081          & 0.1160 &    /   & $\sim$10\text{s}   \\
    DreamFusion & 0.1549        & 0.1781          & 0.0566 &    /   & $\sim$1.5\text{h}   \\
    Magic3d     & 0.1553        & 0.2034          & 0.0661 &    /   & $\sim$1.5\text{h}   \\
    MVDream     & 0.1786        & 0.2237          & 0.1351 &    /   & $\sim$1.5\text{h}   \\
    RichDreamer & 0.1891        & 0.2281          & 0.1503 &    /   & $\sim$2\text{h}   \\
    CLAY        & \tb{0.1948}   & \tb{0.2324}     & \tb{0.1705}& /  & $\sim$45\text{s}   \\\hline
    \tb{Image-to-3D} & (N-I) & (I-I) & & \\\hline
    Shap-E       & 0.6315 & 0.6971 & / & 0.1307      & $\sim$10\text{s} \\
    Wonder3D     & 0.6489 & 0.7220 & / & 0.1520      & $\sim$4\text{min} \\
    DreamCraft3D & 0.6641 & 0.7718 & / & 0.1706 & $\sim$4\text{h} \\
    One-2-3-45++    & 0.6271 & 0.7574 & / & 0.1743      & $\sim$90\text{s} \\
    % Michelangelo & 0.6726 & 0.6310 & / & 0.1899      & $\sim$5 \text{s} \\
    Michelangelo & 0.6726 & / & / & 0.1899          & $\sim$10\text{s} \\
    CLAY         & \tb{0.6848} & \tb{0.7769} & / & \tb{0.2140} & $\sim$45\text{s} \\\hline
    \end{tabular}
    \label{tab:comp_other}
    \vspace{-0.4cm}
\end{table}

\paragraph{Running Time}
Regarding the inference timing breakdown, on a single Nvidia A100 GPU, it takes CLAY about 4 seconds for shape latent generation, 1 seconds to decode the latent due to the efficient adaptive sampling, 8 seconds for mesh processing, and 32 seconds for PBR generation, cumulatively resulting in a total generation time of 45 seconds.

\subsection{Comparisons with SOTA}

We compare our methods with leading text-to-3D approaches, namely Shap-E~\cite{jun2023shap}, DreamFusion~\cite{poole2022dreamfusion}, Magic3D~\cite{lin2023magic3d}, MVDream~\cite{shi2024mvdream}, and RichDreamer~\cite{qiu2023richdreamer}. We utilize the open-source code for Shap-E, MVDream, and RichDreamer, while for DreamFusion and Magic3D, we employ a third-party implementation~\cite{threestudio2023}.

\paragraph{Qualitative Comparison}
On Text-to-3D tasks, Fig.~\ref{fig:comp_text} illustrates the comparison using normal maps, with text inputs such as ``Mythical creature dragon'', ``Stag deer'', ``Interstellar warship'', ``Space rocket'', and ``Eagle, wooden statue''.
Shap-E exhibits faster generation but lacks complete geometry structures. Pure SDS optimization methods like DreamFusion and Magic3D exhibit the multi-face Janus artifacts. MVDream and RichDreamer, which generate multi-view images for SDS, produce consistent geometries but exhibit a deficiency in surface smoothness and require long optimization times. In contrast, CLAY manages to produce high-quality 3D assets in roughly 45 seconds (5 seconds for geometry and 40 seconds for texture). The generated geometries exhibit smooth surfaces without compromising intricate details, better matching the text prompts.

We have further compared the image-to-3D generation quality of between CLAY and SOTA (Shap-E~\cite{jun2023shap}, Wonder3D~\cite{long2023wonder3d}, One-2-3-45++~\cite{liu2023one}, DreamCraft3D~\cite{sun2023dreamcraft3d}, and Michelangelo~\cite{zhao2023michelangelo}). We use the official code of respective techniques except One-2-3-45++ where only its online demo is available. Our evaluations include inputs like Chair, Car, Dragon Head, and Sword, detailed in Fig.~\ref{fig:comp_image}.  Note that Michelangelo produces only geometries and we manually assign a similar color for rendering.
Shap-E, while fast, fails to accurately reconstruct the input images, resulting in incomplete geometries.
Wonder3D, which relies on multi-view images and normal prediction followed by NeuS~\cite{wang2021neus} reconstruction, produces coarse and incomplete geometries due to inconsistencies among the multi-view output. 
One-2-3-45++ is efficient in creating smooth geometries but lacked details and does not fully maintain symmetry, especially on complex objects such as Chairs and Dragons.
DreamCraft3D is an SDS optimization method that produces high-quality output, but is time-consuming and still results in uneven surfaces.
CLAY in contrast manages to quickly generate detailed and high-quality geometries along with high quality PBR textures.

\begin{figure}
    \centering
    \begin{overpic}[width=\linewidth]{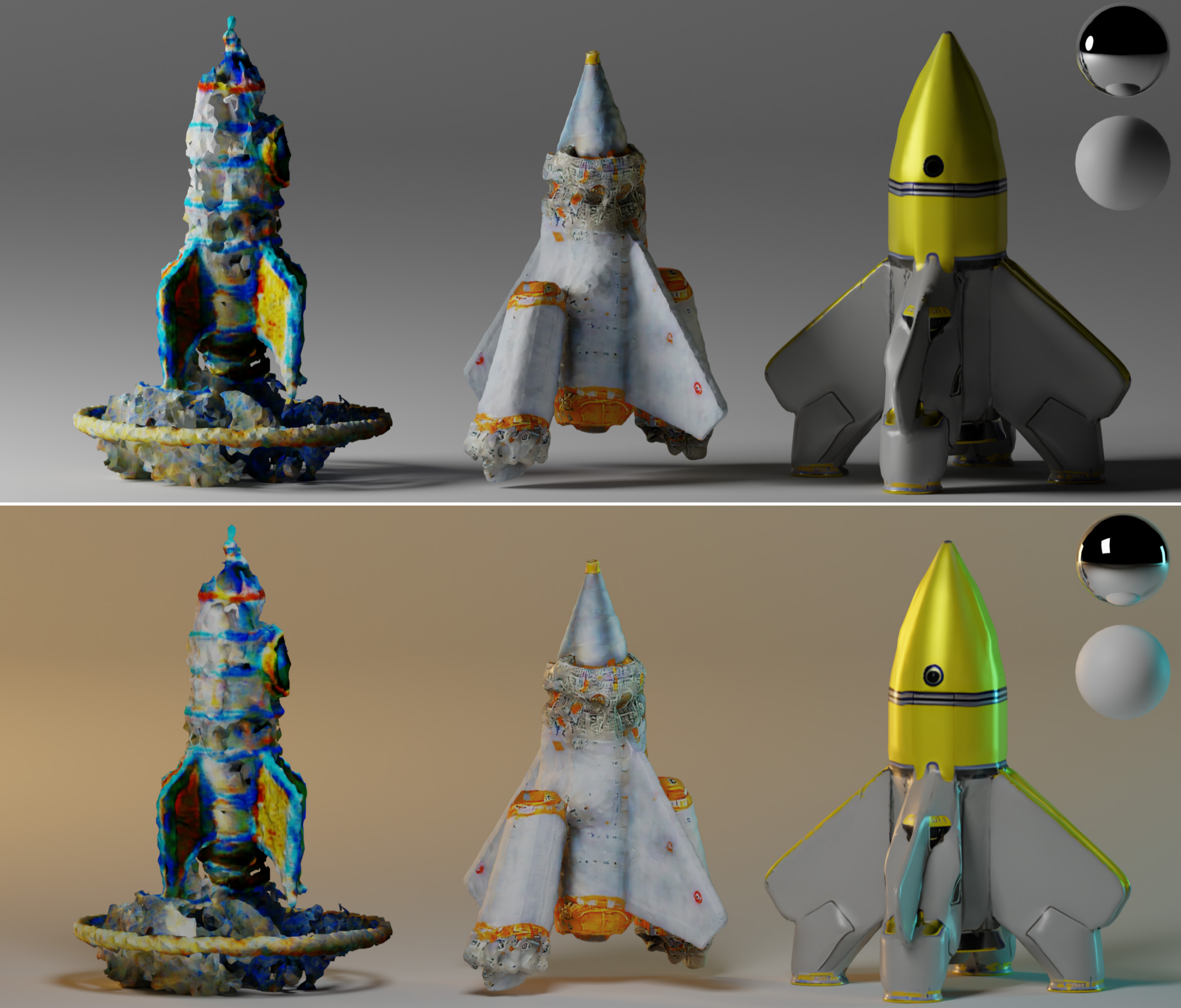}
        \put(12,-4){MVDream}
        \put(41,-4){RichDreamer}
        \put(77,-4){CLAY}
    \end{overpic}
    \vspace{-0.2cm}
    % \vspace{0.05cm}
    \caption{Comparison of rendering results under two distinct lighting conditions. The light probes are displayed at the top-right corner. Our method showcases high-quality rendering with accurate specular highlights, whereas MVDream lacks matching highlights and RichDreamer misses view dependency by modeling highlights as fixed surface textures.}
    \label{fig:pbr}
    \vspace{-0.4cm}
\end{figure}

\paragraph{Quantitative Comparisons}
We perform additional quantitative comparison using a GPT-4 generated test dataset that includes 50 images and 50 text prompts tailored for text-to-3D and image-to-3D evaluations, respectively. In addition to from ULIP-T and ULIP-I, we render 30 views of RGB images and normal maps for each generated 3D asset, respectively. We apply four CLIP-based metrics to these views, calculating the average to provide a comprehensive assessment.
 CLIP(N-I) and CLIP(N-T) gauge the geometric alignment of the normal map with the input image and text, respectively whereas CLIP(I-I) and CLIP(I-T) evaluate the appearance by measuring the similarity of rendered images with the input images and text.
As shown in Table.~\ref{tab:comp_other}, CLAY outperforms SOTA techniques in all metrics.

\paragraph{PBR Material Comparison}

Another key component in CLAY is material generation. Here we show visual comparisons between CLAY and two leading methods, MVDream~\cite{shi2024mvdream} and RichDreamer~\cite{qiu2023richdreamer}, using the text prompt ``Space rocket''. Fig.~\ref{fig:pbr} illustrates that, under varying lighting conditions MVDream without PBR materials cannot fully reproduce specular highlights. RichDreamer, employing an albedo diffusion model, attempts to distinguish the albedo from complex lighting effects. In this case though, the highlights are modeled as fixed surface textures under changing environment lighting, e.g., on the rocket's head. In contrast, CLAY faithfully models PBR materials where the rocket's metallic surfaces exhibit realistic highlights that moves consistently with the moving environment lighting. This also showcases the potential advantages of separating generating geometry and texture.

\begin{figure}[t]
    \vspace{0.1cm}
    \centering
    \includegraphics[width=\linewidth]{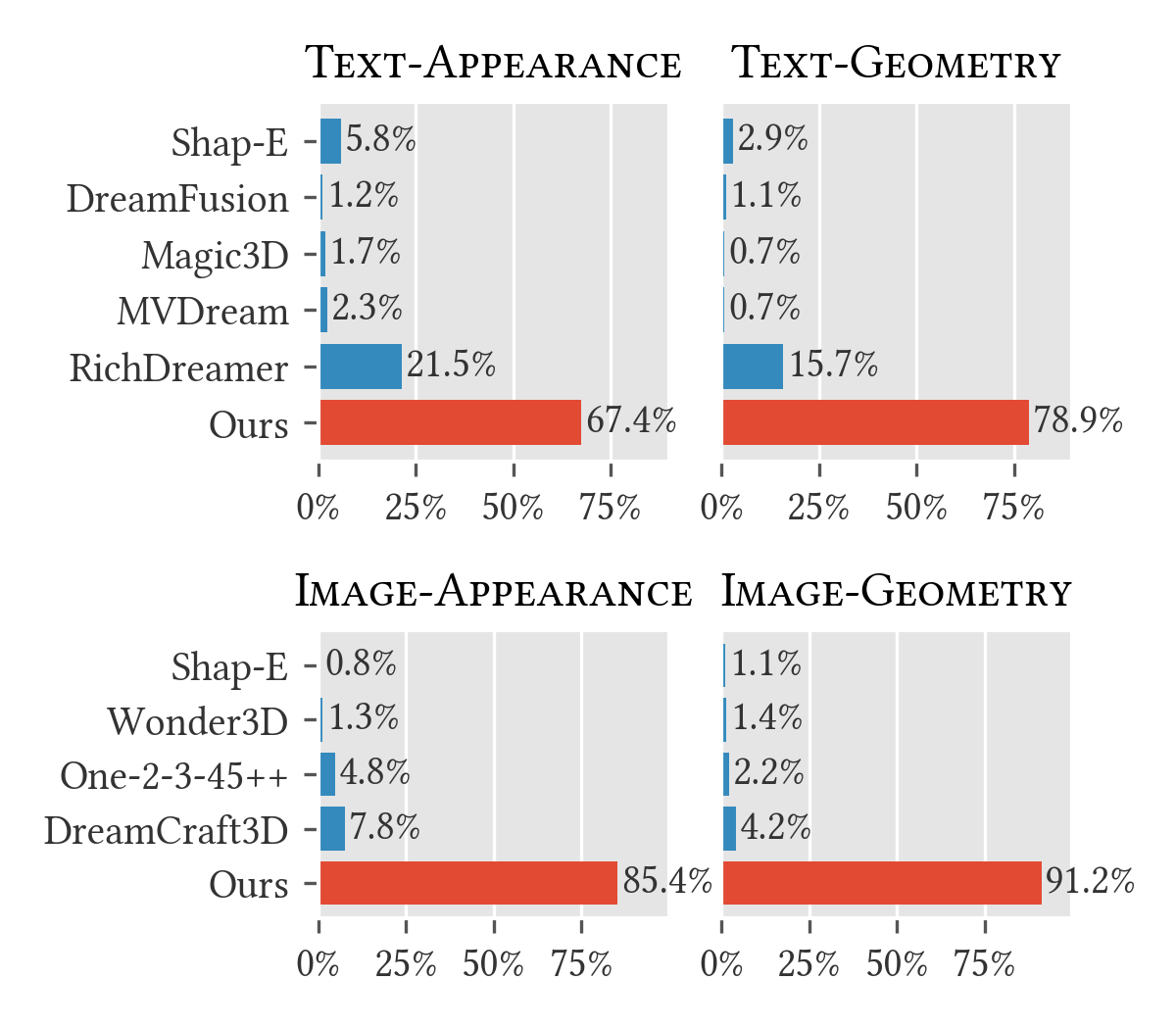}
    \vspace{-0.8cm}
    \caption{User studies of CLAY vs. state-of-the-art methods indicates strong preferences of CLAY in generating both geometry and appearance.}
    \label{fig:user_study}
    \vspace{-0.2cm}
\end{figure}

\paragraph{User studies}

We have conduct a comprehensive user study, structured around two primary evaluations: appearance quality for visualization and geometry quality for modeling.
We have created a test set consisting of 5 text prompts generated by GPT-4 and 15 images generated by Stable Diffusion. % and SDXL. 
A total of 150 volunteers participated in the study, each evaluating 15 randomly chosen questions to determine their preferred method.
We compare CLAY with leading approaches on Text-to-3D and Image-to-3D tasks respectively.
Fig~\ref{fig:user_study} shows that CLAY outperforms others in both appearance and geometry in text-to-3D and image-to-3D tasks. Specifically, CLAY secured 67.4\% of votes for appearance and 78.9\% for geometry in text-to-3D, surpassing the second-ranked RichDreamer, which had a notably longer optimization time of $\sim$2 hours compared to our $\sim$45 seconds. In Image-to-3D, CLAY further garnered 85.4\% and 91.2\% votes in appearance and geometry, respectively.

\section{Discussions and Conclusions}

We have presented CLAY, a large-scale 3D generative model that supports multi-modal controls for high quality 3D asset generation, further bridging the gap between the vivid realms of human imagination and the tangible world of digital creation. By enabling users to effortlessly craft and manipulate digital geometry and textures, CLAY empowers both experts and novices alike to facilitate the seamless transformation of abstract concepts into detailed and realistic 3D models, expanding the horizons of digital artistry and design. At CLAY's core is a large-scale generative framework enabled by a multi-resolution VAE and a DiT to accurately depict continuous surfaces and complex shapes. We have shown how to scale up CLAY efficiently through a progressive training scheme to become a large 3D generative model. Its success is also largely attributed to our elaborately designed geometric data processing pipeline, including a standardized geometry remeshing protocol to ensure consistency in training, and the automatic annotation capabilities by GPT-4V. Comprehensive experimental evaluations and user studies have demonstrated CLAY's efficacy and adaptability. Its high geometry quality, diversity in variety, and material richness position CLAY as one of the leading 3D generator in the field.

\paragraph{Ethics Statement} Same as 2D contents, 3D generation models have the potential to producing deceptive contents. Although we have implemented rigorous scrutiny processes for our training data, the utilization of pretrained feature encoders (CLIP~\cite{radford2021clip} for text encoding and DINO~\cite{oquab2023dinov2} for image encoding) in CLAY introduces a high-level of generalization capability that carries the risk of potential misuse. This means there is a possibility that our model could be used to generate virtual assets or scenes that violate regulations and propagate false information. We are committed to addressing these ethical issues, and along with the whole community, developing strategies to ensure the responsible use of CLAY.

\paragraph{Limitations and Future Work} It is important to note that CLAY is not yet complete end-to-end, as it entails distinct stages for generating geometry and materials, and requires additional steps such as remeshing and UV unwrapping.
An immediate future step is to explore integrated model architectures to integrate geometry and PBR materials. This will require implementing automatic schemes to produce geometry with consistent topology. By far, CLAY has been trained on a substantially large dataset. However, there is still room for improvement in terms of both the quantity and quality of the training data, especially compared with 2D image datasets used to train Stable Diffusion. 
Further, we observe that CLAY shows robustness in generating assets composed of single objects but tends to be vulnerable when dealing with complex ``composed objects'', such as ``a tiger riding a motorcycle'', particularly with text-only inputs. The issue is largely attributed to insufficient training data of composed objects and the lack of detailed textual descriptions of these objects.
The issue can potentially be mitigated through a text-to-image-to-3D workflow, akin to the approaches employed by Wonder3D~\cite{long2023wonder3d} and One-2-3-45++~\cite{liu2023one2345}.
As the community augments the training dataset with a larger and more diverse collection of 3D shapes along with corresponding text descriptions, we expect CLAY as well as its concurrent works to reach a new level of geometry generation, in both quality and complexity.
Finally, we intend to explore extends of CLAY to dynamic object generation. The generated results from CLAY indicate that it may be possible to semantically partition the geometry into meaningful parts, further facilitating motion and interaction, as in 
\citet{singer2023text4d} and \citet{ling2023align}.

\bibliographystyle{ACM-Reference-Format}
\bibliography{sample-bibliography}

%%% -*-BibTeX-*-
%%% Do NOT edit. File created by BibTeX with style
%%% ACM-Reference-Format-Journals [18-Jan-2012].

\begin{thebibliography}{98}

%%% ====================================================================
%%% NOTE TO THE USER: you can override these defaults by providing
%%% customized versions of any of these macros before the \bibliography
%%% command.  Each of them MUST provide its own final punctuation,
%%% except for \shownote{}, \showDOI{}, and \showURL{}.  The latter two
%%% do not use final punctuation, in order to avoid confusing it with
%%% the Web address.
%%%
%%% To suppress output of a particular field, define its macro to expand
%%% to an empty string, or better, \unskip, like this:
%%%
%%% \newcommand{\showDOI}[1]{\unskip}   % LaTeX syntax
%%%
%%% \def \showDOI #1{\unskip}           % plain TeX syntax
%%%
%%% ====================================================================

\ifx \showCODEN    \undefined \def \showCODEN     #1{\unskip}     \fi
\ifx \showDOI      \undefined \def \showDOI       #1{#1}\fi
\ifx \showISBNx    \undefined \def \showISBNx     #1{\unskip}     \fi
\ifx \showISBNxiii \undefined \def \showISBNxiii  #1{\unskip}     \fi
\ifx \showISSN     \undefined \def \showISSN      #1{\unskip}     \fi
\ifx \showLCCN     \undefined \def \showLCCN      #1{\unskip}     \fi
\ifx \shownote     \undefined \def \shownote      #1{#1}          \fi
\ifx \showarticletitle \undefined \def \showarticletitle #1{#1}   \fi
\ifx \showURL      \undefined \def \showURL       {\relax}        \fi
% The following commands are used for tagged output and should be
% invisible to TeX
\providecommand\bibfield[2]{#2}
\providecommand\bibinfo[2]{#2}
\providecommand\natexlab[1]{#1}
\providecommand\showeprint[2][]{arXiv:#2}

\bibitem[Bar-Tal et~al\mbox{.}(2023)]%
        {bar2023multidiffusion}
\bibfield{author}{\bibinfo{person}{Omer Bar-Tal}, \bibinfo{person}{Lior Yariv},
  \bibinfo{person}{Yaron Lipman}, {and} \bibinfo{person}{Tali Dekel}.}
  \bibinfo{year}{2023}\natexlab{}.
\newblock \bibinfo{title}{MultiDiffusion: fusing diffusion paths for controlled
  image generation}.
\newblock , \bibinfo{numpages}{16}~pages.
\newblock


\bibitem[Blattmann et~al\mbox{.}(2023)]%
        {blattmann2023stable}
\bibfield{author}{\bibinfo{person}{Andreas Blattmann}, \bibinfo{person}{Tim
  Dockhorn}, \bibinfo{person}{Sumith Kulal}, \bibinfo{person}{Daniel
  Mendelevitch}, \bibinfo{person}{Maciej Kilian}, \bibinfo{person}{Dominik
  Lorenz}, \bibinfo{person}{Yam Levi}, \bibinfo{person}{Zion English},
  \bibinfo{person}{Vikram Voleti}, \bibinfo{person}{Adam Letts},
  \bibinfo{person}{Varun Jampani}, {and} \bibinfo{person}{Robin Rombach}.}
  \bibinfo{year}{2023}\natexlab{}.
\newblock \bibinfo{title}{Stable Video Diffusion: Scaling Latent Video
  Diffusion Models to Large Datasets}.
\newblock
\newblock
\showeprint[arxiv]{2311.15127}~[cs.CV]


\bibitem[{Blender Online Community}(2024)]%
        {Blender}
\bibfield{author}{\bibinfo{person}{{Blender Online Community}}.}
  \bibinfo{year}{2024}\natexlab{}.
\newblock \bibinfo{title}{Blender - a 3D modelling and rendering package}.
\newblock \bibinfo{howpublished}{\url{http://www.blender.org}}.
\newblock


\bibitem[Chang et~al\mbox{.}(2015)]%
        {chang2015shapenet}
\bibfield{author}{\bibinfo{person}{Angel~X. Chang}, \bibinfo{person}{Thomas
  Funkhouser}, \bibinfo{person}{Leonidas Guibas}, \bibinfo{person}{Pat
  Hanrahan}, \bibinfo{person}{Qixing Huang}, \bibinfo{person}{Zimo Li},
  \bibinfo{person}{Silvio Savarese}, \bibinfo{person}{Manolis Savva},
  \bibinfo{person}{Shuran Song}, \bibinfo{person}{Hao Su},
  \bibinfo{person}{Jianxiong Xiao}, \bibinfo{person}{Li Yi}, {and}
  \bibinfo{person}{Fisher Yu}.} \bibinfo{year}{2015}\natexlab{}.
\newblock \bibinfo{title}{ShapeNet: An Information-Rich 3D Model Repository}.
\newblock
\newblock
\showeprint[arxiv]{1512.03012}~[cs.GR]


\bibitem[Chen et~al\mbox{.}(2023b)]%
        {chen2023text2tex}
\bibfield{author}{\bibinfo{person}{Dave~Zhenyu Chen}, \bibinfo{person}{Yawar
  Siddiqui}, \bibinfo{person}{Hsin-Ying Lee}, \bibinfo{person}{Sergey
  Tulyakov}, {and} \bibinfo{person}{Matthias Nießner}.}
  \bibinfo{year}{2023}\natexlab{b}.
\newblock \showarticletitle{Text2Tex: Text-driven Texture Synthesis via
  Diffusion Models}. In \bibinfo{booktitle}{\emph{2023 IEEE/CVF International
  Conference on Computer Vision (ICCV)}}. \bibinfo{pages}{18512--18522}.
\newblock
\urldef\tempurl%
\url{https://doi.org/10.1109/ICCV51070.2023.01701}
\showDOI{\tempurl}


\bibitem[Chen et~al\mbox{.}(2023a)]%
        {chen2023fantasia3d}
\bibfield{author}{\bibinfo{person}{R. Chen}, \bibinfo{person}{Y. Chen},
  \bibinfo{person}{N. Jiao}, {and} \bibinfo{person}{K. Jia}.}
  \bibinfo{year}{2023}\natexlab{a}.
\newblock \showarticletitle{Fantasia3D: Disentangling Geometry and Appearance
  for High-quality Text-to-3D Content Creation}. In
  \bibinfo{booktitle}{\emph{2023 IEEE/CVF International Conference on Computer
  Vision (ICCV)}}. \bibinfo{publisher}{IEEE Computer Society},
  \bibinfo{address}{Los Alamitos, CA, USA}, \bibinfo{pages}{22189--22199}.
\newblock
\urldef\tempurl%
\url{https://doi.org/10.1109/ICCV51070.2023.02033}
\showDOI{\tempurl}


\bibitem[Chen et~al\mbox{.}(2024)]%
        {GSGEN}
\bibfield{author}{\bibinfo{person}{Zilong Chen}, \bibinfo{person}{Feng Wang},
  {and} \bibinfo{person}{Huaping Liu}.} \bibinfo{year}{2024}\natexlab{}.
\newblock \showarticletitle{Text-to-3D using Gaussian Splatting}. In
  \bibinfo{booktitle}{\emph{Proceedings of the IEEE/CVF conference on computer
  vision and pattern recognition}}.
\newblock


\bibitem[Chen and Zhang(2019)]%
        {chen2019learning}
\bibfield{author}{\bibinfo{person}{Zhiqin Chen} {and} \bibinfo{person}{Hao
  Zhang}.} \bibinfo{year}{2019}\natexlab{}.
\newblock \showarticletitle{Learning implicit fields for generative shape
  modeling}. In \bibinfo{booktitle}{\emph{Proceedings of the IEEE/CVF
  conference on computer vision and pattern recognition}}.
  \bibinfo{pages}{5939--5948}.
\newblock


\bibitem[Cheng et~al\mbox{.}(2023)]%
        {cheng2023sdfusion}
\bibfield{author}{\bibinfo{person}{Y. Cheng}, \bibinfo{person}{H. Lee},
  \bibinfo{person}{S. Tulyakov}, \bibinfo{person}{A. Schwing}, {and}
  \bibinfo{person}{L. Gui}.} \bibinfo{year}{2023}\natexlab{}.
\newblock \showarticletitle{SDFusion: Multimodal 3D Shape Completion,
  Reconstruction, and Generation}. In \bibinfo{booktitle}{\emph{2023 IEEE/CVF
  Conference on Computer Vision and Pattern Recognition (CVPR)}}.
  \bibinfo{publisher}{IEEE Computer Society}, \bibinfo{address}{Los Alamitos,
  CA, USA}, \bibinfo{pages}{4456--4465}.
\newblock
\urldef\tempurl%
\url{https://doi.org/10.1109/CVPR52729.2023.00433}
\showDOI{\tempurl}


\bibitem[Choy et~al\mbox{.}(2016)]%
        {choy20163d}
\bibfield{author}{\bibinfo{person}{Christopher~B Choy}, \bibinfo{person}{Danfei
  Xu}, \bibinfo{person}{JunYoung Gwak}, \bibinfo{person}{Kevin Chen}, {and}
  \bibinfo{person}{Silvio Savarese}.} \bibinfo{year}{2016}\natexlab{}.
\newblock \showarticletitle{3d-r2n2: A unified approach for single and
  multi-view 3d object reconstruction}. In \bibinfo{booktitle}{\emph{Computer
  Vision--ECCV 2016: 14th European Conference, Amsterdam, The Netherlands,
  October 11-14, 2016, Proceedings, Part VIII 14}}. Springer,
  \bibinfo{pages}{628--644}.
\newblock


\bibitem[Deitke et~al\mbox{.}(2023)]%
        {deitke2023objaverse}
\bibfield{author}{\bibinfo{person}{M. Deitke}, \bibinfo{person}{D. Schwenk},
  \bibinfo{person}{J. Salvador}, \bibinfo{person}{L. Weihs},
  \bibinfo{person}{O. Michel}, \bibinfo{person}{E. VanderBilt},
  \bibinfo{person}{L. Schmidt}, \bibinfo{person}{K. Ehsanit},
  \bibinfo{person}{A. Kembhavi}, {and} \bibinfo{person}{A. Farhadi}.}
  \bibinfo{year}{2023}\natexlab{}.
\newblock \showarticletitle{Objaverse: A Universe of Annotated 3D Objects}. In
  \bibinfo{booktitle}{\emph{2023 IEEE/CVF Conference on Computer Vision and
  Pattern Recognition (CVPR)}}. \bibinfo{publisher}{IEEE Computer Society},
  \bibinfo{address}{Los Alamitos, CA, USA}, \bibinfo{pages}{13142--13153}.
\newblock
\urldef\tempurl%
\url{https://doi.org/10.1109/CVPR52729.2023.01263}
\showDOI{\tempurl}


\bibitem[Fan et~al\mbox{.}(2017)]%
        {fan2017point}
\bibfield{author}{\bibinfo{person}{Haoqiang Fan}, \bibinfo{person}{Hao Su},
  {and} \bibinfo{person}{Leonidas~J Guibas}.} \bibinfo{year}{2017}\natexlab{}.
\newblock \showarticletitle{A point set generation network for 3d object
  reconstruction from a single image}. In \bibinfo{booktitle}{\emph{Proceedings
  of the IEEE conference on computer vision and pattern recognition}}.
  \bibinfo{pages}{605--613}.
\newblock


\bibitem[Gesmundo and Maile(2023)]%
        {gesmundo2023composable}
\bibfield{author}{\bibinfo{person}{Andrea Gesmundo} {and}
  \bibinfo{person}{Kaitlin Maile}.} \bibinfo{year}{2023}\natexlab{}.
\newblock \bibinfo{title}{Composable Function-preserving Expansions for
  Transformer Architectures}.
\newblock
\newblock
\showeprint[arxiv]{2308.06103}~[cs.LG]


\bibitem[Groueix et~al\mbox{.}(2018)]%
        {groueix2018papier}
\bibfield{author}{\bibinfo{person}{Thibault Groueix}, \bibinfo{person}{Matthew
  Fisher}, \bibinfo{person}{Vladimir~G Kim}, \bibinfo{person}{Bryan~C Russell},
  {and} \bibinfo{person}{Mathieu Aubry}.} \bibinfo{year}{2018}\natexlab{}.
\newblock \showarticletitle{A papier-m{\^a}ch{\'e} approach to learning 3d
  surface generation}. In \bibinfo{booktitle}{\emph{Proceedings of the IEEE
  conference on computer vision and pattern recognition}}.
  \bibinfo{pages}{216--224}.
\newblock


\bibitem[Guo et~al\mbox{.}(2023)]%
        {threestudio2023}
\bibfield{author}{\bibinfo{person}{Yuan-Chen Guo}, \bibinfo{person}{Ying-Tian
  Liu}, \bibinfo{person}{Ruizhi Shao}, \bibinfo{person}{Christian Laforte},
  \bibinfo{person}{Vikram Voleti}, \bibinfo{person}{Guan Luo},
  \bibinfo{person}{Chia-Hao Chen}, \bibinfo{person}{Zi-Xin Zou},
  \bibinfo{person}{Chen Wang}, \bibinfo{person}{Yan-Pei Cao}, {and}
  \bibinfo{person}{Song-Hai Zhang}.} \bibinfo{year}{2023}\natexlab{}.
\newblock \bibinfo{title}{threestudio: A unified framework for 3D content
  generation}.
\newblock
  \bibinfo{howpublished}{\url{https://github.com/threestudio-project/threestudio}}.
\newblock


\bibitem[Gupta et~al\mbox{.}(2023)]%
        {gupta20233dgen}
\bibfield{author}{\bibinfo{person}{Anchit Gupta}, \bibinfo{person}{Wenhan
  Xiong}, \bibinfo{person}{Yixin Nie}, \bibinfo{person}{Ian Jones}, {and}
  \bibinfo{person}{Barlas Oğuz}.} \bibinfo{year}{2023}\natexlab{}.
\newblock \bibinfo{title}{3DGen: Triplane Latent Diffusion for Textured Mesh
  Generation}.
\newblock
\newblock
\showeprint[arxiv]{2303.05371}~[cs.CV]


\bibitem[Ho et~al\mbox{.}(2020)]%
        {ddpm}
\bibfield{author}{\bibinfo{person}{Jonathan Ho}, \bibinfo{person}{Ajay Jain},
  {and} \bibinfo{person}{Pieter Abbeel}.} \bibinfo{year}{2020}\natexlab{}.
\newblock \showarticletitle{Denoising Diffusion Probabilistic Models}. In
  \bibinfo{booktitle}{\emph{Advances in Neural Information Processing
  Systems}}, \bibfield{editor}{\bibinfo{person}{H.~Larochelle},
  \bibinfo{person}{M.~Ranzato}, \bibinfo{person}{R.~Hadsell},
  \bibinfo{person}{M.F. Balcan}, {and} \bibinfo{person}{H.~Lin}} (Eds.),
  Vol.~\bibinfo{volume}{33}. \bibinfo{publisher}{Curran Associates, Inc.},
  \bibinfo{pages}{6840--6851}.
\newblock
\urldef\tempurl%
\url{https://proceedings.neurips.cc/paper_files/paper/2020/file/4c5bcfec8584af0d967f1ab10179ca4b-Paper.pdf}
\showURL{%
\tempurl}


\bibitem[Hong et~al\mbox{.}(2024)]%
        {hong2023lrm}
\bibfield{author}{\bibinfo{person}{Yicong Hong}, \bibinfo{person}{Kai Zhang},
  \bibinfo{person}{Jiuxiang Gu}, \bibinfo{person}{Sai Bi},
  \bibinfo{person}{Yang Zhou}, \bibinfo{person}{Difan Liu},
  \bibinfo{person}{Feng Liu}, \bibinfo{person}{Kalyan Sunkavalli},
  \bibinfo{person}{Trung Bui}, {and} \bibinfo{person}{Hao Tan}.}
  \bibinfo{year}{2024}\natexlab{}.
\newblock \showarticletitle{{LRM}: Large Reconstruction Model for Single Image
  to 3D}. In \bibinfo{booktitle}{\emph{The Twelfth International Conference on
  Learning Representations}}.
\newblock


\bibitem[Hu et~al\mbox{.}(2022)]%
        {hu2022lora}
\bibfield{author}{\bibinfo{person}{Edward~J Hu}, \bibinfo{person}{Yelong Shen},
  \bibinfo{person}{Phillip Wallis}, \bibinfo{person}{Zeyuan Allen-Zhu},
  \bibinfo{person}{Yuanzhi Li}, \bibinfo{person}{Shean Wang},
  \bibinfo{person}{Lu Wang}, {and} \bibinfo{person}{Weizhu Chen}.}
  \bibinfo{year}{2022}\natexlab{}.
\newblock \showarticletitle{Lo{RA}: Low-Rank Adaptation of Large Language
  Models}. In \bibinfo{booktitle}{\emph{International Conference on Learning
  Representations}}.
\newblock


\bibitem[Huang et~al\mbox{.}(2018a)]%
        {huang2018robust}
\bibfield{author}{\bibinfo{person}{Jingwei Huang}, \bibinfo{person}{Hao Su},
  {and} \bibinfo{person}{Leonidas~J. Guibas}.}
  \bibinfo{year}{2018}\natexlab{a}.
\newblock \bibinfo{title}{Robust Watertight Manifold Surface Generation Method
  for ShapeNet Models}.
\newblock
\newblock
\showeprint[arXiv]{1802.01698}
\urldef\tempurl%
\url{http://arxiv.org/abs/1802.01698}
\showURL{%
\tempurl}


\bibitem[Huang et~al\mbox{.}(2020)]%
        {huang2020manifoldplus}
\bibfield{author}{\bibinfo{person}{Jingwei Huang}, \bibinfo{person}{Yichao
  Zhou}, {and} \bibinfo{person}{Leonidas Guibas}.}
  \bibinfo{year}{2020}\natexlab{}.
\newblock \bibinfo{title}{ManifoldPlus: A Robust and Scalable Watertight
  Manifold Surface Generation Method for Triangle Soups}.
\newblock
\newblock
\showeprint[arxiv]{2005.11621}~[cs.GR]


\bibitem[Huang et~al\mbox{.}(2018b)]%
        {quadriflow}
\bibfield{author}{\bibinfo{person}{Jingwei Huang}, \bibinfo{person}{Yichao
  Zhou}, \bibinfo{person}{Matthias Niessner}, \bibinfo{person}{Jonathan~Richard
  Shewchuk}, {and} \bibinfo{person}{Leonidas~J. Guibas}.}
  \bibinfo{year}{2018}\natexlab{b}.
\newblock \showarticletitle{{QuadriFlow: A Scalable and Robust Method for
  Quadrangulation}}.
\newblock \bibinfo{journal}{\emph{Computer Graphics Forum}}
  \bibinfo{volume}{37} (\bibinfo{year}{2018}).
\newblock
\showISSN{1467-8659}
\urldef\tempurl%
\url{https://doi.org/10.1111/cgf.13498}
\showDOI{\tempurl}


\bibitem[Huang et~al\mbox{.}(2024)]%
        {huang2023dreamtime}
\bibfield{author}{\bibinfo{person}{Yukun Huang}, \bibinfo{person}{Jianan Wang},
  \bibinfo{person}{Yukai Shi}, \bibinfo{person}{Boshi Tang},
  \bibinfo{person}{Xianbiao Qi}, {and} \bibinfo{person}{Lei Zhang}.}
  \bibinfo{year}{2024}\natexlab{}.
\newblock \showarticletitle{DreamTime: An Improved Optimization Strategy for
  Diffusion-Guided 3D Generation}. In \bibinfo{booktitle}{\emph{The Twelfth
  International Conference on Learning Representations}}.
\newblock


\bibitem[Jun and Nichol(2023)]%
        {jun2023shap}
\bibfield{author}{\bibinfo{person}{Heewoo Jun} {and} \bibinfo{person}{Alex
  Nichol}.} \bibinfo{year}{2023}\natexlab{}.
\newblock \bibinfo{title}{Shap-E: Generating Conditional 3D Implicit
  Functions}.
\newblock
\newblock
\showeprint[arxiv]{2305.02463}~[cs.CV]


\bibitem[Kerbl et~al\mbox{.}(2023)]%
        {kerbl3Dgaussians}
\bibfield{author}{\bibinfo{person}{Bernhard Kerbl}, \bibinfo{person}{Georgios
  Kopanas}, \bibinfo{person}{Thomas Leimk{\"u}hler}, {and}
  \bibinfo{person}{George Drettakis}.} \bibinfo{year}{2023}\natexlab{}.
\newblock \showarticletitle{3D Gaussian Splatting for Real-Time Radiance Field
  Rendering}.
\newblock \bibinfo{journal}{\emph{ACM Transactions on Graphics}}
  \bibinfo{volume}{42}, \bibinfo{number}{4} (\bibinfo{date}{July}
  \bibinfo{year}{2023}).
\newblock
\urldef\tempurl%
\url{https://repo-sam.inria.fr/fungraph/3d-gaussian-splatting/}
\showURL{%
\tempurl}


\bibitem[Li et~al\mbox{.}(2023)]%
        {li2023instant3d}
\bibfield{author}{\bibinfo{person}{Sixu Li}, \bibinfo{person}{Chaojian Li},
  \bibinfo{person}{Wenbo Zhu}, \bibinfo{person}{Boyang~(Tony) Yu},
  \bibinfo{person}{Yang~(Katie) Zhao}, \bibinfo{person}{Cheng Wan},
  \bibinfo{person}{Haoran You}, \bibinfo{person}{Huihong Shi}, {and}
  \bibinfo{person}{Yingyan~(Celine) Lin}.} \bibinfo{year}{2023}\natexlab{}.
\newblock \showarticletitle{Instant-3D: Instant Neural Radiance Field Training
  Towards On-Device AR/VR 3D Reconstruction}. In
  \bibinfo{booktitle}{\emph{Proceedings of the 50th Annual International
  Symposium on Computer Architecture}} (Orlando, FL, USA)
  \emph{(\bibinfo{series}{ISCA '23})}. \bibinfo{publisher}{Association for
  Computing Machinery}, \bibinfo{address}{New York, NY, USA}, Article
  \bibinfo{articleno}{6}, \bibinfo{numpages}{13}~pages.
\newblock
\showISBNx{9798400700958}
\urldef\tempurl%
\url{https://doi.org/10.1145/3579371.3589115}
\showDOI{\tempurl}


\bibitem[Li et~al\mbox{.}(2024)]%
        {li2023sweetdreamer}
\bibfield{author}{\bibinfo{person}{Weiyu Li}, \bibinfo{person}{Rui Chen},
  \bibinfo{person}{Xuelin Chen}, {and} \bibinfo{person}{Ping Tan}.}
  \bibinfo{year}{2024}\natexlab{}.
\newblock \showarticletitle{SweetDreamer: Aligning Geometric Priors in 2D
  diffusion for Consistent Text-to-3D}. In \bibinfo{booktitle}{\emph{The
  Twelfth International Conference on Learning Representations}}.
\newblock


\bibitem[Lin et~al\mbox{.}(2023)]%
        {lin2023magic3d}
\bibfield{author}{\bibinfo{person}{C. Lin}, \bibinfo{person}{J. Gao},
  \bibinfo{person}{L. Tang}, \bibinfo{person}{T. Takikawa}, \bibinfo{person}{X.
  Zeng}, \bibinfo{person}{X. Huang}, \bibinfo{person}{K. Kreis},
  \bibinfo{person}{S. Fidler}, \bibinfo{person}{M. Liu}, {and}
  \bibinfo{person}{T. Lin}.} \bibinfo{year}{2023}\natexlab{}.
\newblock \showarticletitle{Magic3D: High-Resolution Text-to-3D Content
  Creation}. In \bibinfo{booktitle}{\emph{2023 IEEE/CVF Conference on Computer
  Vision and Pattern Recognition (CVPR)}}. \bibinfo{publisher}{IEEE Computer
  Society}, \bibinfo{address}{Los Alamitos, CA, USA},
  \bibinfo{pages}{300--309}.
\newblock
\urldef\tempurl%
\url{https://doi.org/10.1109/CVPR52729.2023.00037}
\showDOI{\tempurl}


\bibitem[Lin et~al\mbox{.}(2024)]%
        {lin2024common}
\bibfield{author}{\bibinfo{person}{Shanchuan Lin}, \bibinfo{person}{Bingchen
  Liu}, \bibinfo{person}{Jiashi Li}, {and} \bibinfo{person}{Xiao Yang}.}
  \bibinfo{year}{2024}\natexlab{}.
\newblock \showarticletitle{Common diffusion noise schedules and sample steps
  are flawed}. In \bibinfo{booktitle}{\emph{Proceedings of the IEEE/CVF Winter
  Conference on Applications of Computer Vision}}. \bibinfo{pages}{5404--5411}.
\newblock


\bibitem[Ling et~al\mbox{.}(2024)]%
        {ling2023align}
\bibfield{author}{\bibinfo{person}{Huan Ling}, \bibinfo{person}{Seung~Wook
  Kim}, \bibinfo{person}{Antonio Torralba}, \bibinfo{person}{Sanja Fidler},
  {and} \bibinfo{person}{Karsten Kreis}.} \bibinfo{year}{2024}\natexlab{}.
\newblock \showarticletitle{Align Your Gaussians: Text-to-4D with Dynamic 3D
  Gaussians and Composed Diffusion Models}. In
  \bibinfo{booktitle}{\emph{Proceedings of the IEEE/CVF conference on computer
  vision and pattern recognition}}.
\newblock


\bibitem[Liu et~al\mbox{.}(2024b)]%
        {liu2023one}
\bibfield{author}{\bibinfo{person}{Minghua Liu}, \bibinfo{person}{Ruoxi Shi},
  \bibinfo{person}{Linghao Chen}, \bibinfo{person}{Zhuoyang Zhang},
  \bibinfo{person}{Chao Xu}, \bibinfo{person}{Xinyue Wei},
  \bibinfo{person}{Hansheng Chen}, \bibinfo{person}{Chong Zeng},
  \bibinfo{person}{Jiayuan Gu}, {and} \bibinfo{person}{Hao Su}.}
  \bibinfo{year}{2024}\natexlab{b}.
\newblock \showarticletitle{One-2-3-45++: Fast Single Image to 3D Objects with
  Consistent Multi-View Generation and 3D Diffusion}. In
  \bibinfo{booktitle}{\emph{Proceedings of the IEEE/CVF conference on computer
  vision and pattern recognition}}.
\newblock


\bibitem[Liu et~al\mbox{.}(2023d)]%
        {liu2023one2345}
\bibfield{author}{\bibinfo{person}{Minghua Liu}, \bibinfo{person}{Chao Xu},
  \bibinfo{person}{Haian Jin}, \bibinfo{person}{Linghao Chen},
  \bibinfo{person}{Mukund Varma~T}, \bibinfo{person}{Zexiang Xu}, {and}
  \bibinfo{person}{Hao Su}.} \bibinfo{year}{2023}\natexlab{d}.
\newblock \showarticletitle{One-2-3-45: Any Single Image to 3D Mesh in 45
  Seconds without Per-Shape Optimization}. In
  \bibinfo{booktitle}{\emph{Advances in Neural Information Processing
  Systems}}, \bibfield{editor}{\bibinfo{person}{A.~Oh},
  \bibinfo{person}{T.~Neumann}, \bibinfo{person}{A.~Globerson},
  \bibinfo{person}{K.~Saenko}, \bibinfo{person}{M.~Hardt}, {and}
  \bibinfo{person}{S.~Levine}} (Eds.), Vol.~\bibinfo{volume}{36}.
  \bibinfo{publisher}{Curran Associates, Inc.}, \bibinfo{pages}{22226--22246}.
\newblock
\urldef\tempurl%
\url{https://proceedings.neurips.cc/paper_files/paper/2023/file/4683beb6bab325650db13afd05d1a14a-Paper-Conference.pdf}
\showURL{%
\tempurl}


\bibitem[Liu et~al\mbox{.}(2023c)]%
        {liu2023zero}
\bibfield{author}{\bibinfo{person}{R. Liu}, \bibinfo{person}{R. Wu},
  \bibinfo{person}{B.~Van Hoorick}, \bibinfo{person}{P. Tokmakov},
  \bibinfo{person}{S. Zakharov}, {and} \bibinfo{person}{C. Vondrick}.}
  \bibinfo{year}{2023}\natexlab{c}.
\newblock \showarticletitle{Zero-1-to-3: Zero-shot One Image to 3D Object}. In
  \bibinfo{booktitle}{\emph{2023 IEEE/CVF International Conference on Computer
  Vision (ICCV)}}. \bibinfo{publisher}{IEEE Computer Society},
  \bibinfo{address}{Los Alamitos, CA, USA}, \bibinfo{pages}{9264--9275}.
\newblock
\urldef\tempurl%
\url{https://doi.org/10.1109/ICCV51070.2023.00853}
\showDOI{\tempurl}


\bibitem[Liu et~al\mbox{.}(2023b)]%
        {liu2023hyperhuman}
\bibfield{author}{\bibinfo{person}{Xian Liu}, \bibinfo{person}{Jian Ren},
  \bibinfo{person}{Aliaksandr Siarohin}, \bibinfo{person}{Ivan Skorokhodov},
  \bibinfo{person}{Yanyu Li}, \bibinfo{person}{Dahua Lin},
  \bibinfo{person}{Xihui Liu}, \bibinfo{person}{Ziwei Liu}, {and}
  \bibinfo{person}{Sergey Tulyakov}.} \bibinfo{year}{2023}\natexlab{b}.
\newblock \bibinfo{title}{HyperHuman: Hyper-Realistic Human Generation with
  Latent Structural Diffusion}.
\newblock
\newblock
\showeprint[arxiv]{2310.08579}~[cs.CV]


\bibitem[Liu et~al\mbox{.}(2024a)]%
        {liu2024syncdreamer}
\bibfield{author}{\bibinfo{person}{Yuan Liu}, \bibinfo{person}{Cheng Lin},
  \bibinfo{person}{Zijiao Zeng}, \bibinfo{person}{Xiaoxiao Long},
  \bibinfo{person}{Lingjie Liu}, \bibinfo{person}{Taku Komura}, {and}
  \bibinfo{person}{Wenping Wang}.} \bibinfo{year}{2024}\natexlab{a}.
\newblock \showarticletitle{SyncDreamer: Generating Multiview-consistent Images
  from a Single-view Image}. In \bibinfo{booktitle}{\emph{The Twelfth
  International Conference on Learning Representations}}.
\newblock


\bibitem[Liu et~al\mbox{.}(2023a)]%
        {liu2023unidream}
\bibfield{author}{\bibinfo{person}{Zexiang Liu}, \bibinfo{person}{Yangguang
  Li}, \bibinfo{person}{Youtian Lin}, \bibinfo{person}{Xin Yu},
  \bibinfo{person}{Sida Peng}, \bibinfo{person}{Yan-Pei Cao},
  \bibinfo{person}{Xiaojuan Qi}, \bibinfo{person}{Xiaoshui Huang},
  \bibinfo{person}{Ding Liang}, {and} \bibinfo{person}{Wanli Ouyang}.}
  \bibinfo{year}{2023}\natexlab{a}.
\newblock \bibinfo{title}{UniDream: Unifying Diffusion Priors for Relightable
  Text-to-3D Generation}.
\newblock
\newblock
\showeprint[arxiv]{2312.08754}~[cs.CV]


\bibitem[Long et~al\mbox{.}(2024)]%
        {long2023wonder3d}
\bibfield{author}{\bibinfo{person}{Xiaoxiao Long}, \bibinfo{person}{Yuan-Chen
  Guo}, \bibinfo{person}{Cheng Lin}, \bibinfo{person}{Yuan Liu},
  \bibinfo{person}{Zhiyang Dou}, \bibinfo{person}{Lingjie Liu},
  \bibinfo{person}{Yuexin Ma}, \bibinfo{person}{Song-Hai Zhang},
  \bibinfo{person}{Marc Habermann}, \bibinfo{person}{Christian Theobalt}, {and}
  \bibinfo{person}{Wenping Wang}.} \bibinfo{year}{2024}\natexlab{}.
\newblock \showarticletitle{Wonder3D: Single Image to 3D using Cross-Domain
  Diffusion}. In \bibinfo{booktitle}{\emph{Proceedings of the IEEE/CVF
  conference on computer vision and pattern recognition}}.
\newblock


\bibitem[Long et~al\mbox{.}(2022)]%
        {long2022sparseneus}
\bibfield{author}{\bibinfo{person}{Xiaoxiao Long}, \bibinfo{person}{Cheng Lin},
  \bibinfo{person}{Peng Wang}, \bibinfo{person}{Taku Komura}, {and}
  \bibinfo{person}{Wenping Wang}.} \bibinfo{year}{2022}\natexlab{}.
\newblock \showarticletitle{SparseNeuS: Fast Generalizable Neural Surface
  Reconstruction from Sparse Views}. In \bibinfo{booktitle}{\emph{Computer
  Vision -- ECCV 2022}}, \bibfield{editor}{\bibinfo{person}{Shai Avidan},
  \bibinfo{person}{Gabriel Brostow}, \bibinfo{person}{Moustapha Ciss{\'e}},
  \bibinfo{person}{Giovanni~Maria Farinella}, {and} \bibinfo{person}{Tal
  Hassner}} (Eds.). \bibinfo{publisher}{Springer Nature Switzerland},
  \bibinfo{address}{Cham}, \bibinfo{pages}{210--227}.
\newblock
\showISBNx{978-3-031-19824-3}


\bibitem[Marian(2021)]%
        {mesh_to_sdf}
\bibfield{author}{\bibinfo{person}{Kleineberg Marian}.}
  \bibinfo{year}{2021}\natexlab{}.
\newblock \bibinfo{title}{mesh\_to\_sdf: Calculate signed distance fields for
  arbitrary meshes}.
\newblock
  \bibinfo{howpublished}{\url{https://github.com/marian42/mesh_to_sdf}}.
\newblock


\bibitem[Mescheder et~al\mbox{.}(2019)]%
        {mescheder2019occupancy}
\bibfield{author}{\bibinfo{person}{Lars Mescheder}, \bibinfo{person}{Michael
  Oechsle}, \bibinfo{person}{Michael Niemeyer}, \bibinfo{person}{Sebastian
  Nowozin}, {and} \bibinfo{person}{Andreas Geiger}.}
  \bibinfo{year}{2019}\natexlab{}.
\newblock \showarticletitle{Occupancy networks: Learning 3d reconstruction in
  function space}. In \bibinfo{booktitle}{\emph{Proceedings of the IEEE/CVF
  conference on computer vision and pattern recognition}}.
  \bibinfo{pages}{4460--4470}.
\newblock


\bibitem[Metzer et~al\mbox{.}(2023)]%
        {latentnerf}
\bibfield{author}{\bibinfo{person}{G. Metzer}, \bibinfo{person}{E. Richardson},
  \bibinfo{person}{O. Patashnik}, \bibinfo{person}{R. Giryes}, {and}
  \bibinfo{person}{D. Cohen-Or}.} \bibinfo{year}{2023}\natexlab{}.
\newblock \showarticletitle{Latent-NeRF for Shape-Guided Generation of 3D
  Shapes and Textures}. In \bibinfo{booktitle}{\emph{2023 IEEE/CVF Conference
  on Computer Vision and Pattern Recognition (CVPR)}}. \bibinfo{publisher}{IEEE
  Computer Society}, \bibinfo{address}{Los Alamitos, CA, USA},
  \bibinfo{pages}{12663--12673}.
\newblock
\urldef\tempurl%
\url{https://doi.org/10.1109/CVPR52729.2023.01218}
\showDOI{\tempurl}


\bibitem[Mildenhall et~al\mbox{.}(2021)]%
        {mildenhall2021nerf}
\bibfield{author}{\bibinfo{person}{Ben Mildenhall}, \bibinfo{person}{Pratul~P
  Srinivasan}, \bibinfo{person}{Matthew Tancik}, \bibinfo{person}{Jonathan~T
  Barron}, \bibinfo{person}{Ravi Ramamoorthi}, {and} \bibinfo{person}{Ren Ng}.}
  \bibinfo{year}{2021}\natexlab{}.
\newblock \showarticletitle{Nerf: Representing scenes as neural radiance fields
  for view synthesis}.
\newblock \bibinfo{journal}{\emph{Commun. ACM}} \bibinfo{volume}{65},
  \bibinfo{number}{1} (\bibinfo{year}{2021}), \bibinfo{pages}{99--106}.
\newblock


\bibitem[Nash et~al\mbox{.}(2020)]%
        {nash2020polygen}
\bibfield{author}{\bibinfo{person}{Charlie Nash}, \bibinfo{person}{Yaroslav
  Ganin}, \bibinfo{person}{SM~Ali Eslami}, {and} \bibinfo{person}{Peter
  Battaglia}.} \bibinfo{year}{2020}\natexlab{}.
\newblock \showarticletitle{Polygen: An autoregressive generative model of 3d
  meshes}. In \bibinfo{booktitle}{\emph{International conference on machine
  learning}}. PMLR, \bibinfo{pages}{7220--7229}.
\newblock


\bibitem[Nichol et~al\mbox{.}(2022)]%
        {nichol2022point}
\bibfield{author}{\bibinfo{person}{Alex Nichol}, \bibinfo{person}{Heewoo Jun},
  \bibinfo{person}{Prafulla Dhariwal}, \bibinfo{person}{Pamela Mishkin}, {and}
  \bibinfo{person}{Mark Chen}.} \bibinfo{year}{2022}\natexlab{}.
\newblock \bibinfo{title}{Point-E: A System for Generating 3D Point Clouds from
  Complex Prompts}.
\newblock
\newblock
\showeprint[arxiv]{2212.08751}~[cs.CV]


\bibitem[OpenAI(2023)]%
        {openai2023gpt4v}
\bibfield{author}{\bibinfo{person}{OpenAI}.} \bibinfo{year}{2023}\natexlab{}.
\newblock \bibinfo{title}{GPT-4V: Generative Pre-trained Transformer 4 for
  Vision}.
\newblock \bibinfo{howpublished}{\url{https://www.openai.com/}}.
\newblock


\bibitem[Oquab et~al\mbox{.}(2024)]%
        {oquab2023dinov2}
\bibfield{author}{\bibinfo{person}{Maxime Oquab}, \bibinfo{person}{Timoth{\'e}e
  Darcet}, \bibinfo{person}{Th{\'e}o Moutakanni}, \bibinfo{person}{Huy~V. Vo},
  \bibinfo{person}{Marc Szafraniec}, \bibinfo{person}{Vasil Khalidov},
  \bibinfo{person}{Pierre Fernandez}, \bibinfo{person}{Daniel HAZIZA},
  \bibinfo{person}{Francisco Massa}, \bibinfo{person}{Alaaeldin El-Nouby},
  \bibinfo{person}{Mido Assran}, \bibinfo{person}{Nicolas Ballas},
  \bibinfo{person}{Wojciech Galuba}, \bibinfo{person}{Russell Howes},
  \bibinfo{person}{Po-Yao Huang}, \bibinfo{person}{Shang-Wen Li},
  \bibinfo{person}{Ishan Misra}, \bibinfo{person}{Michael Rabbat},
  \bibinfo{person}{Vasu Sharma}, \bibinfo{person}{Gabriel Synnaeve},
  \bibinfo{person}{Hu Xu}, \bibinfo{person}{Herve Jegou},
  \bibinfo{person}{Julien Mairal}, \bibinfo{person}{Patrick Labatut},
  \bibinfo{person}{Armand Joulin}, {and} \bibinfo{person}{Piotr Bojanowski}.}
  \bibinfo{year}{2024}\natexlab{}.
\newblock \showarticletitle{{DINO}v2: Learning Robust Visual Features without
  Supervision}.
\newblock \bibinfo{journal}{\emph{Transactions on Machine Learning Research}}
  (\bibinfo{year}{2024}).
\newblock
\showISSN{2835-8856}


\bibitem[Park et~al\mbox{.}(2019)]%
        {park2019deepsdf}
\bibfield{author}{\bibinfo{person}{J. Park}, \bibinfo{person}{P. Florence},
  \bibinfo{person}{J. Straub}, \bibinfo{person}{R. Newcombe}, {and}
  \bibinfo{person}{S. Lovegrove}.} \bibinfo{year}{2019}\natexlab{}.
\newblock \showarticletitle{DeepSDF: Learning Continuous Signed Distance
  Functions for Shape Representation}. In \bibinfo{booktitle}{\emph{2019
  IEEE/CVF Conference on Computer Vision and Pattern Recognition (CVPR)}}.
  \bibinfo{publisher}{IEEE Computer Society}, \bibinfo{address}{Los Alamitos,
  CA, USA}, \bibinfo{pages}{165--174}.
\newblock
\urldef\tempurl%
\url{https://doi.org/10.1109/CVPR.2019.00025}
\showDOI{\tempurl}


\bibitem[Peng et~al\mbox{.}(2020)]%
        {peng2020convolutional}
\bibfield{author}{\bibinfo{person}{Songyou Peng}, \bibinfo{person}{Michael
  Niemeyer}, \bibinfo{person}{Lars Mescheder}, \bibinfo{person}{Marc
  Pollefeys}, {and} \bibinfo{person}{Andreas Geiger}.}
  \bibinfo{year}{2020}\natexlab{}.
\newblock \showarticletitle{Convolutional occupancy networks}. In
  \bibinfo{booktitle}{\emph{Computer Vision--ECCV 2020: 16th European
  Conference, Glasgow, UK, August 23--28, 2020, Proceedings, Part III 16}}.
  Springer, \bibinfo{pages}{523--540}.
\newblock


\bibitem[Po et~al\mbox{.}(2023)]%
        {po2023state}
\bibfield{author}{\bibinfo{person}{Ryan Po}, \bibinfo{person}{Wang Yifan},
  \bibinfo{person}{Vladislav Golyanik}, \bibinfo{person}{Kfir Aberman},
  \bibinfo{person}{Jonathan~T Barron}, \bibinfo{person}{Amit~H Bermano},
  \bibinfo{person}{Eric~Ryan Chan}, \bibinfo{person}{Tali Dekel},
  \bibinfo{person}{Aleksander Holynski}, \bibinfo{person}{Angjoo Kanazawa},
  {et~al\mbox{.}}} \bibinfo{year}{2023}\natexlab{}.
\newblock \showarticletitle{State of the art on diffusion models for visual
  computing}.
\newblock \bibinfo{journal}{\emph{arXiv preprint arXiv:2310.07204}}
  (\bibinfo{year}{2023}).
\newblock


\bibitem[Podell et~al\mbox{.}(2023)]%
        {podell2023sdxl}
\bibfield{author}{\bibinfo{person}{Dustin Podell}, \bibinfo{person}{Zion
  English}, \bibinfo{person}{Kyle Lacey}, \bibinfo{person}{Andreas Blattmann},
  \bibinfo{person}{Tim Dockhorn}, \bibinfo{person}{Jonas Müller},
  \bibinfo{person}{Joe Penna}, {and} \bibinfo{person}{Robin Rombach}.}
  \bibinfo{year}{2023}\natexlab{}.
\newblock \bibinfo{title}{SDXL: Improving Latent Diffusion Models for
  High-Resolution Image Synthesis}.
\newblock
\newblock
\showeprint[arxiv]{2307.01952}~[cs.CV]


\bibitem[Poole et~al\mbox{.}(2023)]%
        {poole2022dreamfusion}
\bibfield{author}{\bibinfo{person}{Ben Poole}, \bibinfo{person}{Ajay Jain},
  \bibinfo{person}{Jonathan~T. Barron}, {and} \bibinfo{person}{Ben
  Mildenhall}.} \bibinfo{year}{2023}\natexlab{}.
\newblock \showarticletitle{DreamFusion: Text-to-3D using 2D Diffusion}. In
  \bibinfo{booktitle}{\emph{The Eleventh International Conference on Learning
  Representations}}.
\newblock


\bibitem[Qi et~al\mbox{.}(2017)]%
        {qi2017pointnet2}
\bibfield{author}{\bibinfo{person}{Charles~R. Qi}, \bibinfo{person}{Li Yi},
  \bibinfo{person}{Hao Su}, {and} \bibinfo{person}{Leonidas~J. Guibas}.}
  \bibinfo{year}{2017}\natexlab{}.
\newblock \showarticletitle{PointNet++: deep hierarchical feature learning on
  point sets in a metric space}.
\newblock   \bibinfo{volume}{30} (\bibinfo{year}{2017}),
  \bibinfo{pages}{5105–5114}.
\newblock
\showISBNx{9781510860964}


\bibitem[Qian et~al\mbox{.}(2024)]%
        {qian2023magic123}
\bibfield{author}{\bibinfo{person}{Guocheng Qian}, \bibinfo{person}{Jinjie
  Mai}, \bibinfo{person}{Abdullah Hamdi}, \bibinfo{person}{Jian Ren},
  \bibinfo{person}{Aliaksandr Siarohin}, \bibinfo{person}{Bing Li},
  \bibinfo{person}{Hsin-Ying Lee}, \bibinfo{person}{Ivan Skorokhodov},
  \bibinfo{person}{Peter Wonka}, \bibinfo{person}{Sergey Tulyakov}, {and}
  \bibinfo{person}{Bernard Ghanem}.} \bibinfo{year}{2024}\natexlab{}.
\newblock \showarticletitle{Magic123: One Image to High-Quality 3D Object
  Generation Using Both 2D and 3D Diffusion Priors}. In
  \bibinfo{booktitle}{\emph{The Twelfth International Conference on Learning
  Representations}}.
\newblock


\bibitem[Qiu et~al\mbox{.}(2024)]%
        {qiu2023richdreamer}
\bibfield{author}{\bibinfo{person}{Lingteng Qiu}, \bibinfo{person}{Guanying
  Chen}, \bibinfo{person}{Xiaodong Gu}, \bibinfo{person}{Qi Zuo},
  \bibinfo{person}{Mutian Xu}, \bibinfo{person}{Yushuang Wu},
  \bibinfo{person}{Weihao Yuan}, \bibinfo{person}{Zilong Dong},
  \bibinfo{person}{Liefeng Bo}, {and} \bibinfo{person}{Xiaoguang Han}.}
  \bibinfo{year}{2024}\natexlab{}.
\newblock \showarticletitle{RichDreamer: A Generalizable Normal-Depth Diffusion
  Model for Detail Richness in Text-to-3D}. In
  \bibinfo{booktitle}{\emph{Proceedings of the IEEE/CVF conference on computer
  vision and pattern recognition}}.
\newblock


\bibitem[Radford et~al\mbox{.}(2021)]%
        {radford2021clip}
\bibfield{author}{\bibinfo{person}{Alec Radford}, \bibinfo{person}{Jong~Wook
  Kim}, \bibinfo{person}{Chris Hallacy}, \bibinfo{person}{Aditya Ramesh},
  \bibinfo{person}{Gabriel Goh}, \bibinfo{person}{Sandhini Agarwal},
  \bibinfo{person}{Girish Sastry}, \bibinfo{person}{Amanda Askell},
  \bibinfo{person}{Pamela Mishkin}, \bibinfo{person}{Jack Clark},
  {et~al\mbox{.}}} \bibinfo{year}{2021}\natexlab{}.
\newblock \showarticletitle{Learning transferable visual models from natural
  language supervision}. In \bibinfo{booktitle}{\emph{International conference
  on machine learning}}. PMLR, \bibinfo{pages}{8748--8763}.
\newblock


\bibitem[Ramesh et~al\mbox{.}(2021)]%
        {Dall-e}
\bibfield{author}{\bibinfo{person}{Aditya Ramesh}, \bibinfo{person}{Mikhail
  Pavlov}, \bibinfo{person}{Gabriel Goh}, \bibinfo{person}{Scott Gray},
  \bibinfo{person}{Chelsea Voss}, \bibinfo{person}{Alec Radford},
  \bibinfo{person}{Mark Chen}, {and} \bibinfo{person}{Ilya Sutskever}.}
  \bibinfo{year}{2021}\natexlab{}.
\newblock \showarticletitle{Zero-Shot Text-to-Image Generation}. In
  \bibinfo{booktitle}{\emph{Proceedings of the 38th International Conference on
  Machine Learning, {ICML} 2021, 18-24 July 2021, Virtual Event}}
  \emph{(\bibinfo{series}{Proceedings of Machine Learning Research},
  Vol.~\bibinfo{volume}{139})}, \bibfield{editor}{\bibinfo{person}{Marina
  Meila} {and} \bibinfo{person}{Tong Zhang}} (Eds.).
  \bibinfo{publisher}{{PMLR}}, \bibinfo{pages}{8821--8831}.
\newblock
\urldef\tempurl%
\url{http://proceedings.mlr.press/v139/ramesh21a.html}
\showURL{%
\tempurl}


\bibitem[Reizenstein et~al\mbox{.}(2021)]%
        {reizenstein2021common}
\bibfield{author}{\bibinfo{person}{J. Reizenstein}, \bibinfo{person}{R.
  Shapovalov}, \bibinfo{person}{P. Henzler}, \bibinfo{person}{L. Sbordone},
  \bibinfo{person}{P. Labatut}, {and} \bibinfo{person}{D. Novotny}.}
  \bibinfo{year}{2021}\natexlab{}.
\newblock \showarticletitle{Common Objects in 3D: Large-Scale Learning and
  Evaluation of Real-life 3D Category Reconstruction}. In
  \bibinfo{booktitle}{\emph{2021 IEEE/CVF International Conference on Computer
  Vision (ICCV)}}. \bibinfo{publisher}{IEEE Computer Society},
  \bibinfo{address}{Los Alamitos, CA, USA}, \bibinfo{pages}{10881--10891}.
\newblock
\urldef\tempurl%
\url{https://doi.org/10.1109/ICCV48922.2021.01072}
\showDOI{\tempurl}


\bibitem[Ren et~al\mbox{.}(2024)]%
        {ren2023xcube}
\bibfield{author}{\bibinfo{person}{Xuanchi Ren}, \bibinfo{person}{Jiahui
  Huang}, \bibinfo{person}{Xiaohui Zeng}, \bibinfo{person}{Ken Museth},
  \bibinfo{person}{Sanja Fidler}, {and} \bibinfo{person}{Francis Williams}.}
  \bibinfo{year}{2024}\natexlab{}.
\newblock \showarticletitle{XCube ($\mathcal{X}^3$): Large-Scale 3D Generative
  Modeling using Sparse Voxel Hierarchies}. In
  \bibinfo{booktitle}{\emph{Proceedings of the IEEE/CVF conference on computer
  vision and pattern recognition}}.
\newblock


\bibitem[Richardson et~al\mbox{.}(2023)]%
        {TEXTure}
\bibfield{author}{\bibinfo{person}{Elad Richardson}, \bibinfo{person}{Gal
  Metzer}, \bibinfo{person}{Yuval Alaluf}, \bibinfo{person}{Raja Giryes}, {and}
  \bibinfo{person}{Daniel Cohen-Or}.} \bibinfo{year}{2023}\natexlab{}.
\newblock \showarticletitle{TEXTure: Text-Guided Texturing of 3D Shapes}. In
  \bibinfo{booktitle}{\emph{ACM SIGGRAPH 2023 Conference Proceedings}} (Los
  Angeles, CA, USA) \emph{(\bibinfo{series}{SIGGRAPH '23})}.
  \bibinfo{publisher}{Association for Computing Machinery},
  \bibinfo{address}{New York, NY, USA}, Article \bibinfo{articleno}{54},
  \bibinfo{numpages}{11}~pages.
\newblock
\showISBNx{9798400701597}
\urldef\tempurl%
\url{https://doi.org/10.1145/3588432.3591503}
\showDOI{\tempurl}


\bibitem[Rombach et~al\mbox{.}(2022)]%
        {sd}
\bibfield{author}{\bibinfo{person}{R. Rombach}, \bibinfo{person}{A. Blattmann},
  \bibinfo{person}{D. Lorenz}, \bibinfo{person}{P. Esser}, {and}
  \bibinfo{person}{B. Ommer}.} \bibinfo{year}{2022}\natexlab{}.
\newblock \showarticletitle{High-Resolution Image Synthesis with Latent
  Diffusion Models}. In \bibinfo{booktitle}{\emph{2022 IEEE/CVF Conference on
  Computer Vision and Pattern Recognition (CVPR)}}. \bibinfo{publisher}{IEEE
  Computer Society}, \bibinfo{address}{Los Alamitos, CA, USA},
  \bibinfo{pages}{10674--10685}.
\newblock
\urldef\tempurl%
\url{https://doi.org/10.1109/CVPR52688.2022.01042}
\showDOI{\tempurl}


\bibitem[Saharia et~al\mbox{.}(2022)]%
        {imagen}
\bibfield{author}{\bibinfo{person}{Chitwan Saharia}, \bibinfo{person}{William
  Chan}, \bibinfo{person}{Saurabh Saxena}, \bibinfo{person}{Lala Li},
  \bibinfo{person}{Jay Whang}, \bibinfo{person}{Emily~L Denton},
  \bibinfo{person}{Kamyar Ghasemipour}, \bibinfo{person}{Raphael
  Gontijo~Lopes}, \bibinfo{person}{Burcu Karagol~Ayan}, \bibinfo{person}{Tim
  Salimans}, \bibinfo{person}{Jonathan Ho}, \bibinfo{person}{David~J Fleet},
  {and} \bibinfo{person}{Mohammad Norouzi}.} \bibinfo{year}{2022}\natexlab{}.
\newblock \showarticletitle{Photorealistic Text-to-Image Diffusion Models with
  Deep Language Understanding}. In \bibinfo{booktitle}{\emph{Advances in Neural
  Information Processing Systems}},
  \bibfield{editor}{\bibinfo{person}{S.~Koyejo}, \bibinfo{person}{S.~Mohamed},
  \bibinfo{person}{A.~Agarwal}, \bibinfo{person}{D.~Belgrave},
  \bibinfo{person}{K.~Cho}, {and} \bibinfo{person}{A.~Oh}} (Eds.),
  Vol.~\bibinfo{volume}{35}. \bibinfo{publisher}{Curran Associates, Inc.},
  \bibinfo{pages}{36479--36494}.
\newblock
\urldef\tempurl%
\url{https://proceedings.neurips.cc/paper_files/paper/2022/file/ec795aeadae0b7d230fa35cbaf04c041-Paper-Conference.pdf}
\showURL{%
\tempurl}


\bibitem[Seo et~al\mbox{.}(2024)]%
        {seo2024let}
\bibfield{author}{\bibinfo{person}{Junyoung Seo}, \bibinfo{person}{Wooseok
  Jang}, \bibinfo{person}{Min-Seop Kwak}, \bibinfo{person}{Hyeonsu Kim},
  \bibinfo{person}{Jaehoon Ko}, \bibinfo{person}{Junho Kim},
  \bibinfo{person}{Jin-Hwa Kim}, \bibinfo{person}{Jiyoung Lee}, {and}
  \bibinfo{person}{Seungryong Kim}.} \bibinfo{year}{2024}\natexlab{}.
\newblock \showarticletitle{Let 2D Diffusion Model Know 3D-Consistency for
  Robust Text-to-3D Generation}. In \bibinfo{booktitle}{\emph{The Twelfth
  International Conference on Learning Representations}}.
\newblock


\bibitem[Shen et~al\mbox{.}(2021)]%
        {shen2021deep}
\bibfield{author}{\bibinfo{person}{Tianchang Shen}, \bibinfo{person}{Jun Gao},
  \bibinfo{person}{Kangxue Yin}, \bibinfo{person}{Ming-Yu Liu}, {and}
  \bibinfo{person}{Sanja Fidler}.} \bibinfo{year}{2021}\natexlab{}.
\newblock \showarticletitle{Deep Marching Tetrahedra: a Hybrid Representation
  for High-Resolution 3D Shape Synthesis}. In
  \bibinfo{booktitle}{\emph{Advances in Neural Information Processing
  Systems}}, \bibfield{editor}{\bibinfo{person}{A.~Beygelzimer},
  \bibinfo{person}{Y.~Dauphin}, \bibinfo{person}{P.~Liang}, {and}
  \bibinfo{person}{J.~Wortman Vaughan}} (Eds.).
\newblock


\bibitem[Shi et~al\mbox{.}(2023)]%
        {shi2023zero123++}
\bibfield{author}{\bibinfo{person}{Ruoxi Shi}, \bibinfo{person}{Hansheng Chen},
  \bibinfo{person}{Zhuoyang Zhang}, \bibinfo{person}{Minghua Liu},
  \bibinfo{person}{Chao Xu}, \bibinfo{person}{Xinyue Wei},
  \bibinfo{person}{Linghao Chen}, \bibinfo{person}{Chong Zeng}, {and}
  \bibinfo{person}{Hao Su}.} \bibinfo{year}{2023}\natexlab{}.
\newblock \bibinfo{title}{Zero123++: a Single Image to Consistent Multi-view
  Diffusion Base Model}.
\newblock
\newblock
\showeprint[arxiv]{2310.15110}~[cs.CV]


\bibitem[Shi et~al\mbox{.}(2024)]%
        {shi2024mvdream}
\bibfield{author}{\bibinfo{person}{Yichun Shi}, \bibinfo{person}{Peng Wang},
  \bibinfo{person}{Jianglong Ye}, \bibinfo{person}{Long Mai},
  \bibinfo{person}{Kejie Li}, {and} \bibinfo{person}{Xiao Yang}.}
  \bibinfo{year}{2024}\natexlab{}.
\newblock \showarticletitle{{MVD}ream: Multi-view Diffusion for 3D Generation}.
  In \bibinfo{booktitle}{\emph{The Twelfth International Conference on Learning
  Representations}}.
\newblock


\bibitem[Siddiqui et~al\mbox{.}(2024)]%
        {siddiqui2023meshgpt}
\bibfield{author}{\bibinfo{person}{Yawar Siddiqui}, \bibinfo{person}{Antonio
  Alliegro}, \bibinfo{person}{Alexey Artemov}, \bibinfo{person}{Tatiana
  Tommasi}, \bibinfo{person}{Daniele Sirigatti}, \bibinfo{person}{Vladislav
  Rosov}, \bibinfo{person}{Angela Dai}, {and} \bibinfo{person}{Matthias
  Nießner}.} \bibinfo{year}{2024}\natexlab{}.
\newblock \showarticletitle{MeshGPT: Generating Triangle Meshes with
  Decoder-Only Transformers}. In \bibinfo{booktitle}{\emph{Proceedings of the
  IEEE/CVF conference on computer vision and pattern recognition}}.
\newblock


\bibitem[Singer et~al\mbox{.}(2023)]%
        {singer2023text4d}
\bibfield{author}{\bibinfo{person}{Uriel Singer}, \bibinfo{person}{Shelly
  Sheynin}, \bibinfo{person}{Adam Polyak}, \bibinfo{person}{Oron Ashual},
  \bibinfo{person}{Iurii Makarov}, \bibinfo{person}{Filippos Kokkinos},
  \bibinfo{person}{Naman Goyal}, \bibinfo{person}{Andrea Vedaldi},
  \bibinfo{person}{Devi Parikh}, \bibinfo{person}{Justin Johnson}, {and}
  \bibinfo{person}{Yaniv Taigman}.} \bibinfo{year}{2023}\natexlab{}.
\newblock \showarticletitle{Text-To-4D Dynamic Scene Generation}. In
  \bibinfo{booktitle}{\emph{International Conference on Machine Learning,
  {ICML} 2023, 23-29 July 2023, Honolulu, Hawaii, {USA}}}
  \emph{(\bibinfo{series}{Proceedings of Machine Learning Research},
  Vol.~\bibinfo{volume}{202})}, \bibfield{editor}{\bibinfo{person}{Andreas
  Krause}, \bibinfo{person}{Emma Brunskill}, \bibinfo{person}{Kyunghyun Cho},
  \bibinfo{person}{Barbara Engelhardt}, \bibinfo{person}{Sivan Sabato}, {and}
  \bibinfo{person}{Jonathan Scarlett}} (Eds.). \bibinfo{publisher}{{PMLR}},
  \bibinfo{pages}{31915--31929}.
\newblock
\urldef\tempurl%
\url{https://proceedings.mlr.press/v202/singer23a.html}
\showURL{%
\tempurl}


\bibitem[Sun et~al\mbox{.}(2024)]%
        {sun2023dreamcraft3d}
\bibfield{author}{\bibinfo{person}{Jingxiang Sun}, \bibinfo{person}{Bo Zhang},
  \bibinfo{person}{Ruizhi Shao}, \bibinfo{person}{Lizhen Wang},
  \bibinfo{person}{Wen Liu}, \bibinfo{person}{Zhenda Xie}, {and}
  \bibinfo{person}{Yebin Liu}.} \bibinfo{year}{2024}\natexlab{}.
\newblock \showarticletitle{DreamCraft3D: Hierarchical 3D Generation with
  Bootstrapped Diffusion Prior}. In \bibinfo{booktitle}{\emph{The Twelfth
  International Conference on Learning Representations}}.
\newblock


\bibitem[Tang et~al\mbox{.}(2019)]%
        {tang2019skeleton}
\bibfield{author}{\bibinfo{person}{Jiapeng Tang}, \bibinfo{person}{Xiaoguang
  Han}, \bibinfo{person}{Junyi Pan}, \bibinfo{person}{Kui Jia}, {and}
  \bibinfo{person}{Xin Tong}.} \bibinfo{year}{2019}\natexlab{}.
\newblock \showarticletitle{A skeleton-bridged deep learning approach for
  generating meshes of complex topologies from single rgb images}. In
  \bibinfo{booktitle}{\emph{Proceedings of the ieee/cvf conference on computer
  vision and pattern recognition}}. \bibinfo{pages}{4541--4550}.
\newblock


\bibitem[Tang et~al\mbox{.}(2021a)]%
        {tang2021skeletonnet}
\bibfield{author}{\bibinfo{person}{Jiapeng Tang}, \bibinfo{person}{Xiaoguang
  Han}, \bibinfo{person}{Mingkui Tan}, \bibinfo{person}{Xin Tong}, {and}
  \bibinfo{person}{Kui Jia}.} \bibinfo{year}{2021}\natexlab{a}.
\newblock \showarticletitle{Skeletonnet: A topology-preserving solution for
  learning mesh reconstruction of object surfaces from rgb images}.
\newblock \bibinfo{journal}{\emph{IEEE transactions on pattern analysis and
  machine intelligence}} \bibinfo{volume}{44}, \bibinfo{number}{10}
  (\bibinfo{year}{2021}), \bibinfo{pages}{6454--6471}.
\newblock


\bibitem[Tang et~al\mbox{.}(2021b)]%
        {tang2021sa}
\bibfield{author}{\bibinfo{person}{Jiapeng Tang}, \bibinfo{person}{Jiabao Lei},
  \bibinfo{person}{Dan Xu}, \bibinfo{person}{Feiying Ma}, \bibinfo{person}{Kui
  Jia}, {and} \bibinfo{person}{Lei Zhang}.} \bibinfo{year}{2021}\natexlab{b}.
\newblock \showarticletitle{Sa-convonet: Sign-agnostic optimization of
  convolutional occupancy networks}. In \bibinfo{booktitle}{\emph{Proceedings
  of the IEEE/CVF International Conference on Computer Vision}}.
  \bibinfo{pages}{6504--6513}.
\newblock


\bibitem[Tang et~al\mbox{.}(2024)]%
        {tang2023dreamgaussian}
\bibfield{author}{\bibinfo{person}{Jiaxiang Tang}, \bibinfo{person}{Jiawei
  Ren}, \bibinfo{person}{Hang Zhou}, \bibinfo{person}{Ziwei Liu}, {and}
  \bibinfo{person}{Gang Zeng}.} \bibinfo{year}{2024}\natexlab{}.
\newblock \showarticletitle{DreamGaussian: Generative Gaussian Splatting for
  Efficient 3D Content Creation}. In \bibinfo{booktitle}{\emph{The Twelfth
  International Conference on Learning Representations}}.
\newblock


\bibitem[Vaswani et~al\mbox{.}(2017)]%
        {AttentionAllYouNeed}
\bibfield{author}{\bibinfo{person}{Ashish Vaswani}, \bibinfo{person}{Noam
  Shazeer}, \bibinfo{person}{Niki Parmar}, \bibinfo{person}{Jakob Uszkoreit},
  \bibinfo{person}{Llion Jones}, \bibinfo{person}{Aidan~N Gomez},
  \bibinfo{person}{\L~ukasz Kaiser}, {and} \bibinfo{person}{Illia Polosukhin}.}
  \bibinfo{year}{2017}\natexlab{}.
\newblock \showarticletitle{Attention is All you Need}. In
  \bibinfo{booktitle}{\emph{Advances in Neural Information Processing
  Systems}}, \bibfield{editor}{\bibinfo{person}{I.~Guyon},
  \bibinfo{person}{U.~Von Luxburg}, \bibinfo{person}{S.~Bengio},
  \bibinfo{person}{H.~Wallach}, \bibinfo{person}{R.~Fergus},
  \bibinfo{person}{S.~Vishwanathan}, {and} \bibinfo{person}{R.~Garnett}}
  (Eds.), Vol.~\bibinfo{volume}{30}. \bibinfo{publisher}{Curran Associates,
  Inc.}
\newblock
\urldef\tempurl%
\url{https://proceedings.neurips.cc/paper_files/paper/2017/file/3f5ee243547dee91fbd053c1c4a845aa-Paper.pdf}
\showURL{%
\tempurl}


\bibitem[Wang et~al\mbox{.}(2021a)]%
        {wang2021neus}
\bibfield{author}{\bibinfo{person}{Peng Wang}, \bibinfo{person}{Lingjie Liu},
  \bibinfo{person}{Yuan Liu}, \bibinfo{person}{Christian Theobalt},
  \bibinfo{person}{Taku Komura}, {and} \bibinfo{person}{Wenping Wang}.}
  \bibinfo{year}{2021}\natexlab{a}.
\newblock \showarticletitle{NeuS: Learning Neural Implicit Surfaces by Volume
  Rendering for Multi-view Reconstruction}.
\newblock \bibinfo{journal}{\emph{Advances in Neural Information Processing
  Systems}}  \bibinfo{volume}{34} (\bibinfo{year}{2021}),
  \bibinfo{pages}{27171--27183}.
\newblock


\bibitem[Wang et~al\mbox{.}(2024)]%
        {wang2023pf}
\bibfield{author}{\bibinfo{person}{Peng Wang}, \bibinfo{person}{Hao Tan},
  \bibinfo{person}{Sai Bi}, \bibinfo{person}{Yinghao Xu},
  \bibinfo{person}{Fujun Luan}, \bibinfo{person}{Kalyan Sunkavalli},
  \bibinfo{person}{Wenping Wang}, \bibinfo{person}{Zexiang Xu}, {and}
  \bibinfo{person}{Kai Zhang}.} \bibinfo{year}{2024}\natexlab{}.
\newblock \showarticletitle{{PF}-{LRM}: Pose-Free Large Reconstruction Model
  for Joint Pose and Shape Prediction}. In \bibinfo{booktitle}{\emph{The
  Twelfth International Conference on Learning Representations}}.
\newblock


\bibitem[Wang(2022)]%
        {wang2022mesh2sdf}
\bibfield{author}{\bibinfo{person}{Peng-Shuai Wang}.}
  \bibinfo{year}{2022}\natexlab{}.
\newblock \bibinfo{title}{mesh2sdf}.
\newblock \bibinfo{howpublished}{\url{https://github.com/wang-ps/mesh2sdf}}.
\newblock
\newblock
\shownote{Converts an input mesh to a signed distance field (SDF)}.


\bibitem[Wang et~al\mbox{.}(2022)]%
        {Wang-2022-dualocnn}
\bibfield{author}{\bibinfo{person}{Peng-Shuai Wang}, \bibinfo{person}{Yang
  Liu}, {and} \bibinfo{person}{Xin Tong}.} \bibinfo{year}{2022}\natexlab{}.
\newblock \showarticletitle{Dual octree graph networks for learning adaptive
  volumetric shape representations}.
\newblock \bibinfo{journal}{\emph{ACM Trans. Graph.}} \bibinfo{volume}{41},
  \bibinfo{number}{4}, Article \bibinfo{articleno}{103} (\bibinfo{date}{jul}
  \bibinfo{year}{2022}), \bibinfo{numpages}{15}~pages.
\newblock
\showISSN{0730-0301}
\urldef\tempurl%
\url{https://doi.org/10.1145/3528223.3530087}
\showDOI{\tempurl}


\bibitem[Wang et~al\mbox{.}(2021b)]%
        {wang2021realesrgan}
\bibfield{author}{\bibinfo{person}{X. Wang}, \bibinfo{person}{L. Xie},
  \bibinfo{person}{C. Dong}, {and} \bibinfo{person}{Y. Shan}.}
  \bibinfo{year}{2021}\natexlab{b}.
\newblock \showarticletitle{Real-ESRGAN: Training Real-World Blind
  Super-Resolution with Pure Synthetic Data}. In \bibinfo{booktitle}{\emph{2021
  IEEE/CVF International Conference on Computer Vision Workshops (ICCVW)}}.
  \bibinfo{publisher}{IEEE Computer Society}, \bibinfo{address}{Los Alamitos,
  CA, USA}, \bibinfo{pages}{1905--1914}.
\newblock
\urldef\tempurl%
\url{https://doi.org/10.1109/ICCVW54120.2021.00217}
\showDOI{\tempurl}


\bibitem[Wang et~al\mbox{.}(2023)]%
        {wang2023prolificdreamer}
\bibfield{author}{\bibinfo{person}{Zhengyi Wang}, \bibinfo{person}{Cheng Lu},
  \bibinfo{person}{Yikai Wang}, \bibinfo{person}{Fan Bao},
  \bibinfo{person}{Chongxuan Li}, \bibinfo{person}{Hang Su}, {and}
  \bibinfo{person}{Jun Zhu}.} \bibinfo{year}{2023}\natexlab{}.
\newblock \showarticletitle{ProlificDreamer: High-Fidelity and Diverse
  Text-to-3D Generation with Variational Score Distillation}. In
  \bibinfo{booktitle}{\emph{Thirty-seventh Conference on Neural Information
  Processing Systems}}.
\newblock


\bibitem[Wu et~al\mbox{.}(2024)]%
        {wu2024hd}
\bibfield{author}{\bibinfo{person}{Jinbo Wu}, \bibinfo{person}{Xiaobo Gao},
  \bibinfo{person}{Xing Liu}, \bibinfo{person}{Zhengyang Shen},
  \bibinfo{person}{Chen Zhao}, \bibinfo{person}{Haocheng Feng},
  \bibinfo{person}{Jingtuo Liu}, {and} \bibinfo{person}{Errui Ding}.}
  \bibinfo{year}{2024}\natexlab{}.
\newblock \showarticletitle{Hd-fusion: Detailed text-to-3d generation
  leveraging multiple noise estimation}. In
  \bibinfo{booktitle}{\emph{Proceedings of the IEEE/CVF Winter Conference on
  Applications of Computer Vision}}. \bibinfo{pages}{3202--3211}.
\newblock


\bibitem[Wu et~al\mbox{.}(2023)]%
        {wu2023omniobject3d}
\bibfield{author}{\bibinfo{person}{T. Wu}, \bibinfo{person}{J. Zhang},
  \bibinfo{person}{X. Fu}, \bibinfo{person}{Y. Wang}, \bibinfo{person}{J. Ren},
  \bibinfo{person}{L. Pan}, \bibinfo{person}{W. Wu}, \bibinfo{person}{L. Yang},
  \bibinfo{person}{J. Wang}, \bibinfo{person}{C. Qian}, \bibinfo{person}{D.
  Lin}, {and} \bibinfo{person}{Z. Liu}.} \bibinfo{year}{2023}\natexlab{}.
\newblock \showarticletitle{OmniObject3D: Large-Vocabulary 3D Object Dataset
  for Realistic Perception, Reconstruction and Generation}. In
  \bibinfo{booktitle}{\emph{2023 IEEE/CVF Conference on Computer Vision and
  Pattern Recognition (CVPR)}}. \bibinfo{publisher}{IEEE Computer Society},
  \bibinfo{address}{Los Alamitos, CA, USA}, \bibinfo{pages}{803--814}.
\newblock
\urldef\tempurl%
\url{https://doi.org/10.1109/CVPR52729.2023.00084}
\showDOI{\tempurl}


\bibitem[Xiong et~al\mbox{.}(2020)]%
        {xiong2020layer}
\bibfield{author}{\bibinfo{person}{Ruibin Xiong}, \bibinfo{person}{Yunchang
  Yang}, \bibinfo{person}{Di He}, \bibinfo{person}{Kai Zheng},
  \bibinfo{person}{Shuxin Zheng}, \bibinfo{person}{Chen Xing},
  \bibinfo{person}{Huishuai Zhang}, \bibinfo{person}{Yanyan Lan},
  \bibinfo{person}{Liwei Wang}, {and} \bibinfo{person}{Tieyan Liu}.}
  \bibinfo{year}{2020}\natexlab{}.
\newblock \showarticletitle{On layer normalization in the transformer
  architecture}. In \bibinfo{booktitle}{\emph{International Conference on
  Machine Learning}}. PMLR, \bibinfo{pages}{10524--10533}.
\newblock


\bibitem[Xu et~al\mbox{.}(2023)]%
        {Rui2023GCNO}
\bibfield{author}{\bibinfo{person}{Rui Xu}, \bibinfo{person}{Zhiyang Dou},
  \bibinfo{person}{Ningna Wang}, \bibinfo{person}{Shiqing Xin},
  \bibinfo{person}{Shuangmin Chen}, \bibinfo{person}{Mingyan Jiang},
  \bibinfo{person}{Xiaohu Guo}, \bibinfo{person}{Wenping Wang}, {and}
  \bibinfo{person}{Changhe Tu}.} \bibinfo{year}{2023}\natexlab{}.
\newblock \showarticletitle{Globally Consistent Normal Orientation for Point
  Clouds by Regularizing the Winding-Number Field}.
\newblock \bibinfo{journal}{\emph{ACM Trans. Graph.}} \bibinfo{volume}{42},
  \bibinfo{number}{4}, Article \bibinfo{articleno}{111} (\bibinfo{date}{jul}
  \bibinfo{year}{2023}), \bibinfo{numpages}{15}~pages.
\newblock
\showISSN{0730-0301}
\urldef\tempurl%
\url{https://doi.org/10.1145/3592129}
\showDOI{\tempurl}


\bibitem[Xu et~al\mbox{.}(2024)]%
        {xu2023dmv3d}
\bibfield{author}{\bibinfo{person}{Yinghao Xu}, \bibinfo{person}{Hao Tan},
  \bibinfo{person}{Fujun Luan}, \bibinfo{person}{Sai Bi}, \bibinfo{person}{Peng
  Wang}, \bibinfo{person}{Jiahao Li}, \bibinfo{person}{Zifan Shi},
  \bibinfo{person}{Kalyan Sunkavalli}, \bibinfo{person}{Gordon Wetzstein},
  \bibinfo{person}{Zexiang Xu}, {and} \bibinfo{person}{Kai Zhang}.}
  \bibinfo{year}{2024}\natexlab{}.
\newblock \showarticletitle{{DMV}3D: Denoising Multi-view Diffusion Using 3D
  Large Reconstruction Model}. In \bibinfo{booktitle}{\emph{The Twelfth
  International Conference on Learning Representations}}.
\newblock


\bibitem[Xue et~al\mbox{.}(2023)]%
        {xue2023ulip}
\bibfield{author}{\bibinfo{person}{Le Xue}, \bibinfo{person}{Ning Yu},
  \bibinfo{person}{Shu Zhang}, \bibinfo{person}{Junnan Li},
  \bibinfo{person}{Roberto Martín-Martín}, \bibinfo{person}{Jiajun Wu},
  \bibinfo{person}{Caiming Xiong}, \bibinfo{person}{Ran Xu},
  \bibinfo{person}{Juan~Carlos Niebles}, {and} \bibinfo{person}{Silvio
  Savarese}.} \bibinfo{year}{2023}\natexlab{}.
\newblock \bibinfo{title}{ULIP-2: Towards Scalable Multimodal Pre-training for
  3D Understanding}.
\newblock
\newblock
\showeprint[arxiv]{2305.08275}~[cs.CV]


\bibitem[Yariv et~al\mbox{.}(2024)]%
        {yariv2023mosaic}
\bibfield{author}{\bibinfo{person}{Lior Yariv}, \bibinfo{person}{Omri Puny},
  \bibinfo{person}{Natalia Neverova}, \bibinfo{person}{Oran Gafni}, {and}
  \bibinfo{person}{Yaron Lipman}.} \bibinfo{year}{2024}\natexlab{}.
\newblock \showarticletitle{Mosaic-SDF for 3D Generative Models}. In
  \bibinfo{booktitle}{\emph{Proceedings of the IEEE/CVF conference on computer
  vision and pattern recognition}}.
\newblock


\bibitem[Ye et~al\mbox{.}(2023)]%
        {ye2023ip-adapter}
\bibfield{author}{\bibinfo{person}{Hu Ye}, \bibinfo{person}{Jun Zhang},
  \bibinfo{person}{Sibo Liu}, \bibinfo{person}{Xiao Han}, {and}
  \bibinfo{person}{Wei Yang}.} \bibinfo{year}{2023}\natexlab{}.
\newblock \bibinfo{title}{IP-Adapter: Text Compatible Image Prompt Adapter for
  Text-to-Image Diffusion Models}.
\newblock
\newblock
\showeprint[arxiv]{2308.06721}~[cs.CV]


\bibitem[Yin et~al\mbox{.}(2023)]%
        {yin2023shapegpt}
\bibfield{author}{\bibinfo{person}{Fukun Yin}, \bibinfo{person}{Xin Chen},
  \bibinfo{person}{Chi Zhang}, \bibinfo{person}{Biao Jiang},
  \bibinfo{person}{Zibo Zhao}, \bibinfo{person}{Jiayuan Fan},
  \bibinfo{person}{Gang Yu}, \bibinfo{person}{Taihao Li}, {and}
  \bibinfo{person}{Tao Chen}.} \bibinfo{year}{2023}\natexlab{}.
\newblock \bibinfo{title}{ShapeGPT: 3D Shape Generation with A Unified
  Multi-modal Language Model}.
\newblock
\newblock
\showeprint[arxiv]{2311.17618}~[cs.CV]


\bibitem[Yu et~al\mbox{.}(2023b)]%
        {yu2023points}
\bibfield{author}{\bibinfo{person}{Chaohui Yu}, \bibinfo{person}{Qiang Zhou},
  \bibinfo{person}{Jingliang Li}, \bibinfo{person}{Zhe Zhang},
  \bibinfo{person}{Zhibin Wang}, {and} \bibinfo{person}{Fan Wang}.}
  \bibinfo{year}{2023}\natexlab{b}.
\newblock \showarticletitle{Points-to-3D: Bridging the Gap between Sparse
  Points and Shape-Controllable Text-to-3D Generation}. In
  \bibinfo{booktitle}{\emph{Proceedings of the 31st ACM International
  Conference on Multimedia}} (, Ottawa ON, Canada,) \emph{(\bibinfo{series}{MM
  '23})}. \bibinfo{publisher}{Association for Computing Machinery},
  \bibinfo{address}{New York, NY, USA}, \bibinfo{pages}{6841–6850}.
\newblock
\showISBNx{9798400701085}
\urldef\tempurl%
\url{https://doi.org/10.1145/3581783.3612232}
\showDOI{\tempurl}


\bibitem[Yu et~al\mbox{.}(2023a)]%
        {Yu_2023_CVPR}
\bibfield{author}{\bibinfo{person}{X. Yu}, \bibinfo{person}{M. Xu},
  \bibinfo{person}{Y. Zhang}, \bibinfo{person}{H. Liu}, \bibinfo{person}{C.
  Ye}, \bibinfo{person}{Y. Wu}, \bibinfo{person}{Z. Yan}, \bibinfo{person}{C.
  Zhu}, \bibinfo{person}{Z. Xiong}, \bibinfo{person}{T. Liang},
  \bibinfo{person}{G. Chen}, \bibinfo{person}{S. Cui}, {and}
  \bibinfo{person}{X. Han}.} \bibinfo{year}{2023}\natexlab{a}.
\newblock \showarticletitle{MVImgNet: A Large-scale Dataset of Multi-view
  Images}. In \bibinfo{booktitle}{\emph{2023 IEEE/CVF Conference on Computer
  Vision and Pattern Recognition (CVPR)}}. \bibinfo{publisher}{IEEE Computer
  Society}, \bibinfo{address}{Los Alamitos, CA, USA},
  \bibinfo{pages}{9150--9161}.
\newblock
\urldef\tempurl%
\url{https://doi.org/10.1109/CVPR52729.2023.00883}
\showDOI{\tempurl}


\bibitem[Zhang et~al\mbox{.}(2023c)]%
        {zhang20233dshape2vecset}
\bibfield{author}{\bibinfo{person}{Biao Zhang}, \bibinfo{person}{Jiapeng Tang},
  \bibinfo{person}{Matthias Nie\ss{}ner}, {and} \bibinfo{person}{Peter Wonka}.}
  \bibinfo{year}{2023}\natexlab{c}.
\newblock \showarticletitle{3DShape2VecSet: A 3D Shape Representation for
  Neural Fields and Generative Diffusion Models}.
\newblock \bibinfo{journal}{\emph{ACM Trans. Graph.}} \bibinfo{volume}{42},
  \bibinfo{number}{4}, Article \bibinfo{articleno}{92} (\bibinfo{date}{jul}
  \bibinfo{year}{2023}), \bibinfo{numpages}{16}~pages.
\newblock
\showISSN{0730-0301}
\urldef\tempurl%
\url{https://doi.org/10.1145/3592442}
\showDOI{\tempurl}


\bibitem[Zhang et~al\mbox{.}(2023a)]%
        {dreamface}
\bibfield{author}{\bibinfo{person}{Longwen Zhang}, \bibinfo{person}{Qiwei Qiu},
  \bibinfo{person}{Hongyang Lin}, \bibinfo{person}{Qixuan Zhang},
  \bibinfo{person}{Cheng Shi}, \bibinfo{person}{Wei Yang}, \bibinfo{person}{Ye
  Shi}, \bibinfo{person}{Sibei Yang}, \bibinfo{person}{Lan Xu}, {and}
  \bibinfo{person}{Jingyi Yu}.} \bibinfo{year}{2023}\natexlab{a}.
\newblock \showarticletitle{DreamFace: Progressive Generation of Animatable 3D
  Faces under Text Guidance}.
\newblock \bibinfo{journal}{\emph{ACM Trans. Graph.}} \bibinfo{volume}{42},
  \bibinfo{number}{4}, Article \bibinfo{articleno}{138} (\bibinfo{date}{jul}
  \bibinfo{year}{2023}), \bibinfo{numpages}{16}~pages.
\newblock
\showISSN{0730-0301}
\urldef\tempurl%
\url{https://doi.org/10.1145/3592094}
\showDOI{\tempurl}


\bibitem[Zhang et~al\mbox{.}(2023b)]%
        {zhang2023controlnet}
\bibfield{author}{\bibinfo{person}{L. Zhang}, \bibinfo{person}{A. Rao}, {and}
  \bibinfo{person}{M. Agrawala}.} \bibinfo{year}{2023}\natexlab{b}.
\newblock \showarticletitle{Adding Conditional Control to Text-to-Image
  Diffusion Models}. In \bibinfo{booktitle}{\emph{2023 IEEE/CVF International
  Conference on Computer Vision (ICCV)}}. \bibinfo{publisher}{IEEE Computer
  Society}, \bibinfo{address}{Los Alamitos, CA, USA},
  \bibinfo{pages}{3813--3824}.
\newblock
\urldef\tempurl%
\url{https://doi.org/10.1109/ICCV51070.2023.00355}
\showDOI{\tempurl}


\bibitem[Zhang et~al\mbox{.}(2023d)]%
        {zhang2023optimized}
\bibfield{author}{\bibinfo{person}{Youjia Zhang}, \bibinfo{person}{Junqing Yu},
  \bibinfo{person}{Zikai Song}, {and} \bibinfo{person}{Wei Yang}.}
  \bibinfo{year}{2023}\natexlab{d}.
\newblock \bibinfo{title}{Optimized View and Geometry Distillation from
  Multi-view Diffuser}.
\newblock
\newblock
\showeprint[arxiv]{2312.06198}~[cs.CV]


\bibitem[Zhao et~al\mbox{.}(2023)]%
        {zhao2023michelangelo}
\bibfield{author}{\bibinfo{person}{Zibo Zhao}, \bibinfo{person}{Wen Liu},
  \bibinfo{person}{Xin Chen}, \bibinfo{person}{Xianfang Zeng},
  \bibinfo{person}{Rui Wang}, \bibinfo{person}{Pei Cheng}, \bibinfo{person}{Bin
  Fu}, \bibinfo{person}{Tao Chen}, \bibinfo{person}{Gang Yu}, {and}
  \bibinfo{person}{Shenghua Gao}.} \bibinfo{year}{2023}\natexlab{}.
\newblock \showarticletitle{Michelangelo: Conditional 3d shape generation based
  on shape-image-text aligned latent representation}.
\newblock \bibinfo{journal}{\emph{Advances in neural information processing
  systems}} (\bibinfo{year}{2023}).
\newblock


\bibitem[Zheng et~al\mbox{.}(2023)]%
        {zheng2023locally}
\bibfield{author}{\bibinfo{person}{Xin-Yang Zheng}, \bibinfo{person}{Hao Pan},
  \bibinfo{person}{Peng-Shuai Wang}, \bibinfo{person}{Xin Tong},
  \bibinfo{person}{Yang Liu}, {and} \bibinfo{person}{Heung-Yeung Shum}.}
  \bibinfo{year}{2023}\natexlab{}.
\newblock \showarticletitle{Locally Attentional SDF Diffusion for Controllable
  3D Shape Generation}.
\newblock \bibinfo{journal}{\emph{ACM Trans. Graph.}} \bibinfo{volume}{42},
  \bibinfo{number}{4}, Article \bibinfo{articleno}{91} (\bibinfo{date}{jul}
  \bibinfo{year}{2023}), \bibinfo{numpages}{13}~pages.
\newblock
\showISSN{0730-0301}
\urldef\tempurl%
\url{https://doi.org/10.1145/3592103}
\showDOI{\tempurl}


\bibitem[Zhu et~al\mbox{.}(2024)]%
        {zhu2023hifa}
\bibfield{author}{\bibinfo{person}{Junzhe Zhu}, \bibinfo{person}{Peiye Zhuang},
  {and} \bibinfo{person}{Sanmi Koyejo}.} \bibinfo{year}{2024}\natexlab{}.
\newblock \showarticletitle{{HIFA}: High-fidelity Text-to-3D Generation with
  Advanced Diffusion Guidance}. In \bibinfo{booktitle}{\emph{The Twelfth
  International Conference on Learning Representations}}.
\newblock


\bibitem[Zou et~al\mbox{.}(2024)]%
        {zou2023triplane}
\bibfield{author}{\bibinfo{person}{Zi-Xin Zou}, \bibinfo{person}{Zhipeng Yu},
  \bibinfo{person}{Yuan-Chen Guo}, \bibinfo{person}{Yangguang Li},
  \bibinfo{person}{Ding Liang}, \bibinfo{person}{Yan-Pei Cao}, {and}
  \bibinfo{person}{Song-Hai Zhang}.} \bibinfo{year}{2024}\natexlab{}.
\newblock \showarticletitle{Triplane Meets Gaussian Splatting: Fast and
  Generalizable Single-View 3D Reconstruction with Transformers}. In
  \bibinfo{booktitle}{\emph{Proceedings of the IEEE/CVF conference on computer
  vision and pattern recognition}}.
\newblock


\end{thebibliography}

\end{document}